\documentclass[manuscript]{acmart}
\AtBeginDocument{%
  }

\usepackage{multirow}
\usepackage{multirow}
\usepackage{subcaption}

\begin{document}

\title{DMSC: Dynamic Multi-Scale Coordination Framework for Time Series Forecasting}

\author{Haonan Yang}
\email{yanghaonan21@nudt.edu.cn}

\author{Jianchao Tang}
\authornotemark[1]
\email{tangjianchao14@nudt.edu.cn}

\author{Zhuo Li}
\email{lizhuo19@nudt.edu.cn}
\affiliation{%
  \institution{National University of Defense Technology}
  \city{Changsha}
  \state{Hunan}
  \country{China}
}


\begin{abstract}
  Time Series Forecasting (TSF) faces persistent challenges in modeling intricate temporal dependencies across different scales. Despite recent advances leveraging different decomposition operations and novel architectures based on CNN, MLP or Transformer, existing methods still struggle with static decomposition strategies, fragmented dependency modeling, and inflexible fusion mechanisms, limiting their ability to model intricate temporal dependencies. To explicitly solve the mentioned three problems respectively, we propose a novel Dynamic Multi-Scale Coordination Framework (DMSC) with Multi-Scale Patch Decomposition block (EMPD), Triad Interaction Block (TIB) and Adaptive Scale Routing MoE block (ASR-MoE). Specifically, EMPD is designed as a built-in component to dynamically segment sequences into hierarchical patches with exponentially scaled granularities, eliminating predefined scale constraints through input-adaptive patch adjustment. TIB then jointly models intra-patch, inter-patch, and cross-variable dependencies within each layer's decomposed representations. EMPD and TIB are jointly integrated into layers forming a multi-layer progressive cascade architecture, where coarse-grained representations from earlier layers adaptively guide fine-grained feature extraction in subsequent layers via gated pathways. And ASR-MoE dynamically fuses multi-scale predictions by leveraging specialized global and local experts with temporal-aware weighting. Comprehensive experiments on thirteen real-world benchmarks demonstrate that DMSC consistently maintains state-of-the-art (SOTA) performance and superior computational efficiency for TSF tasks.
\end{abstract}




\maketitle

\section{Introduction}
Time Series Forecasting (TSF) constitutes a pivotal capability across numerous fields such as energy consumption \cite{Alvarez2011,Guo2015}, healthcare monitoring \cite{Luan2019,Kaimin2022}, transportation scheduling \cite{Chenjuan2020,Jin2021}, weather forecasting \cite{Kaifeng2023,Haixu2023}, economics \cite{Chen2023,Yux2023} and urban computing \cite{Zhong2024,Yuan2024}. Time series data exhibit intricate temporal dependencies, which include nonlinear relationships, multi-scale patterns (e.g., trends and seasonality), and dynamic variable couplings \cite{Qiuz2024}. Due to the inherent non-stationarity \cite{GuyPNason2006} and complex interconnections \cite{Seasonal2016}, traditional approaches often fail to adequately capture the underlying patterns in time series data, making effectively dependency modeling a critical challenge for enhancing predictive accuracy \cite{Zezhi2025}. In recent years, deep learning has demonstrated a strong capability to capture complex dependency relationships in time series data, making it a formidable tool for TSF \cite{Zonglei2023}.

However, intricate dependencies in time series inherently manifest across multiscales: coarser granularities typically encapsulate long-term trends while finer resolutions capture short-term fluctuations and periodic patterns. Conventional single-scale modeling approaches often fail to balance fine-grained local details and holistic global trends \cite{TimeMixer2025}. To effectively model multi-scale temporal dependencies, recent research has pioneered multi-scale modeling frameworks that concurrently extract global and local patterns, \textcolor{black}{thereby} mitigating the representational constraints inherent in single-scale approaches. While previous implementations often employ moving average downsampling \cite{TimeMixer2024,TimeKAN2025}, which remains insufficient in capturing local semantic information. In contrast, more effective alternatives focus on patch-wise and variable-wise decomposition operations that decompose \textcolor{black}{sequences} across \textcolor{black}{both} feature and temporal dimensions. Such decomposition operations better preserve both global trends and local semantics, and multi-length patches inherently facilitate the learning of multi-scale representations. Critically, prevailing methods suffer from two limitations: 1) reliance on fixed-scale decomposition strategies that lack dynamic scale adaptation; 2) fragmented processing of temporal dependencies and cross-variable interactions, which \textcolor{black}{compromise} holistic dependency modeling.

The primary advantage of multi-scale feature extraction lies in its capacity to mine complementary information across diverse temporal granularities and hierarchical levels, thereby overcoming the limitations of single-scale modeling \cite{TimeMixer2024}. Following feature capture, the effective design of prediction heads for multi-scale fusion is equally important. Conventional approaches typically employ linear projection layers \cite{TVNet2025}, additive combinations of multiple projections \cite{TimeMixer2024}, or spectral amplitude-weighted aggregation \cite{TimesNet2023}. However, these simplistic fusion methods exhibit fundamental inadequacies: they fail to leverage heterogeneous dominance patterns across multiscales, disregard dynamic inter-scale importance, and incur quadratic growth of model parameters with increasing input lengths and scale counts, thereby severely compromising deployment efficiency \cite{Qiuz2024}.

To effectively \textcolor{black}{model} complex temporal dependencies which inherently manifest across multiple patterns and scales, we propose a Dynamic Multi-Scale Coordination framework (DMSC) for TSF. This framework dynamically decomposes time series and extracts features across scales. It effectively captures intricate dependencies across temporal resolutions and cross-variable interactions. Meanwhile, it can generate adaptive predictions based on learned representations via temporal-aware fusion. Specifically, DMSC processes intricate dependencies through a multi-layer progressive architecture and an Adaptive Scale Routing Mixture-of-Experts (ASR-MoE). The multi-layer progressive cascade architecture incorporates Embedded Multi-Scale Patch Decomposition Block (EMPD), Triad Interaction Block (TIB), in which EMPD transforms the input into 3D representations at varying granularities and TIB models heterogeneous dependencies within each layer. In this architecture, coarse-grained representations from shallow layers adaptively guide fine-grained feature extraction in subsequent layers via gated pathways. After multi-scale feature extraction, ASR-MoE performs temporal-aware weighted dynamic fusion for multi-scale prediction, thus enabling adaptive integration of cross-scale features to \textcolor{black}{enhance} forecasting accuracy. DMSC formulates a unified framework that incorporates dynamic patch decomposition, deep models, and sparse MoE principles to achieve full-spectrum multi-scale coordination. \textcolor{black}{Rather than claiming an entirely new forecasting paradigm, DMSC presents a principled system-level integration.} This dynamic synergy enables the DMSC framework to achieve SOTA performance across 13 real-world benchmarks while maintaining high efficiency and low computational cost. Our contributions are as follows:

\begin{itemize}
	\item A novel multi-layer progressive cascade architecture is designed to jointly integrate Embedded Multi-Scale Patch Decomposition Block (EMPD) and Triad Interaction Block (TIB) across layers. Unlike fixed-scale approaches, EMPD dynamically adjusts patch granularities by employing a lightweight network based on the temporal characteristics of input data, and TIB jointly models intra-patch, inter-patch, and inter-variable dependencies through gated feature fusion to form a coarse-to-fine feature pyramid, where coarse-grained representations from earlier layers adaptively guide fine-grained extraction in subsequent layers via gated residual pathways.
	\item Adaptive Scale Routing Mixture-of-Experts (ASR-MoE) is proposed to resolve static fusion limitations in prediction. It establishes a hierarchical expert architecture to explicitly \textcolor{black}{decouple} long-term and short-term dependencies, in which global-shared experts capture common long-term temporal dependencies, while local-specialized experts model different short-term variations. Significantly, a temporal-aware weighting aggregator is designed to dynamically compute scale-specific prediction contributions with historical memory.
	\item Extensive experiments demonstrate that DMSC achieves SOTA performance and superior efficiency across multiple TSF benchmarks.
\end{itemize}

\section{Related Work}

\subsection{Time Series Forecasting Models}
Existing deep models for TSF tasks can be broadly categorized into MLP-based, CNN-based, \textcolor{black}{GNN-based} and Transformer-based architectures \cite{TSL2024}. MLP-based models \cite{chi2021t,NBEATS2019,FreTS2023} typically capture temporal dependencies through predefined decomposition and fully-connected layers. While these models demonstrate efficiency in capturing intra-series patterns, they exhibit notable limitations in modeling long-term dependencies and complex inter-series relationships. In contrast, CNN-based models \cite{TCN2018,SCINet2022,MiCN2023} leverage different convolutions to effectively capture local dependencies. Although they remain constrained in modeling global relationships, recent efforts have mitigated this through large-kernel convolutions \cite{moderntcn2024} and frequency-domain analysis \cite{TimesNet2023}. \textcolor{black}{Meanwhile, GNN-based methods have also emerged as a compelling paradigm for multivariate time series forecasting by explicitly modeling inter-variable dependencies as graph structures. Representative works include STGCN \cite{stgcn}, which integrates graph convolutions with temporal convolutions to capture spatio-temporal correlations; MTGNN \cite{mtgnn}, which learns a self-adaptive adjacency matrix to discover hidden spatial dependencies; and AGCRN \cite{agcrn}, which adaptively generates graph structures for each time step. However those methods typically rely on predefined or statically learned graphs, which may be insufficient to capture the dynamically evolving variable couplings inherent in time series.} Nevertheless, Transformer-based models \cite{wang20241,Autoformer2021,Informer2021} have emerged as a powerful paradigm, leveraging self-attention mechanisms \cite{Vaswani2017} to effectively capture long-range dependencies and persistent temporal relationships. However, Transformer-based models face substantial criticism due to their permutation invariance and quadratic computational complexity limitations in recent \textcolor{black}{research} \cite{tang2024,Ailing2023}. Despite significant progress, most existing deep architectures rely on single or fixed decomposition strategies, and often model dependencies in a fragmented manner, lacking mechanisms for dynamic, adaptive multi-scale decomposition and coordinated dependency modeling.

\subsection{Decomposition Operations in TSF}
To effectively capture intricate temporal dependencies, numerous studies have adopted a multi-scale modeling perspective, employing diverse decomposition operations on time series. TimesNet \cite{TimesNet2023} learns frequency-adaptive periods in the frequency domain and transforms series into 2D tensors, explicitly modeling intra-period and inter-period variations. TimeMixer \cite{TimeMixer2024} decomposes sequences into seasonal and trend components \cite{Autoformer2021}, then applies mixing strategies at different granularities to integrate multi-scale features. However, simply using moving average inadequately preserves local semantic patterns. Diverging from these operations, several methods focus on variable-wise and patch-wise disentanglement. PatchTST \cite{PatchTST2023} segments series into subseries-level patches as input tokens and employs channel independence to capture local semantics. ITransformer \cite{iTransformer2023} embeds entire series as single tokens to model extended global representations. TimeXer \cite{TimeXer2024} hierarchically represents variables through dual disentanglement, variable-wise for cross-channel interactions and patch-wise for endogenous variables. However, these decomposition techniques are constrained by static or predefined scale settings, hindering their adaptability to complex temporal patterns. \textcolor{black}{In contrast to these methods that apply decomposition, mixing, or disentanglement in relative isolation, DMSC integrates dynamic decomposition with dependency modeling and prediction fusion within a single coordinated framework, where the three stages actively interact rather than being pipelined.}

\textcolor{black}{Meanwhile, It is important to distinguish the proposed dynamic decomposition from conventional multi-scale architectures that apply multiple fixed patch sizes in parallel (e.g., \cite{liu2021, yifan2024}). Those approaches enumerate a predefined set of scales simultaneously, which introduces two limitations: (i) the scale set must be manually specified per dataset, and (ii) cross-scale interactions are confined to late fusion. In contrast, EMPD produces a single input-adaptive scale per layer and cascades layers hierarchically from coarse to fine, enabling early and progressive cross-scale guidance through gated pathways. This design trades parallel scale enumeration for dynamic scale selection and hierarchical refinement.}

\subsection{Multiscale Prediction Operations in TSF}
Beyond multi-scale decomposition operations, researchers have also investigated multi-scale operations in prediction heads. TimeMixer \cite{TimeMixer2024} employs dedicated projectors to generate scale-specific predictions. Diverging from simplistic additive aggregation, TimeMixer++ \cite{TimeMixer2025} utilizes frequency-domain amplitude analysis to perform weighted aggregation of projector outputs. TimeKAN \cite{TimeKAN2025} adopts frequency transformation and padding operations to unify multi-scale representations into identical dimensions, thereby enabling holistic scale integration. Current multi-scale fusion mechanisms predominantly rely on static weighting or fixed designs. They fail to dynamically prioritize the importance of different scales with intricate temporal dependencies. Different from these fragmented approaches, our proposed DMSC framework explores full-spectrum multi-scale coordination across embedding, extraction, and prediction stages, achieving both superior forecasting accuracy and high computational efficiency.

\begin{figure}[!t]
	\centering
	\includegraphics[width=\textwidth]{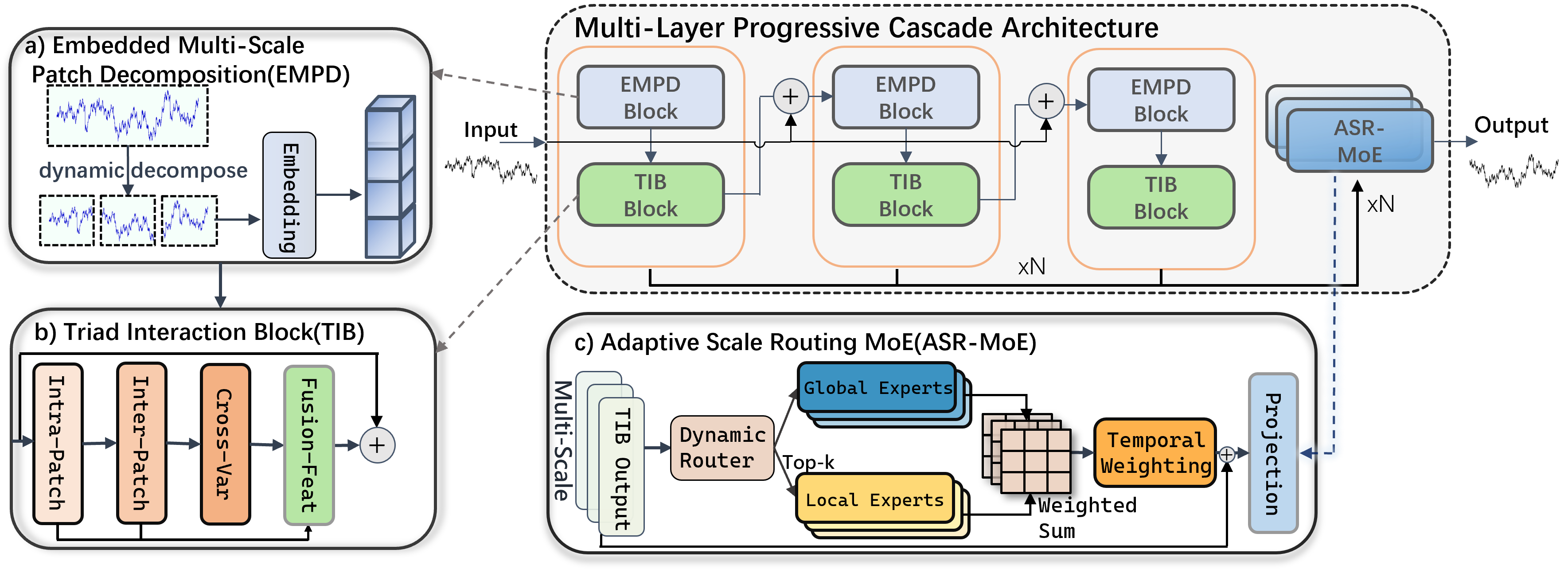}
	\caption{The overall architecture of the DMSC framework. EMPD dynamically segments input series into hierarchical patches, TIB jointly models three dependencies, and ASR-MoE fuses multiscale predictions adaptively.}
	\label{model}
\end{figure}

\section{Dynamic Multi-Scale Coordination Framework}
Time series forecasting addresses the fundamental challenge of predicting future values ${Y} = {y_{t+1}, ..., y_{t+h}} \in \mathbb{R}^{h\times{C}}$ from historical observations ${X} = {x_1, ..., x_t} \in \mathbb{R}^{t\times{C}}$, where $h$ is the prediction horizon, $t$ is the history horizon, and $C$ is the number of variables. To model complex multi-scale dependencies inherent in real-world time series, we propose the Dynamic Multi-Scale Coordination (DMSC) framework. \textcolor{black}{DMSC is not a collection of individual novelties, but a coordinated system where dynamic decomposition, triad interaction, and sparse expert fusion are jointly designed to interact - coarse scales guide fine scales, intra - and inter - patch features condition cross - variable gating, and multi - scale representations are routed adaptively to experts.} As illustrated in Figure \ref{model}, DMSC establishes full-spectrum multi-scale coordination across three key stages: embedding, feature extraction, and prediction, realized through EMPD, TIB and ASR-MoE components.

\textcolor{black}{(i) Embedding. The EMPD block (Section \ref{EMPD}) dynamically decomposes the input series into hierarchical patches, adapting the patch granularity to the input characteristics. (ii) Feature extraction. The TIB block (Section \ref{TIB}) jointly models intra-patch, inter-patch, and cross-variable dependencies within each layer, and the multi-layer cascade (Section \ref{MPCA}) progressively refines representations from coarse to fine. (iii) Prediction. The ASR-MoE block (Section \ref{ASR-MoE}) fuses multi-scale representations via a mixture of global and local experts with temporal-aware weighting. The following subsections detail each component.}

\subsection{Multi-Layer Progressive Cascade Architecture}\label{MPCA}
\textcolor{black}{Hierarchical coarse-to-fine architectures have been widely adopted in both computer vision \cite{Tsung2017, liu2021} and time series analysis \cite{TimeMixer2024, TimeKAN2025} to progressively refine representations across scales.} The multi-layer progressive cascade architecture facilitates hierarchical feature learning through stacked EMPD-TIB units, forming a hierarchical feature structure. Within this structure, coarse-grained representations from earlier layers adaptively guide fine-grained feature extraction in subsequent layers via gated pathways. Formally, given input $\mathbf{X} \in \mathbb{R}^{C \times L}$ \textcolor{black}{with $L$ the sequence length}, the $l$-th layer processes:

\begin{equation}
	\label{eq:Layer a}
	\mathcal{T} = \{\mathbf{F}_1, \mathbf{F}_2,..., \mathbf{F}_l\},
\end{equation}
\begin{equation}
	\label{eq:Layer b}
	\mathbf{F}_l = \textbf{TIB}_l(\mathbf{Z}_l),
\end{equation}
\begin{equation}
	\label{eq:Layer c}
	\mathbf{Z}_l = \textbf{EMPD}_l(\mathcal{P}_l(\mathbf{F}_{l-1}) \oplus \mathbf{X}), 
\end{equation}
where $\mathcal{T}$ denotes the set of hierarchical features generated by this architecture for ASR-MoE, with each $\mathbf{F}_l \in \mathbb{R}^{C \times{D}}$ representing the output of $l$-th layer, \textcolor{black}{where $D$ is the projection dimension}. Here, $\mathcal{P}_l$ represents a gated projection matrix that adaptively modulates the residual information flow. \textcolor{black}{And for the first layer ($l=1$), the gated pathway $\mathcal{P}_1$ and the previous layer's output $\mathbf{F}_0$ do not exist. Instead, EMPD directly processes the raw input:
\begin{equation}
    \label{eq:Layer d}
    \quad \mathbf{F}_1 = \textbf{TIB}_1(\mathbf{Z}_1),
\end{equation}
\begin{equation}
    \label{eq:Layer e}
    \mathbf{Z}_1 = \textbf{EMPD}_1(\mathbf{X}),
\end{equation}}
This cascade flow enables progressive refinement of multi-scale representations, where $\mathbf{F}_{l-1}$ adaptively guides the computation of $\mathbf{Z}_l$ with residual connections, and $\textbf{TIB}_l$ iteratively enhances representations by jointly modeling intra-patch, inter-patch, and cross-variable dependencies. By combining dynamic patch decomposition (EMPD) with triadic dependency modeling (TIB), our cascade architecture achieves input-aware progressive refinement of temporal granularities. This integration enables coherent modeling of both short-term dynamics and long-range trends while maintaining parameter efficiency. \textcolor{black}{The progressive cascade thus produces a set of hierarchical features $\mathcal{T} = {\mathbf{F}_1, \dots, \mathbf{F}_l}$. These features are subsequently fed to ASR-MoE for multi-scale prediction.}

\subsection{Embedded Multi-Scale Patch Decomposition}\label{EMPD}
Patch decomposition serves as a fundamental operation that transforms time series into structured representations, preserving local semantics while enabling hierarchical pattern discovery. \textcolor{black}{Unlike conventional single or fixed-scale approaches \cite{tang2024, Autoformer2021},} the EMPD block introduces adaptive and hierarchical decomposition mechanism as a built-in component, dynamically adjusting patch granularities based on sequence characteristics. This approach eliminates rigid predefined patch configurations while natively integrating multi-scale decomposition into our framework.

Specifically, given an input time series tensor $\mathbf{X} \in \mathbb{R}^{C\times{L}}$(where $C$ denotes the number of variates and $L$ the sequence length), EMPD first computes a scale factor $\alpha$ through a lightweight neural network:
\begin{equation}
	\label{eq:EMPD factor}
	\alpha = \mathcal{N}_{\theta}(\Phi_{\mathrm{GAP}}(\mathbf{X}))
\end{equation}
where $\mathcal{N}_{\theta}$ denotes a lightweight MLP with sigmoid activation, and $\Phi_{\mathrm{GAP}}$ denotes a global average pooling operation that compresses the temporal dimension. The factor $\alpha\in[0, 1]$ dynamically determines the base patch length, which subsequently undergoes exponential decay across layers as follows:
\begin{equation}
    \label{eq:EMPD patch1}
	\mathbf{P}_{{base}} = [\mathbf{P}_{min} + \alpha\cdot(\mathbf{P}_{max} - \mathbf{P}_{min})],
\end{equation}
\begin{equation}
	\label{eq:EMPD patch2}
	\mathbf{P}_{l} = \mathrm{max}(\mathbf{P}_{min}, [\mathbf{P}_{base}/\tau^l]),
\end{equation}
where $\mathbf{P}_{max}$ and $\mathbf{P}_{min}$ denote the parameters for minimum and maximum patch bounds, respectively. $\mathbf{P}_{l}$ represents the patch length at the $l$-th layer, \textcolor{black}{and $\tau$ is the patch decay rate that controls how rapidly the patch length shrinks from coarse to fine layers, with a default value of $\tau = 2$ in our experiments.} This design ensures that shallow layers process coarse-grained dependencies while deeper layers capture fine-grained ones. EMPD then applies replication padding to mitigate boundary effects, followed by hierarchical patch unfolding operation:
\begin{equation}
	\label{eq:EMPD unfold}
	\mathbf{X}_p^l = \mathrm{Unfold}_{P_l, S_l}(\mathrm{Padding}_{S_l}(X)),
\end{equation}
where $\mathrm{Padding}_{S_l}(\cdot)$ denotes the replication padding operation with stride \textcolor{black}{$S_l = P_l // 2$} and $\mathrm{Unfold}_{P_l, S_l}$ unfolds the padded sequence into patches of length $P_l$ using a stride of $S_l$. This operation produces a 3D patch tensor $\mathbf{X}_u^l \in \mathbb{R}^{C\times{N_l}\times{P_l}}$. Finally, EMPD projects $\mathbf{X}_p^l$ into a unified embedding space via a linear projection:
\begin{equation}
	\label{eq:EMPD proj}
	\mathbf{Z}^l = \mathrm{Projector}(\mathbf{X}_p^l),
\end{equation}
This direct linear projection efficiently maps patch-wise temporal features into a unified embedding space, avoiding redundant flattening operations while preserving the hierarchical structure of the multi-scale patches.

\textcolor{black}{In summary, EMPD produces a hierarchy of multi-scale patch embeddings ${\mathbf{Z}^l}_{l=1}^L$ that preserve both local semantics and global structure, forming the foundation for subsequent dependency modeling.}

\subsection{Triad Interaction Block}\label{TIB}
\textcolor{black}{Temporal dependencies in multivariate time series manifest along three orthogonal dimensions: local continuity within a time segment, long-range relationships across segments, and inter-series correlations. Modeling only a subset of these leads to fragmented representations. The Triad Interaction Block(TIB) is designed to capture all three jointly in a unified block.}

Specifically, TIB is designed to model heterogeneous dependencies within the multi-scale patch representations generated by EMPD, through joint capture of intra-patch, inter-patch, and cross-variable interactions. Given the input tensor $\mathbf{Z}^l \in\mathbb{R}^{C\times{N_l}\times{D}}$, TIB integrates these three complementary dependency types within a coherent framework. Specifically, TIB first processes intra-patch dependencies using depth-wise separable convolutions through operation $\mathbf{\zeta}_\text{intra}$ to capture fine-grained local patterns:
\begin{equation}
	\label{eq:TIB intra}
	\mathbf{F}_{\text{intra}}^l = \mathbf{\zeta}_\text{intra}(\mathbf{Z}^l),
\end{equation}
where $\mathbf{\zeta}_\text{intra}$ comprises depth-wise convolution and $\mathrm{Conv1d}$ to capture local temporal contexts and project features back to the embedding space. This operation preserves intra-patch continuity while effectively extracting local representations.

For inter-patch dynamics $\mathbf{F}_{\text{intra}}^l$, TIB employs dilated convolutions combined with adaptive pooling through operation $\mathbf{\zeta}_\text{inter}$:
\begin{equation}
	\label{eq:TIB inter}
	\mathbf{F}_{\text{inter}}^l = \mathbf{\zeta}_\text{inter}(\mathbf{F}_{\text{intra}}^l),
\end{equation}
where $\mathbf{\zeta}_\text{inter}$ consists of the dilated convolution and adaptive pooling operations, capturing broader temporal contexts and patch-level information without increasing computational cost. 

Then TIB models dependencies between different features by generating adaptive weighting coefficients $\mathbf{G}^l\in \mathbf{R}^{{C}\times{1}}$ from globally aggregated inter-patch features $\bar{\mathbf{F}}_{{inter}}^l \in \mathbf{R}^{{1}\times{D}}$ as follows:
\begin{equation}
	\label{eq:TIB inter1}
	\mathbf{G}^l = \sigma\left(\mathrm{MLP}(\bar{\mathbf{F}}_{\text{inter}}^l)\right),
\end{equation}
\begin{equation}
	\label{eq:TIB inter2}
	\mathbf{F}_{\text{cross}}^l = \mathbf{G}^l \odot \mathbf{F}_{\text{inter}}^l,
\end{equation}
where $\bar{\mathbf{F}}_{\text{inter}}^l$ denotes the global average of $\mathbf{F}_{\text{inter}}^l$, and the sigmoid function $\sigma$ acts as a feature-specific gate mechanism. $\mathbf{G}^l$ are applied element-wise to $\mathbf{F}_{\text{inter}}^l$ via the Hadamard product $\odot$, adaptively scaling each feature's contribution based on its global relevance and enabling context-aware cross-variable interactions. \textcolor{black}{The gating mechanism draws inspiration from the Squeeze-and-Excitation block \cite{sae2018}, adapted here to model cross-variable dependencies in multi-scale temporal representations.}

Finally, the three dependency representations are adaptively fused using learned gating weights. \textcolor{black}{After computing $\mathbf{F}_{\text{fused}}^l$ from the multi‑scale patches, we apply average pooling over the $N_l$ patch dimension of $\mathbf{Z}^l$ to obtain $\widetilde{\mathbf{Z}}^l \in \mathbb{R}^{C \times D}$,} and the resulting features are integrated with residual connection and normalization:
\begin{equation}
	\label{eq:TIB fuse}
	\mathbf{F}^l_{\text{fused}} = \sum_{i=1}^{3}\mathbf{G}^l_i\odot\mathbf{F}^l_i,
\end{equation}
\begin{equation}
	\label{eq:TIB out}
	\mathbf{F}^l_{\text{output}} = \mathrm{LayerNorm}(\mathbf{F}^l_{\text{fused}} + \text{AvgPool}_{N_l}(\mathbf{Z}^l)\big),
\end{equation}
where $\mathbf{G}^l_i$ represents the learned gating weights for the $i$-th representations. This architecture enables TIB to dynamically balance the contributions of different dependency types based on the temporal characteristics of input data. \textcolor{black}{Consequently, TIB enriches each scale's representation by jointly modeling three complementary forms of temporal dependency, yielding compact features $\mathbf{F}_l$ ready for multi-scale prediction.}

\subsection{Adaptive Scale Routing Mixture-of-Experts}\label{ASR-MoE}
\textcolor{black}{The mixture-of-experts (MoE) paradigm \cite{Shazeer2017} introduces conditional computation by routing inputs to specialized sub-networks. Recent work has adapted MoE to time series tasks \cite{timemoe2025}.} \textcolor{black}{However, existing multi-scale fusion methods typically use static linear combinations or simple weighted summation, which cannot capture the dynamic dominance of different scales under varying temporal conditions.} To address the limitations of static fusion mechanisms in multi-scale forecasting, we propose the Adaptive Scale Routing Mixture-of-Experts (ASR-MoE) block. This dynamic prediction head captures temporal dependencies at different horizons through a hierarchy of experts (global-shared and local-specialized experts) and adaptively aggregates scale-specific predictions based on temporal patterns. 

Specifically, ASR-MoE incorporates two distinct expert groups to handle different temporal granularities. Global experts $\mathcal{E}^g=\{\mathcal{G}_1, \mathcal{G}_2,..., \mathcal{G}_m\}$ capture common long-term dependencies through deep nonlinear transformations, modeling persistent trends and long-periodic patterns that commonly \textcolor{black}{exist} in time series. Local experts $\mathcal{E}^l=\{\mathcal{L}_1, \mathcal{L}_2,..., \mathcal{L}_n\}$ specialize in diverse short-term variations through shallower networks, detecting short-periodic patterns and high-frequency fluctuations that vary dynamically across stages. This explicit decoupling enables specialized handling of complex temporal dynamics while maintaining high parameter efficiency.

To compute the weighting for experts at each scale, a dynamic routing mechanism assigns input-dependent weights, thereby balancing the contributions of global and local dependencies. Given a scale-specific feature $\mathbf{F}_l$, the router generates weights, and only the top-$K$ local experts are activated via sparse routing:
\begin{equation}
    \label{eq:MoE 2}
    \mathbf{\hat{\Omega}}^l_L, \mathcal{I} = 
    \begin{cases} 
    \mathbf{\Omega}^l_L, & \mathbf{\Omega}^l_L \in \mathrm{Top-K}(\mathbf{\Omega}_l, K) \\ 
    0,                   &  \text{otherwise} 
    \end{cases}
\end{equation}
\begin{equation}
    \label{eq:MoE 11}
    \mathbf{\Omega}_G = \mathrm{Sigmoid}(\mathrm{MLP}(\mathbf{F}_l)),
\end{equation}
\begin{equation}
	\label{eq:MoE 1}
	\mathbf{\Omega}_L = \mathrm{Softmax}(\mathrm{MLP}(\mathbf{F}_l)),
\end{equation}

where $\mathbf{\Omega}_G$ and $\mathbf{\Omega}_L$ represent the global and local expert weight matrices, respectively. $\mathcal{I}$ denotes the set of indices for the selected experts. Finally, a temporal-aware weighting module fuses outputs from all scales by historical scale importance. The scale-specific predictions $\hat{\mathbf{Y}}^l$ are fused using time-dependent weights:
\begin{equation}
	\label{eq:MoE 3}
	\mathbf{w} = \mathrm{Softmax}\left(\mathrm{MLP}\left(\bigoplus^L_{l=1}\phi(\mathbf{F}_l)\cdot\mathbf{w}_{hist}\right)\right),
\end{equation}
where $\phi(\cdot)$ denotes temporal descriptors, $\mathbf{w}_{hist}$  represents the historical weighting memory, and $\bigoplus$ denotes the concatenation operation. \textcolor{black}{The learnable memory vector $\mathbf{w}_{hist}$ provides a stable, dataset - level prior over scale importance, preventing the scale weights from fluctuating excessively due to per - sample temporal descriptors alone and enabling the model to retain the dominant scale patterns learned from the training distribution.} The final prediction is obtained by integrating the multi-scale outputs:
\begin{equation}
	\label{eq:MoE 4}
	\hat{\mathbf{Y}} = \sum_{l=1}^{L}\mathbf{w}_l\left(\sum_{m=1}^{M}{\mathbf{\Omega}^m_G\mathcal{G}_m}(\mathbf{F}_l)+\sum_{n\in{\mathcal{I}}}{\hat{\mathbf{\Omega}}^n_L\mathcal{L}_n}(\mathbf{F}_l)\right),
\end{equation}

Meanwhile, \textcolor{black}{to encourage balanced utilization of all experts, ASR-MoE incorporates an auxiliary balance loss defined as the negative entropy of the routing weights:
\begin{equation}
	\label{eq:MoE 5}
	\mathcal{L}_{balance} = -\lambda \cdot \mathbb{E}\big[H(\Omega)\big] = -\lambda \cdot \mathbb{E}\Big[-\sum_j \Omega_j \log \Omega_j\Big],
\end{equation}
where $\lambda$ is a hyperparameter controlling the strength of the load-balancing regularization, and $\mathbb{E}[\cdot]$ denotes the empirical mean computed over the samples in the current mini-batch. Minimizing $\mathcal{L}_{balance}$ therefore maximizes the entropy, promoting uniform expert usage.}

The overall loss function for DMSC is defined as follows:
\begin{equation}
	\label{eq:MoE 6}
	\mathcal{L} = \mathcal{L}_{pred} + \mathcal{L}_{{balance}},
\end{equation}
where $\mathcal{L}_{pred}$ denotes the Mean Squared Error (MSE) loss.

By integrating hierarchical expert specialization, dynamic routing, and temporal-aware weighting, ASR-MoE adaptively prioritizes relevant scales and experts, achieving a synergistic balance between long-term trend capture and short-term detail refinement in time series forecasting.

\section{Experiments}
To comprehensively evaluate the performance and effectiveness of the proposed DMSC framework, we conduct extensive experiments on 13 real-world TSF benchmarks across multiple domains and temporal resolutions.

\subsubsection{Datasets.} For long-term forecasting, we conduct experiments on nine well-established benchmarks, including (1) ETT (ETTh1, ETTh2, ETTm1, ETTm2) contains 7 features of electricity transformer data from July 2016 to July 2018, which was sampled at hourly (ETTh1, ETTh2) and 15-minute (ETTm1, ETTm2) intervals. (2) Electricity (ECL) records hourly electricity consumption data of 321 clients from 2012 to 2014. (3) Exchange collects the daily exchange-rate data from eight different countries. (4) Solar-Energy contains the solar power production of 137 PV plants in 2006, which is sampled every 10 minutes. (5) Weather includes 21 meteorological factors collected every 10 minutes from the Max Planck Biogeochemistry Institute's Weather Station in 2020. (6) Traffic contains hourly road occupancy rates from 862 sensors on San Francisco freeways. Meanwhile, we conduct short-term forecasting experiments on PEMS (PEMS03, PEMS04, PEMS07, PEMS08), which contains the public traffic network data in California with a 5-minute interval. The details of datasets are listed in Table \ref{datasets}. 
\begin{table}[htbp]
	\centering
	\caption{Dataset Descriptions. Num is the number of variable. Dataset size is organized in (Train, Validation, Test).}
	\label{datasets}
	\resizebox{0.95\textwidth}{!}{
	\begin{tabular}{c c c c c c c}
		\hline
		\textbf{Name} & \textbf{Domain} & \textbf{Length} & \textbf{Num} & \textbf{Prediction Length} & \textbf{Dataset Size} & \textbf{Freq. (m)} \\
		\hline
		ETTh1 & Temperature & 14400 & 7 & \{96,192,336,720\} & (8545,2881,2881) & 60 \\
		\hline
		ETTh2 & Temperature & 14400 & 7 & \{96,192,336,720\} & (8545,2881,2881) & 60 \\
		\hline
		ETTm1 & Temperature & 57600 & 7 & \{96,192,336,720\} & (34465,11521,11521) & 15 \\
		\hline
		ETTm2 & Temperature & 57600 & 7 & \{96,192,336,720\} & (34465,11521,11521) & 15 \\
		\hline
		Electricity & Electricity & 26304 & 321 & \{96,192,336,720\} & (18317,2633,5261) & 60 \\
		\hline
		Exchange & Exchange Rate & 7588 & 8 & \{96,192,336,720\} & (5120,665,1422) & 1440 \\
		\hline
		Traffic & Road Occupancy & 17544 & 862 & \{96,192,336,720\} & (12185,1757,3509) & 60 \\
		\hline
		Weather & Weather & 52696 & 21 & \{96,192,336,720\} & (36792,5271,10540) & 10 \\
		\hline
		Solar-Energy & Energy & 52179 & 137 & \{96,192,336,720\} & (36601,5161,10417) & 10 \\
		\hline
		PEMS03 & Traffic Flow & 26208 & 358 & \{12,24,48,96\} & (15617,5135,5135) & 5 \\
		\hline
		PEMS04 & Traffic Flow & 16992 & 307 & \{12,24,48,96\} & (10172,3375,3375) & 5 \\
		\hline
		PEMS07 & Traffic Flow & 28224 & 883 & \{12,24,48,96\} & (16711,5622,5622) & 5 \\
		\hline
		PEMS08 & Traffic Flow & 17856 & 170 & \{12,24,48,96\} & (10690,3548,3548) & 5 \\
		\hline
	\end{tabular}
	}
\end{table}

\subsubsection{Implementation Details.} All experiments are implemented in PyTorch, and conducted on a single NVIDIA RTX 4090 24GB GPU. We utilize ADAM optimizer with an initial learning rate $10^{-3}$ and L2 loss for model optimization. For DMSC, we set layers of progressive cascade framework to 2 - 5, embedding dimension is set to {64, 128, 256, 512}. The patch decay rate is set to 2 for \textcolor{black}{exponential} degradation. \textcolor{black}{The validation set is used to monitor training and to select hyperparameters. After each epoch, we evaluate the model on the validation set and apply early stopping when the validation loss ceases to improve. The optimal architecture configuration, including the number of cascade layers $L$, is chosen based on validation performance. Final results are computed exclusively on the held-out test set. All hyperparameter configurations are as listed in Table \ref{hyperparams}.} All experiments are based on the framework of TimesNet. And all the baselines are implemented based on the configurations of original paper and its code.
\begin{table}[htbp]
\centering
\caption{Hyperparameter configurations of DMSC.}
\label{hyperparams}
\resizebox{0.7\textwidth}{!}{
\begin{tabular}{lcc}
\toprule
\textbf{Hyperparameter} & \textbf{Value / Range} & \textbf{Default} \\
\midrule
\multicolumn{3}{c}{\textit{Architecture}} \\
\midrule
Progressive cascade layers $L$ & $\{2, 3, 4, 5\}$ & $3$ \\
Embedding dimension $d_{\text{model}}$ & $\{64, 128, 256, 512\}$ & $128$ \\
EMPD patch bounds $(P_{\min}, P_{\max})$ & $(8, 64)$ & $(8, 64)$ \\
EMPD patch decay rate $\tau$ & $\{2, 3, 4\}$ & $2$ \\
\midrule
\multicolumn{3}{c}{\textit{ASR-MoE}} \\
\midrule
Global experts $m$ & $\{2, 4\}$ & $2$ \\
Local experts $n$ & $\{6, 8, 10\}$ & $8$ \\
Top-$K$ for local routing & $\{1, 2, 4\}$ & $2$ \\
Balance loss weight $\lambda$ & $\{0.1, 0.2, 0.3\}$ & $0.2$ \\
Expert hidden dimension & $\{d_{\text{model}}, 2d_{\text{model}}\}$ & $2d_{\text{model}}$ \\
Expert depth (global) & $\{2, 3, 4\}$ & $3$ \\
Expert depth (local) & $\{1, 2\}$ & $2$ \\
\midrule
\multicolumn{3}{c}{\textit{Training}} \\
\midrule
Optimizer & ADAM & ADAM \\
Learning rate & $10^{-3}$ & $10^{-3}$ \\
Batch size & $\{16, 32, 64\}$ & $32$ \\
Loss function & MSR & MSE \\
\midrule
\multicolumn{3}{c}{\textit{Data}} \\
\midrule
Look-back window & $96$ & $96$ \\
Prediction lengths & $\{96, 192, 336, 720\}$ & — \\
Normalization & Reversible instance normalization & — \\
\bottomrule
\end{tabular}
}
\vspace{-2mm}
\end{table}

\subsubsection{Baselines.} We compare our framework with ten SOTA models of TSF, including transformer-based models: iTransformer\cite{iTransformer2023}, PatchTST\cite{PatchTST2023}, Autoformer\cite{Autoformer2021}, TimeXer\cite{TimeXer2024}; MLP-based models: DLinear\cite{Ailing2023}, PatchMLP\cite{tang2024}, AMD\cite{AMD2025}; CNN-based models: TimesNet\cite{TimesNet2023}, TimeMixer\cite{TimeMixer2024}, and a novel architecture TimeKAN\cite{TimeKAN2025}.

\begin{table*}[!t]
	\centering
	\small
	\setlength{\tabcolsep}{2pt}
	\caption{Long-term forecasting results. All the results are selected from 4 different prediction lengths \{96, 192, 336, 720\}, and the look-back length is fixed to 96 for all baselines. A lower MSE or MAE indicates a better prediction, with the best in $\textbf{boldface}$ and second in \underline{underline}.}
	\label{compl}
	\resizebox{\textwidth}{!}{
		\begin{tabular}{c|c|cc|cc|cc|cc|cc|cc|cc|cc|cc|cc|cc}
			\toprule
			\multicolumn{2}{c}{\textbf{Models}} & \multicolumn{2}{c}{\textbf{DMSC(Ours)}} & \multicolumn{2}{c}{\textbf{TimeMixer}} & \multicolumn{2}{c}{\textbf{iTransformer}} & \multicolumn{2}{c}{\textbf{PatchTST}} & \multicolumn{2}{c}{\textbf{Dlinear}} & \multicolumn{2}{c}{\textbf{TimesNet}} & \multicolumn{2}{c}{\textbf{Autoformer}} & \multicolumn{2}{c}{\textbf{TimeXer}} & \multicolumn{2}{c}{\textbf{PatchMLP}} & \multicolumn{2}{c}{\textbf{TimeKAN}} & \multicolumn{2}{c}{\textbf{AMD}}\\
			\multicolumn{2}{c}{}& \multicolumn{2}{c|}{\textbf{(Ours)}} & \multicolumn{2}{c|}{\textbf{2024}} & \multicolumn{2}{c|}{\textbf{2023}} & \multicolumn{2}{c|}{\textbf{2023}} & \multicolumn{2}{c|}{\textbf{2023}} & \multicolumn{2}{c|}{\textbf{2023}} & \multicolumn{2}{c|}{\textbf{2021}} & \multicolumn{2}{c|}{\textbf{2024}} & \multicolumn{2}{c|}{\textbf{2024}} & \multicolumn{2}{c|}{\textbf{2025}} & \multicolumn{2}{c}{\textbf{2025}}\\
			\cmidrule(lr){3-4} \cmidrule(lr){5-6} \cmidrule(lr){7-8} \cmidrule(lr){9-10} \cmidrule(lr){11-12} \cmidrule(lr){13-14} \cmidrule(lr){15-16} \cmidrule(lr){17-18} \cmidrule(lr){19-20} \cmidrule(lr){21-22} \cmidrule(lr){23-24}
			\multicolumn{2}{c}{\textbf{Metric}} & MSE & MAE & MSE & MAE & MSE & MAE & MSE & MAE & MSE & MAE & MSE & MAE & MSE & MAE & MSE & MAE & MSE & MAE & MSE & MAE & MSE & MAE\\
			\midrule
			\multirow{5}{*}{\rotatebox{90}{ETTh1}} & 96 & $\textbf{0.370}$ & $\textbf{0.395}$ & $0.379$ & $\underline{0.397}$ & $0.387$ & $0.405$ & $\underline{0.378}$ & $0.399$ & $0.397$ & $0.412$ & $0.415$ & $0.429$ & $0.589$ & $0.526$ & $0.386$ & $0.404$ & $0.393$ & $0.405$ & $0.387$ & $0.401$ & $0.394$ & $0.404$\\
			& 192 & $\textbf{0.408}$ & $\textbf{0.420}$ & $0.430$ & $0.429$ & $0.441$ & $0.436$ & $0.427$ & $0.429$ & $0.446$ & $0.441$ & $0.479$ & $0.466$ & $0.653$ & $0.551$ & $0.429$ & $0.435$ & $0.443$ & $0.434$ & $\underline{0.415}$ & $\underline{0.423}$ & $0.444$ & $0.432$\\
			& 336 & $\textbf{0.421}$ & $\textbf{0.433}$ & $0.493$ & $0.459$ & $0.494$ & $0.463$ & $0.468$ & $0.455$ & $0.489$ & $0.467$ & $0.517$ & $0.482$ & $0.715$ & $0.581$ & $0.484$ & $0.457$ & $0.486$ & $0.456$ & $\underline{0.453}$ & $\underline{0.443}$ & $0.485$ & $0.451$\\
			& 720 & $\underline{0.462}$ & $\textbf{0.457}$ & $0.522$ & $0.493$ & $0.488$ & $0.483$ & $0.508$ & $0.497$ & $0.513$ & $0.511$ & $0.505$ & $0.490$ & $0.726$ & $0.601$ & $0.544$ & $0.513$ & $0.509$ & $0.487$ & $\textbf{0.461}$ & $\underline{0.463}$ & $0.486$ & $0.472$\\
			\cmidrule(lr){2-24}
			& Avg & $\textbf{0.415}$ & $\textbf{0.426}$ & $0.456$ & $0.444$ & $0.452$ & $0.446$ & $0.445$ & $0.445$ & $0.461$ & $0.457$ & $0.479$ & $0.466$ & $0.670$ & $0.564$ & $0.460$ & $0.452$ & $0.458$ & $0.445$ & $\underline{0.429}$ & $\underline{0.432}$ & $0.452$ & $0.439$\\
			\midrule
			\multirow{5}{*}{\rotatebox{90}{ETTh2}} & 96 & $\textbf{0.275}$ & $\textbf{0.329}$ & $0.290$ & $0.341$ & $0.301$ & $0.350$ & $0.295$ & $0.347$ & $0.341$ & $0.394$ & $0.316$ & $0.358$ & $0.443$ & $0.459$ & $\underline{0.284}$ & $\underline{0.337}$ & $0.311$ & $0.358$ & $0.291$ & $0.342$ & $0.397$ & $0.451$\\
			& 192 & $\textbf{0.359}$ & $\textbf{0.383}$ & $\underline{0.366}$ & $0.394$ & $0.380$ & $0.399$ & $0.378$ & $0.401$ & $0.482$ & $0.479$ & $0.415$ & $0.414$ & $0.500$ & $0.506$ & $\underline{0.366}$ & $\underline{0.391}$ & $0.404$ & $0.415$ & $0.376$ & $0.393$ & $0.501$ & $0.501$\\
			& 336 & $\textbf{0.376}$ & $\textbf{0.398}$ & $0.425$ & $0.433$ & $\underline{0.423}$ & $\underline{0.431}$ & $0.425$ & $0.442$ & $0.591$ & $0.541$ & $0.452$ & $0.448$ & $0.506$ & $0.502$ & $0.438$ & $0.438$ & $0.447$ & $0.451$ & $0.437$ & $0.443$ & $0.611$ & $0.563$\\
			& 720 & $0.412$ & $\textbf{0.424}$ & $\textbf{0.405}$ & $\underline{0.431}$ & $0.431$ & $0.447$ & $0.436$ & $0.456$ & $0.839$ & $0.661$ & $0.461$ & $0.463$ & $0.503$ & $0.509$ & $\underline{0.407}$ & $0.449$ & $0.462$ & $0.464$ & $0.451$ & $0.458$ & $0.956$ & $0.718$\\
			\cmidrule(lr){2-24}
			& Avg & $\textbf{0.355}$ & $\textbf{0.383}$ & $\underline{0.372}$ & $\underline{0.400}$ & $0.384$ & $0.407$ & $0.383$ & $0.412$ & $0.563$ & $0.519$ & $0.411$ & $0.421$ & $0.488$ & $0.494$ & $0.374$ & $0.404$ & $0.406$ & $0.422$ & $0.389$ & $0.409$ & $0.616$ & $0.558$\\
			\midrule
			\multirow{5}{*}{\rotatebox{90}{ETTm1}} & 96 & $\textbf{0.304}$ & $\textbf{0.335}$ & $0.319$ & $0.359$ & $0.341$ & $0.376$ & $0.326$ & $0.366$ & $0.346$ & $0.374$ & $0.336$ & $0.375$ & $0.564$ & $0.506$ & $\underline{0.318}$ & $\underline{0.356}$ & $0.321$ & $0.362$ & $0.326$ & $0.365$ & $0.331$ & $0.363$\\
			& 192 & $\textbf{0.346}$ & $\textbf{0.364}$ & $\underline{0.360}$ & $0.384$ & $0.382$ & $0.396$ & $0.365$ & $0.387$ & $0.382$ & $0.391$ & $0.377$ & $0.395$ & $0.586$ & $0.516$ & $0.373$ & $0.389$ & $0.364$ & $\underline{0.381}$ & $\underline{0.360}$ & $0.384$ & $0.373$ & $0.382$\\
			& 336 & $\underline{0.372}$ & $\underline{0.392}$ & $0.395$ & $0.406$ & $0.420$ & $0.421$ & $0.392$ & $0.406$ & $0.415$ & $0.415$ & $0.418$ & $0.420$ & $0.679$ & $0.547$ & $0.412$ & $\textbf{0.387}$ & $0.396$ & $0.404$ & $\textbf{0.369}$ & $0.402$ & $0.405$ & $0.403$\\
			& 720 & $\underline{0.451}$ & $\underline{0.440}$ & $0.461$ & $0.444$ & $0.487$ & $0.456$ & $0.461$ & $0.443$ & $0.473$ & $0.451$ & $0.541$ & $0.481$ & $0.715$ & $0.567$ & $0.460$ & $0.450$ & $0.468$ & $0.443$ & $\textbf{0.449}$ & $\textbf{0.437}$ & $0.467$ & $0.437$\\
			\cmidrule(lr){2-24}
			& Avg & $\textbf{0.368}$ & $\textbf{0.383}$ & $0.384$ & $0.398$ & $0.408$ & $0.412$ & $0.386$ & $0.400$ & $0.404$ & $0.408$ & $0.418$ & $0.418$ & $0.636$ & $0.534$ & $0.391$ & $0.395$ & $0.387$ & $0.398$ & $\underline{0.376}$ & $0.397$ & $0.394$ & $\underline{0.396}$\\
			\midrule
			\multirow{5}{*}{\rotatebox{90}{ETTm2}} & 96 & $\underline{0.174}$ & $\underline{0.256}$ & $0.179$ & $0.261$ & $0.186$ & $0.272$ & $0.184$ & $0.269$ & $0.193$ & $0.293$ & $0.188$ & $0.268$ & $0.554$ & $0.469$ & $\textbf{0.172}$ & $\textbf{0.254}$ & $0.176$ & $0.259$ & $0.176$ & $0.263$ & $0.187$ & $0.271$\\
			& 192 & $\textbf{0.233}$ & $\textbf{0.295}$ & $\underline{0.238}$ & $\underline{0.301}$ & $0.252$ & $0.312$ & $0.247$ & $0.307$ & $0.284$ & $0.361$ & $0.250$ & $0.306$ & $0.609$ & $0.497$ & $0.241$ & $0.302$ & $0.246$ & $0.304$ & $0.239$ & $\underline{0.301}$ & $0.251$ & $0.309$\\
			& 336 & $\textbf{0.296}$ & $\textbf{0.331}$ & $\underline{0.299}$ & $\underline{0.339}$ & $0.315$ & $0.351$ & $0.313$ & $0.354$ & $0.382$ & $0.429$ & $0.306$ & $0.341$ & $0.401$ & $0.409$ & $0.301$ & $0.340$ & $0.309$ & $0.344$ & $0.304$ & $0.346$ & $0.309$ & $0.346$\\
			& 720 & $\textbf{0.370}$ & $\textbf{0.385}$ & $0.395$ & $\underline{0.394}$ & $0.415$ & $0.408$ & $0.409$ & $0.406$ & $0.558$ & $0.525$ & $0.420$ & $0.405$ & $0.443$ & $0.433$ & $\underline{0.394}$ & $0.395$ & $0.417$ & $0.408$ & $0.410$ & $0.408$ & $0.407$ & $0.400$\\
			\cmidrule(lr){2-24}
			& Avg & $\textbf{0.268}$ & $\textbf{0.317}$ & $0.278$ & $0.324$ & $0.292$ & $0.336$ & $0.288$ & $0.334$ & $0.354$ & $0.402$ & $0.291$ & $0.330$ & $0.502$ & $0.452$ & $\underline{0.277}$ & $\underline{0.323}$ & $0.287$ & $0.329$ & $0.282$ & $0.330$ & $0.288$ & $0.332$\\
			\midrule
			\multirow{5}{*}{\rotatebox{90}{Electricity}} & 96 & $\textbf{0.138}$ & $\textbf{0.223}$ & $0.161$ & $0.252$ & $\underline{0.148}$ & $\underline{0.241}$ & $0.181$ & $0.274$ & $0.211$ & $0.302$ & $0.163$ & $0.267$ & $0.232$ & $0.347$ & $0.241$ & $0.244$ & $0.167$ & $0.264$ & $0.177$ & $0.270$ & $0.185$ & $0.267$\\
			& 192 & $\underline{0.160}$ & $\textbf{0.258}$ & $0.176$ & $0.269$ & $0.167$ & $0.248$ & $0.187$ & $0.280$ & $0.211$ & $0.305$ & $0.184$ & $0.284$ & $0.363$ & $0.447$ & $\textbf{0.159}$ & $\underline{0.260}$ & $0.181$ & $0.276$ & $0.185$ & $0.276$ & $0.190$ & $0.272$\\
			& 336 & $\textbf{0.167}$ & $\textbf{0.253}$ & $0.193$ & $0.283$ & $0.179$ & $\underline{0.271}$ & $0.204$ & $0.296$ & $0.223$ & $0.319$ & $0.196$ & $0.297$ & $0.599$ & $0.595$ & $\underline{0.177}$ & $0.276$ & $0.203$ & $0.303$ & $0.201$ & $0.292$ & $0.205$ & $0.288$\\
			& 720 & $\underline{0.213}$ & $\underline{0.299}$ & $0.232$ & $0.316$ & $\textbf{0.208}$ & $\textbf{0.298}$ & $0.246$ & $0.328$ & $0.258$ & $0.351$ & $0.232$ & $0.325$ & $0.775$ & $0.701$ & $0.229$ & $0.321$ & $0.251$ & $0.341$ & $0.241$ & $0.323$ & $0.246$ & $0.321$\\
			\cmidrule(lr){2-24}
			& Avg & $\textbf{0.170}$ & $\textbf{0.258}$ & $0.190$ & $0.280$ & $\underline{0.176}$ & $\underline{0.265}$ & $0.205$ & $0.295$ & $0.226$ & $0.319$ & $0.194$ & $0.293$ & $0.492$ & $0.523$ & $0.201$ & $0.275$ & $0.200$ & $0.296$ & $0.201$ & $0.290$ & $0.206$ & $0.287$\\
			\midrule
			\multirow{5}{*}{\rotatebox{90}{Exchange}} & 96 & $\textbf{0.082}$ & $\textbf{0.201}$ & $\underline{0.087}$ & $\underline{0.204}$ & $0.088$ & $0.208$ & $0.094$ & $0.213$ & $0.098$ & $0.233$ & $0.115$ & $0.242$ & $0.158$ & $0.290$ & $0.094$ & $0.214$ & $0.094$ & $0.217$ & $0.092$ & $0.212$ & $0.088$ & $0.208$\\
			& 192 & $\textbf{0.174}$ & $\underline{0.298}$ & $\underline{0.177}$ & $\textbf{0.295}$ & $0.180$ & $0.303$ & $0.182$ & $0.303$ & $0.186$ & $0.325$ & $0.216$ & $0.333$ & $0.299$ & $0.406$ & $0.182$ & $0.303$ & $0.187$ & $0.311$ & $0.180$ & $0.300$ & $0.182$ & $0.305$\\
			& 336 & $\textbf{0.315}$ & $\textbf{0.405}$ & $0.328$ & $\underline{0.414}$ & $0.331$ & $0.418$ & $0.347$ & $0.426$ & $\underline{0.325}$ & $0.434$ & $0.375$ & $0.444$ & $0.470$ & $0.511$ & $0.384$ & $0.448$ & $0.342$ & $0.424$ & $0.352$ & $0.430$ & $0.332$ & $0.417$\\
			& 720 & $\underline{0.772}$ & $\textbf{0.659}$ & $0.847$ & $0.692$ & $0.848$ & $0.695$ & $0.931$ & $0.724$ & $\textbf{0.746}$ & $0.663$ & $1.012$ & $0.765$ & $1.228$ & $0.869$ & $0.932$ & $0.724$ & $0.908$ & $0.715$ & $0.912$ & $0.715$ & $0.861$ & $\underline{0.701}$\\
			\cmidrule(lr){2-24}
			& Avg & $\textbf{0.336}$ & $\textbf{0.391}$ & $0.359$ & $\underline{0.401}$ & $0.362$ & $0.406$ & $0.389$ & $0.417$ & $\underline{0.339}$ & $0.414$ & $0.430$ & $0.446$ & $0.539$ & $0.519$ & $0.398$ & $0.422$ & $0.383$ & $0.417$ & $0.384$ & $0.414$ & $0.366$ & $0.408$\\
			\midrule
			\multirow{5}{*}{\rotatebox{90}{Weather}} & 96 & $\underline{0.160}$ & $\underline{0.210}$ & $0.164$ & $\underline{0.210}$ & $0.176$ & $0.216$ & $0.174$ & $0.215$ & $0.195$ & $0.252$ & $0.172$ & $0.221$ & $0.301$ & $0.364$ & $\textbf{0.157}$ & $\textbf{0.205}$ & $0.168$ & $0.214$ & $0.163$ & $0.210$ & $0.194$ & $0.236$\\
			& 192 & $\underline{0.207}$ & $\underline{0.250}$ & $0.208$ & $0.251$ & $0.225$ & $0.257$ & $0.221$ & $0.256$ & $0.239$ & $0.299$ & $0.220$ & $0.260$ & $0.335$ & $0.385$ & $\textbf{0.204}$ & $\textbf{0.248}$ & $0.215$ & $0.255$ & $0.209$ & $0.251$ & $0.239$ & $0.271$\\
			& 336 & $\textbf{0.253}$ & $\textbf{0.284}$ & $0.264$ & $0.292$ & $0.281$ & $0.299$ & $0.280$ & $0.297$ & $0.282$ & $0.333$ & $0.280$ & $0.302$ & $0.352$ & $0.376$ & $0.264$ & $0.293$ & $0.272$ & $0.295$ & $\underline{0.263}$ & $\underline{0.291}$ & $0.290$ & $0.306$\\
			& 720 & $\textbf{0.313}$ & $\textbf{0.332}$ & $0.343$ & $\underline{0.343}$ & $0.361$ & $0.353$ & $0.357$ & $0.349$ & $0.345$ & $0.381$ & $0.353$ & $0.350$ & $0.405$ & $0.401$ & $0.343$ & $\underline{0.343}$ & $0.351$ & $0.346$ & $\underline{0.341}$ & $0.344$ & $0.362$ & $0.352$\\
			\cmidrule(lr){2-24}
			& Avg & $\textbf{0.233}$ & $\textbf{0.269}$ & $0.245$ & $0.274$ & $0.261$ & $0.281$ & $0.258$ & $0.279$ & $0.265$ & $0.316$ & $0.256$ & $0.283$ & $0.348$ & $0.382$ & $\underline{0.242}$ & $\underline{0.272}$ & $0.252$ & $0.277$ & $0.244$ & $0.274$ & $0.271$ & $0.291$\\
			\midrule
			\multirow{5}{*}{\rotatebox{90}{Traffic}} & 96 & $\textbf{0.389}$ & $\textbf{0.259}$ & $0.476$ & $0.292$ & $\underline{0.393}$ & $\underline{0.269}$ & $0.459$ & $0.299$ & $0.712$ & $0.438$ & $0.593$ & $0.317$ & $0.663$ & $0.403$ & $0.428$ & $0.271$ & $0.513$ & $0.352$ & $0.598$ & $0.382$ & $0.546$ & $0.346$\\
			& 192 & $\textbf{0.405}$ & $\textbf{0.258}$ & $0.501$ & $0.301$ & $\underline{0.412}$ & $\underline{0.277}$ & $0.469$ & $0.303$ & $0.662$ & $0.417$ & $0.618$ & $0.327$ & $0.915$ & $0.557$ & $0.447$ & $0.280$ & $0.509$ & $0.350$ & $0.579$ & $0.365$ & $0.529$ & $0.335$\\
			& 336 & $\textbf{0.398}$ & $\underline{0.286}$ & $0.514$ & $0.314$ & $\underline{0.424}$ & $\textbf{0.283}$ & $0.483$ & $0.309$ & $0.669$ & $0.419$ & $0.642$ & $0.341$ & $1.217$ & $0.704$ & $0.472$ & $0.289$ & $0.533$ & $0.360$ & $0.572$ & $0.361$ & $0.540$ & $0.339$\\
			& 720 & $\textbf{0.437}$ & $\textbf{0.290}$ & $0.545$ & $0.320$ & $\underline{0.459}$ & $\underline{0.301}$ & $0.518$ & $0.326$ & $0.709$ & $0.437$ & $0.679$ & $0.350$ & $1.317$ & $0.755$ & $0.517$ & $0.307$ & $0.599$ & $0.395$ & $0.609$ & $0.384$ & $0.576$ & $0.358$\\
			\cmidrule(lr){2-24}
			& Avg & $\textbf{0.407}$ & $\textbf{0.274}$ & $0.509$ & $0.307$ & $\underline{0.422}$ & $\underline{0.283}$ & $0.482$ & $0.309$ & $0.688$ & $0.428$ & $0.633$ & $0.334$ & $1.028$ & $0.605$ & $0.466$ & $0.287$ & $0.539$ & $0.364$ & $0.590$ & $0.373$ & $0.547$ & $0.345$\\
			\midrule
			\multirow{5}{*}{\rotatebox{90}{Solar}} & 96 & $\textbf{0.187}$ & $\textbf{0.231}$ & $\underline{0.196}$ & $0.263$ & $0.207$ & $\underline{0.237}$ & $0.216$ & $0.274$ & $0.290$ & $0.378$ & $0.223$ & $0.256$ & $0.552$ & $0.524$ & $0.198$ & $0.244$ & $0.239$ & $0.272$ & $0.248$ & $0.302$ & $0.310$ & $0.312$\\
			& 192 & $\textbf{0.216}$ & $\textbf{0.261}$ & $0.245$ & $0.279$ & $0.242$ & $\underline{0.264}$ & $0.250$ & $0.294$ & $0.320$ & $0.398$ & $0.262$ & $0.272$ & $0.696$ & $0.605$ & $\underline{0.226}$ & $0.270$ & $0.291$ & $0.298$ & $0.291$ & $0.321$ & $0.348$ & $0.330$\\
			& 336 & $\textbf{0.223}$ & $\textbf{0.269}$ & $\underline{0.235}$ & $0.287$ & $0.251$ & $\underline{0.274}$ & $0.265$ & $0.302$ & $0.353$ & $0.415$ & $0.287$ & $0.299$ & $0.816$ & $0.677$ & $0.239$ & $0.281$ & $0.325$ & $0.316$ & $0.307$ & $0.332$ & $0.399$ & $0.355$\\
			& 720 & $\textbf{0.226}$ & $\textbf{0.271}$ & $\underline{0.236}$ & $0.286$ & $0.251$ & $0.276$ & $0.266$ & $0.298$ & $0.357$ & $0.413$ & $0.298$ & $0.318$ & $0.844$ & $0.731$ & $0.242$ & $\underline{0.282}$ & $0.336$ & $0.321$ & $0.310$ & $0.329$ & $0.393$ & $0.349$\\
			\cmidrule(lr){2-24}
			& Avg & $\textbf{0.213}$ & $\textbf{0.258}$ & $0.228$ & $0.279$ & $0.238$ & $\underline{0.263}$ & $0.249$ & $0.292$ & $0.330$ & $0.401$ & $0.268$ & $0.286$ & $0.727$ & $0.634$ & $\underline{0.226}$ & $0.269$ & $0.298$ & $0.301$ & $0.289$ & $0.321$ & $0.363$ & $0.337$\\
			\midrule
			\multicolumn{2}{c|}{\textbf{1st Count}} & $\textbf{26}$ & $\textbf{28}$ & $1$ & $1$ & $1$ & $2$ & $0$ & $0$ & $1$ & $0$ & $0$ & $0$ & $0$ & $0$ & $4$ & $4$ & $0$ & $0$ & $3$ & $1$ & $0$ & $0$ \\
			\bottomrule
		\end{tabular}
	}
\end{table*}

\begin{table*}[!t]
	\centering
	\small
	\setlength{\tabcolsep}{2pt}
	\caption{Short-term forecasting results. All the results are selected from 4 different prediction lengths \{12, 24, 48, 96\}, and the look-back length is fixed to 96 for all baselines.}
	\label{comps}
    \resizebox{\textwidth}{!}{
	\begin{tabular}{c|c|cc|cc|cc|cc|cc|cc|cc|cc|cc|cc|cc}
		\toprule
		\multicolumn{2}{c}{\textbf{Models}} & \multicolumn{2}{c}{\textbf{DMSC}} & \multicolumn{2}{c}{\textbf{TimeMixer}} & \multicolumn{2}{c}{\textbf{iTransformer}} & \multicolumn{2}{c}{\textbf{PatchTST}} & \multicolumn{2}{c}{\textbf{Dlinear}} & \multicolumn{2}{c}{\textbf{TimesNet}} & \multicolumn{2}{c}{\textbf{Autoformer}} & \multicolumn{2}{c}{\textbf{TimeXer}} & \multicolumn{2}{c}{\textbf{PatchMLP}} & \multicolumn{2}{c}{\textbf{TimeKAN}} & \multicolumn{2}{c}{\textbf{AMD}}\\
		 \multicolumn{2}{c}{}& \multicolumn{2}{c|}{\textbf{(Ours)}} & \multicolumn{2}{c|}{\textbf{2024}} & \multicolumn{2}{c|}{\textbf{2023}} & \multicolumn{2}{c|}{\textbf{2023}} & \multicolumn{2}{c|}{\textbf{2023}} & \multicolumn{2}{c|}{\textbf{2023}} & \multicolumn{2}{c|}{\textbf{2021}} & \multicolumn{2}{c|}{\textbf{2024}} & \multicolumn{2}{c|}{\textbf{2024}} & \multicolumn{2}{c|}{\textbf{2025}} & \multicolumn{2}{c}{\textbf{2025}}\\
		\cmidrule(lr){3-4} \cmidrule(lr){5-6} \cmidrule(lr){7-8} \cmidrule(lr){9-10} \cmidrule(lr){11-12} \cmidrule(lr){13-14} \cmidrule(lr){15-16} \cmidrule(lr){17-18} \cmidrule(lr){19-20} \cmidrule(lr){21-22} \cmidrule(lr){23-24}
		\multicolumn{2}{c}{\textbf{Metric}} & MSE & MAE & MSE & MAE & MSE & MAE & MSE & MAE & MSE & MAE & MSE & MAE & MSE & MAE & MSE & MAE & MSE & MAE & MSE & MAE & MSE & MAE\\
		\midrule
		\multirow{5}{*}{\rotatebox{90}{PEMS03}} & 12 & $\textbf{0.066}$ & $\textbf{0.171}$ & $0.084$ & $0.194$ & $0.069$ & $\underline{0.175}$ & $0.105$ & $0.216$ & $0.122$ & $0.245$ & $0.088$ & $0.195$ & $0.224$ & $0.346$ & $\underline{0.068}$ & $0.179$ & $0.118$ & $0.191$ & $0.095$ & $0.210$ & $0.106$ & $0.225$\\
		& 24 & $\underline{0.092}$ & $\textbf{0.203}$ & $0.130$ & $0.244$ & $0.099$ & $0.210$ & $0.198$ & $0.296$ & $0.202$ & $0.320$ & $0.118$ & $0.224$ & $0.492$ & $0.513$ & $\textbf{0.089}$ & $\underline{0.204}$ & $0.204$ & $0.231$ & $0.166$ & $0.281$ & $0.174$ & $0.294$\\
		& 48 & $\textbf{0.136}$ & $\textbf{0.251}$ & $0.218$ & $0.317$ & $0.164$ & $0.275$ & $0.472$ & $0.466$ & $0.334$ & $0.428$ & $0.169$ & $0.268$ & $0.392$ & $0.459$ & $\underline{0.137}$ & $\underline{0.253}$ & $0.213$ & $0.314$ & $0.314$ & $0.393$ & $0.333$ & $0.417$\\
		& 96 & $\textbf{0.230}$ & $\underline{0.339}$ & $0.327$ & $0.398$ & $0.711$ & $0.651$ & $0.458$ & $0.493$ & $0.459$ & $0.517$ & $\underline{0.239}$ & $\textbf{0.330}$ & $0.944$ & $0.749$ & $0.427$ & $0.483$ & $0.347$ & $0.421$ & $0.558$ & $0.543$ & $0.510$ & $0.536$\\
		\cmidrule(lr){2-24}
		& Avg & $\textbf{0.131}$ & $\textbf{0.241}$ & $0.190$ & $0.288$ & $0.261$ & $0.328$ & $0.308$ & $0.368$ & $0.279$ & $0.377$ & $\underline{0.154}$ & $\underline{0.254}$ & $0.513$ & $0.517$ & $0.180$ & $0.280$ & $0.220$ & $0.289$ & $0.283$ & $0.357$ & $0.281$ & $0.368$\\
		\midrule
		\multirow{5}{*}{\rotatebox{90}{PEMS04}} & 12 & $\textbf{0.078}$ & $\textbf{0.183}$ & $0.105$ & $0.216$ & $0.766$ & $0.709$ & $0.116$ & $0.230$ & $0.147$ & $0.272$ & $\underline{0.092}$ & $\underline{0.202}$ & $0.211$ & $0.341$ & $0.293$ & $0.397$ & $0.109$ & $0.204$ & $0.107$ & $0.222$ & $0.131$ & $0.256$\\
		& 24 & $\textbf{0.102}$ & $\textbf{0.215}$ & $0.168$ & $0.280$ & $0.799$ & $0.728$ & $0.216$ & $0.314$ & $0.225$ & $0.340$ & $\underline{0.111}$ & $\underline{0.224}$ & $0.394$ & $0.471$ & $0.308$ & $0.409$ & $0.129$ & $0.248$ & $0.178$ & $0.294$ & $0.196$ & $0.311$\\
		& 48 & $\textbf{0.147}$ & $\textbf{0.261}$ & $0.270$ & $0.359$ & $1.041$ & $0.882$ & $0.503$ & $0.489$ & $0.356$ & $0.437$ & $\underline{0.152}$ & $\underline{0.266}$ & $0.429$ & $0.463$ & $0.339$ & $0.425$ & $0.213$ & $0.326$ & $0.329$ & $0.409$ & $0.344$ & $0.420$\\
		& 96 & $\textbf{0.190}$ & $\underline{0.316}$ & $0.377$ & $0.439$ & $1.045$ & $0.886$ & $0.623$ & $0.586$ & $0.453$ & $0.505$ & $\underline{0.197}$ & $\textbf{0.308}$ & $0.853$ & $0.703$ & $0.367$ & $0.441$ & $0.361$ & $0.441$ & $0.572$ & $0.563$ & $0.638$ & $0.602$\\
		\cmidrule(lr){2-24}
		& Avg & $\textbf{0.129}$ & $\textbf{0.244}$ & $0.230$ & $0.324$ & $0.913$ & $0.801$ & $0.365$ & $0.405$ & $0.295$ & $0.389$ & $\underline{0.138}$ & $\underline{0.250}$ & $0.472$ & $0.495$ & $0.327$ & $0.418$ & $0.203$ & $0.305$ & $0.297$ & $0.372$ & $0.327$ & $0.397$\\
		\midrule
		\multirow{5}{*}{\rotatebox{90}{PEMS07}} & 12 & $\textbf{0.059}$ & $\textbf{0.157}$ & $0.070$ & $0.173$ & $0.068$ & $0.169$ & $0.093$ & $0.206$ & $0.116$ & $0.241$ & $0.075$ & $0.179$ & $0.207$ & $0.335$ & $\underline{0.061}$ & $\underline{0.165}$ & $0.107$ & $0.178$ & $0.085$ & $0.199$ & $0.096$ & $0.223$\\
		& 24 & $\underline{0.078}$ & $\underline{0.179}$ & $0.109$ & $0.215$ & $0.087$ & $0.190$ & $0.195$ & $0.295$ & $0.209$ & $0.327$ & $0.083$ & $0.198$ & $0.314$ & $0.412$ & $\textbf{0.071}$ & $\textbf{0.177}$ & $0.112$ & $0.216$ & $0.149$ & $0.268$ & $0.231$ & $0.360$\\
		& 48 & $\underline{0.104}$ & $\underline{0.211}$ & $0.199$ & $0.296$ & $0.122$ & $0.231$ & $0.485$ & $0.469$ & $0.397$ & $0.456$ & $0.128$ & $0.235$ & $0.595$ & $0.553$ & $\textbf{0.100}$ & $\textbf{0.208}$ & $0.168$ & $0.282$ & $0.292$ & $0.383$ & $0.143$ & $0.272$\\
		& 96 & $\underline{0.128}$ & $\underline{0.226}$ & $0.283$ & $0.341$ & $0.159$ & $0.267$ & $0.979$ & $0.716$ & $0.592$ & $0.552$ & $0.150$ & $0.253$ & $0.556$ & $0.563$ & $\textbf{0.120}$ & $\textbf{0.221}$ & $0.275$ & $0.373$ & $0.531$ & $0.535$ & $0.312$ & $0.417$\\
		\cmidrule(lr){2-24}
		& Avg & $\underline{0.092}$ & $\underline{0.193}$ & $0.165$ & $0.256$ & $0.109$ & $0.214$ & $0.438$ & $0.422$ & $0.329$ & $0.394$ & $0.109$ & $0.216$ & $0.418$ & $0.466$ & $\textbf{0.088}$ & $\textbf{0.192}$ & $0.166$ & $0.262$ & $0.264$ & $0.346$ & $0.196$ & $0.318$\\
		\midrule
		\multirow{5}{*}{\rotatebox{90}{PEMS08}} & 12 & $\textbf{0.076}$ & $\textbf{0.178}$ & $0.100$ & $0.208$ & $\underline{0.081}$ & $\underline{0.183}$ & $0.109$ & $0.222$ & $0.153$ & $0.259$ & $0.158$ & $0.192$ & $0.295$ & $0.391$ & $0.146$ & $0.198$ & $0.096$ & $0.206$ & $0.105$ & $0.219$ & $0.141$ & $0.262$\\
		& 24 & $0.124$ & $0.223$ & $0.171$ & $0.279$ & $\underline{0.118}$ & $\underline{0.222}$ & $0.205$ & $0.306$ & $0.238$ & $0.357$ & $\textbf{0.112}$ & $\textbf{0.219}$ & $0.345$ & $0.419$ & $0.171$ & $0.221$ & $0.144$ & $0.256$ & $0.179$ & $0.291$ & $0.243$ & $0.350$\\
		& 48 & $\textbf{0.172}$ & $\underline{0.216}$ & $0.317$ & $0.380$ & $\underline{0.202}$ & $0.292$ & $0.493$ & $0.484$ & $0.473$ & $0.515$ & $0.231$ & $\textbf{0.198}$ & $0.503$ & $0.495$ & $0.220$ & $0.270$ & $0.253$ & $0.351$ & $0.343$ & $0.405$ & $0.439$ & $0.476$\\
		& 96 & $\textbf{0.278}$ & $\textbf{0.296}$ & $0.468$ & $0.459$ & $0.395$ & $0.415$ & $0.582$ & $0.548$ & $0.748$ & $0.646$ & $\underline{0.291}$ & $0.338$ & $1.268$ & $0.857$ & $0.285$ & $\underline{0.303}$ & $0.482$ & $0.492$ & $0.720$ & $0.592$ & $0.886$ & $0.684$\\
		\cmidrule(lr){2-24}
		& Avg & $\textbf{0.162}$ & $\textbf{0.228}$ & $0.264$ & $0.332$ & $0.199$ & $0.278$ & $0.347$ & $0.390$ & $0.403$ & $0.444$ & $\underline{0.198}$ & $\underline{0.236}$ & $0.603$ & $0.541$ & $0.206$ & $0.248$ & $0.244$ & $0.326$ & $0.337$ & $0.377$ & $0.427$ & $0.443$\\
		\midrule
		\multicolumn{2}{c|}{\textbf{1st Count}} & $\textbf{11}$ & $\textbf{9}$ & $0$ & $0$ & $0$ & $0$ & $0$ & $0$ & $0$ & $0$ & $1$ & $4$ & $0$ & $0$ & $4$ & $3$ & $0$ & $0$ & $0$ & $0$ & $0$ & $0$ \\
		\bottomrule
	\end{tabular}
}
\end{table*}

\subsection{Main Results}
The main results for long-term and short-term forecasting are presented in Table \ref{compl} and Table \ref{comps}, in which the lower MSE and MAE values indicates superior forecasting performance. The proposed DMSC framework demonstrates consistent SOTA performance across all 13 benchmarks, achieving the lowest MAE and MSE in most experimental settings. These results confirm the robust effectiveness and generalizability of DMSC for both long-term and short-term forecasting tasks. Notably, multi-scale architectures like TimesNet and PatchMLP exhibit performance instability, which we attribute to their inflexible fusion mechanisms that fails to fully leverage the captured multi-scale features. The competitive performance of TimeXer and iTransformer underscores the effectiveness of patch-wise and variate-wise strategy for information representation, but their fixed decomposition schemes inherently constrain the model's potential. Furthermore, compared to iTransformer and Dlinear, DMSC achieves considerable improvement in data with more variate(Traffic, PEMS), highlighting the necessity of jointly dependency modeling. Collectively, the above observations validate the efficacy of DMSC in addressing the core challenges of time series forecasting.

\subsection{Model Analysis}

\subsubsection{Ablation Study.}
To rigorously evaluate the contribution and effectiveness of each component within DMSC, we conduct systematic ablation experiments on Electricity, Weather, Traffic and Solar datasets. We first ablate each of the three core blocks individually. As listed in Table \ref{ablation}, first three results highlight the complementary synergy of the three blocks. We then perform further ablation analyses for individual components, as detailed in Table \ref{ablation}. For EMPD, \textcircled{\scriptsize{1}} employs predefined static patch size for exponentially decaying. For TIB, \textcircled{\scriptsize{2}} retains only intra-patch feature extraction without jointly modeling the three dependency representations, while \textcircled{\scriptsize{3}} removes dynamic fusion mechanism $\mathbf{F}^l_{\text{fused}}$ in TIB. For ASR-MoE, \textcircled{\scriptsize{4}} replace the prediction heads with a simple summation of multiple linear layers, \textcircled{\scriptsize{5}} and \textcircled{\scriptsize{6}} eliminate global experts and local experts respectively, \textcolor{black}{and \textcircled{\scriptsize{7}} remove historical weighting memory $\mathbf{w}_{hist}$.} Based on these ablation studies, we have the following observations.

\begin{table}[htbp]
	\centering
	\setlength{\tabcolsep}{2pt}
	\caption{Ablation results. First three results were achieved by ablating each of three core blocks individually. Then \textcircled{\scriptsize{1}} - \textcircled{\scriptsize{7}} are further ablation analyses for each block. All results are averaged from prediction length \{96, 192, 336, 720\}.}
	\label{ablation}
    \resizebox{\textwidth}{!}{
	\begin{tabular}{cc|cc|cc|cc|cc|cc|cc|cc|cc|cc|cc|cc}
		\toprule
		\multicolumn{2}{c|}{\textbf{Design}} & \multicolumn{2}{c}{\textbf{DMSC(Ours)}} & \multicolumn{2}{c}{\textbf{w/o EMPD}} & \multicolumn{2}{c}{\textbf{w/o TIB}} & \multicolumn{2}{c}{\textbf{w/o MoE}} & \multicolumn{2}{c}{\textcircled{\scriptsize{1}}\textbf{static}} & \multicolumn{2}{c}{\textcircled{\scriptsize{2}}\textbf{only $\mathbf{F}_{intra}$}} & \multicolumn{2}{c}{\textcircled{\scriptsize{3}}\textbf{w/o $\mathbf{F}^l_{{fused}}$}} & \multicolumn{2}{c}{\textcircled{\scriptsize{4}}\textbf{Agg.}} &
		\multicolumn{2}{c}{\textcircled{\scriptsize{5}}\textbf{w/o $\mathcal{E}^g$}} & \multicolumn{2}{c}{\textcircled{\scriptsize{6}}\textbf{w/o $\mathcal{E}^l$}} &
		\multicolumn{2}{c}{\textcolor{black}{\textcircled{\scriptsize{7}}\textbf{w/o $\mathbf{w}_{hist}$}}}\\
		\cmidrule(lr){3-4} \cmidrule(lr){5-6} \cmidrule(lr){7-8} \cmidrule(lr){9-10} \cmidrule(lr){11-12} \cmidrule(lr){13-14} \cmidrule(lr){15-16} \cmidrule(lr){17-18} \cmidrule(lr){19-20} \cmidrule(lr){21-22} \cmidrule(lr){23-24}
	 	\multicolumn{2}{c|}{\textbf{Metric}} & MSE & MAE & MSE & MAE & MSE & MAE & MSE & MAE & MSE & MAE & MSE & MAE & MSE & MAE & MSE & MAE & MSE & MAE & MSE & MAE & \textcolor{black}{MSE} & \textcolor{black}{MAE}\\
		\midrule
		\multicolumn{2}{c|}{Electricity} & $\textbf{0.170}$ & $\textbf{0.258}$ & $0.181$ & $0.270$ & $0.185$ & $0.273$ & $0.196$ & $0.285$ & $0.179$ & $0.271$ & $0.187$ & $0.276$ & $0.186$ & $0.277$ & $0.194$ & $0.281$ & $0.191$ & $0.280$ & $0.199$ & $0.285$ & \textcolor{black}{$0.189$} & \textcolor{black}{$0.278$}\\
		\midrule
		\multicolumn{2}{c|}{Weather} & $\textbf{0.233}$ & $\textbf{0.269}$ & $0.248$ & $0.275$ & $0.255$ & $0.278$ & $0.251$ & $0.280$ & $0.245$ & $0.272$ & $0.249$ & $0.275$ & $0.247$ & $0.273$ & $0.254$ & $0.281$ & $0.250$ & $0.279$ & $0.243$ & $0.271$ & \textcolor{black}{$0.247$} & \textcolor{black}{$0.275$}\\
		\midrule
		\multicolumn{2}{c|}{Traffic} & $\textbf{0.407}$ & $\textbf{0.274}$ & $0.503$ & $0.323$ & $0.497$ & $0.321$ & $0.503$ & $0.320$ & $0.478$ & $0.313$ & $0.489$ & $0.315$ & $0.483$ & $0.314$ & $0.510$ & $0.312$ & $0.504$ & $0.324$ & $0.488$ & $0.326$ & \textcolor{black}{$0.483$} & \textcolor{black}{$0.301$}\\
		\midrule
		\multicolumn{2}{c|}{Solar} & $\textbf{0.213}$ & $\textbf{0.258}$ & $0.248$ & $0.278$ & $0.264$ & $0.293$& $0.271$ & $0.305$ & $0.243$ & $0.276$ & $0.275$ & $0.298$ & $0.277$ & $0.299$ & $0.278$ & $0.299$ & $0.265$ & $0.293$ & $0.275$ & $0.303$ & \textcolor{black}{$0.255$} & \textcolor{black}{$0.284$}\\
		\bottomrule
	\end{tabular}
}
\end{table}

For EMPD, replacing the input-adaptive decomposition with either predefined static patches or raw data embedding leads to performance degradation. This demonstrates the necessity of adaptive patch-wise decomposition and its lightweight network for dynamic granularity adjustment. For TIB, replacing joint modeling with standard convolution layers or only intra-patch features resulted in performance drops, validating the necessity of comprehensive joint modeling of triadic dependencies. Similarly, ablation of dynamic fusion mechanism in TIB confirms its effectiveness in fusing heterogeneous dependencies, as static handling undermines holistic cross-scale dependency capture. For ASR-MoE, replacing the adaptive prediction mechanism with a single prediction head or summed multiple linear projections leads to substantial performance declines. This indicates that single heads inadequately utilize multi-scale features, while simplistic aggregation overlooks scale-specific dependency importance. Meanwhile, removing either global or local experts degrades performance, demonstrating the necessity of maintaining distinct experts for different temporal patterns. \textcolor{black}{Moreover, removing the historical weighting memory causes a noticeable drop in MSE, confirming that $\mathbf{w}_{hist}$ provides a useful scale prior that stabilizes the fusion.} The elaborate design of ASR-MoE effectively harnesses diverse temporal dynamics through collective specialized expert utilization. Collectively, these ablation studies confirm that dynamic and multi-scale modeling is central to DMSC. All of three blocks and the proposed progressive cascade architecture together enable robust multiscale modeling and hierarchical feature learning, with consistent performance drops from ablating any component validate architectural integrity of DMSC.

\textcolor{black}{To complement the performance ablation, we report the FLOPs, train time, and inference time of some variants in Table \ref{ablation_efficiency}. The table shows that each module introduces only a modest increase in computation. Removing EMPD’s dynamic adaptation or simplifying TIB leads to slight reductions in FLOPs and latency, but these savings are small compared to the corresponding performance drops. Replacing ASR - MoE with a simple linear fusion yields the largest efficiency gain, yet the forecasting accuracy degrades substantially, confirming the value of sparse expert routing. Importantly, the full DMSC maintains competitive inference speed while achieving the best accuracy, demonstrating that the additional complexity of each component is well justified.}
\begin{table}[htbp]
    \centering
    \caption{Computational efficiency of ablation variants on Weather dataset with 96-look-back length and 96-prediction length. All metrics are measured per batch (batch size 128).}
    \label{ablation_efficiency}
    \resizebox{0.65\textwidth}{!}{
    \begin{tabular}{l|ccc|c}
        \toprule
        {Design} & FLOPs(G) & Inference Time(ms/batch) & GPU Memory(MB) & MSE \\
        \hline
        Full DMSC & 3.4   & 1.6   & 304   & 0.160 \\
        \hline
        w/o EMPD         & 3.2   & 2.4   & 296   & 0.164 \\
        w/o TIB          & 2.8   & 2.1   & 307   & 0.168 \\
        w/o MoE          & 2.6   & 1.3   & 325   & 0.171 \\
        static patch design          & 4.1   & 2.8   & 398   & 0.167 \\
        w/o $\mathbf{F}^l_{\text{fused}}$ & 3.5   & 2.7   & 307   & 0.168 \\
        w/o $\mathbf{w}_{hist}$           & 3.4   & 1.6   & 306   & 0.169 \\
        \bottomrule
    \end{tabular}
    }
\end{table}

\subsubsection{ASR-MoE Expert Specialization Analysis}
To verify that the global and local experts in ASR-MoE indeed develop distinct and meaningful specializations, we conduct two sets of analyses: (i) a quantitative study of expert utilization and load balancing, and (ii) a qualitative case study that reveals how experts respond to different temporal regimes.

\begin{figure}[!t]
    \centering
    \begin{subfigure}[b]{0.46\textwidth}
        \centering
        \includegraphics[width=\textwidth]{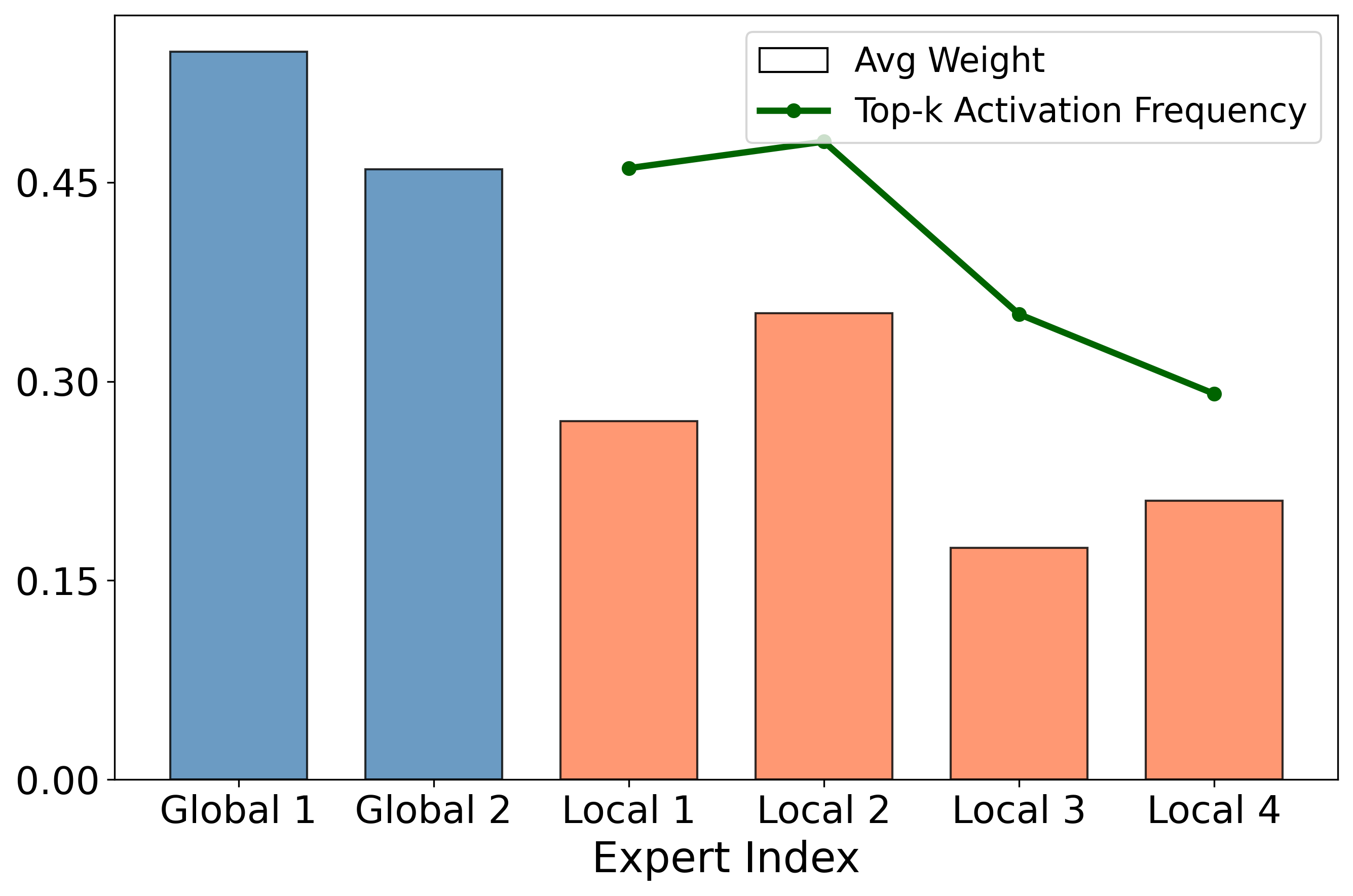}
        \caption{On Electricity dataset.}
        \label{expertload1}
    \end{subfigure}
    \hfill
    \begin{subfigure}[b]{0.46\textwidth}
        \centering
        \includegraphics[width=\textwidth]{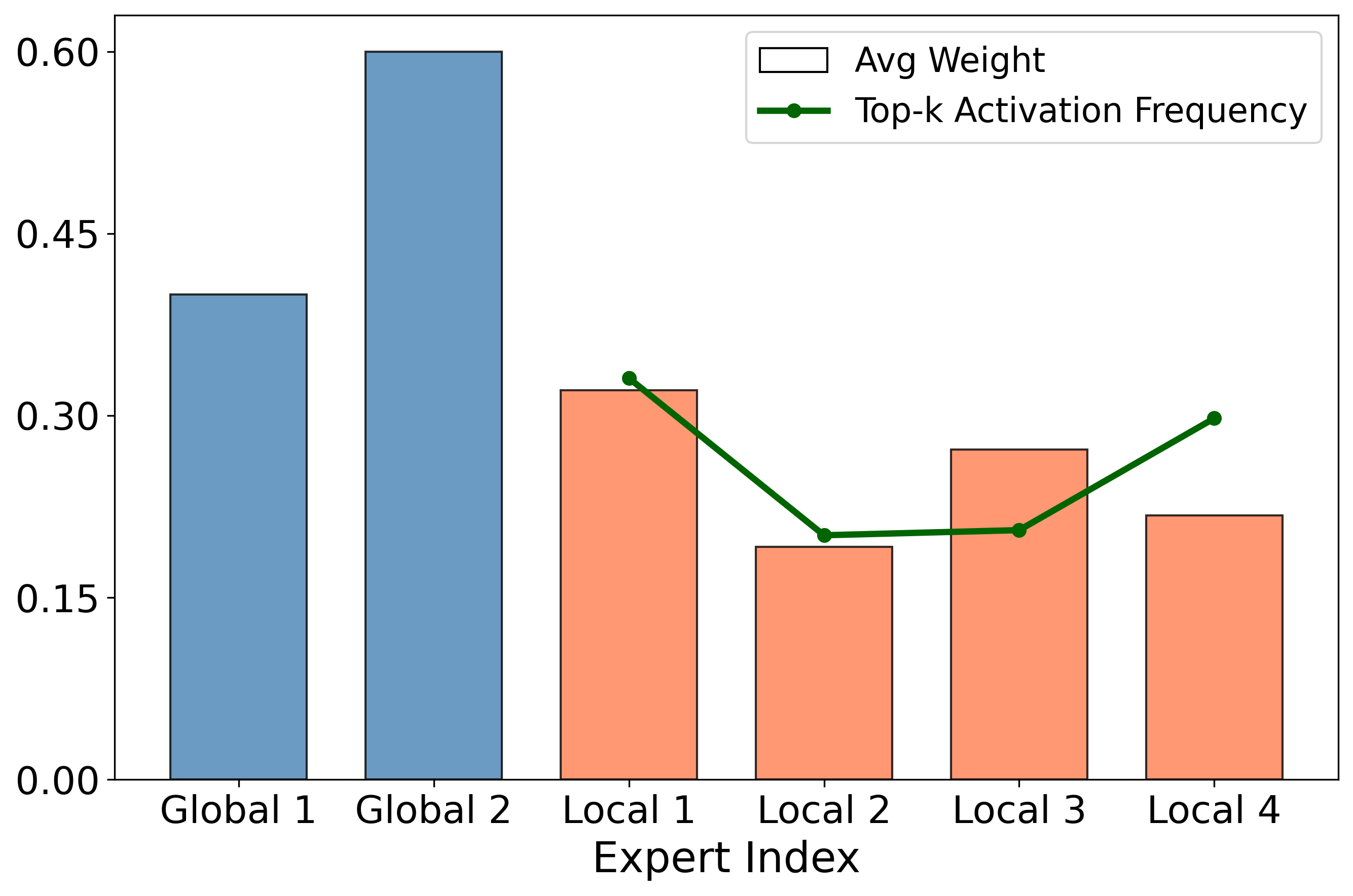}
        \caption{On Weather dataset.}
        \label{expertload2}
    \end{subfigure}
    \caption{Analysis of expert utilization and load balancing. Left is on Electricity dataset, and right is on Weather dataset.}
    \label{expertload}
\end{figure}

\textcolor{black}{\textbf{Expert Utilization and Load Balancing.} To examine whether all experts are actively engaged during inference or whether the routing collapses to a small subset, we compute the average activation frequency and the average activation weight over all test samples for each expert. As shown in Figure \ref{expertload}, both global experts and all local experts are consistently utilized on both datasets. The average weights are reasonably balanced across experts, indicating that the auxiliary entropy loss effectively promotes load sharing.}

\begin{figure}[!t]
        \centering
        \begin{subfigure}[b]{0.49\textwidth}
            \centering
            \includegraphics[width=\textwidth]{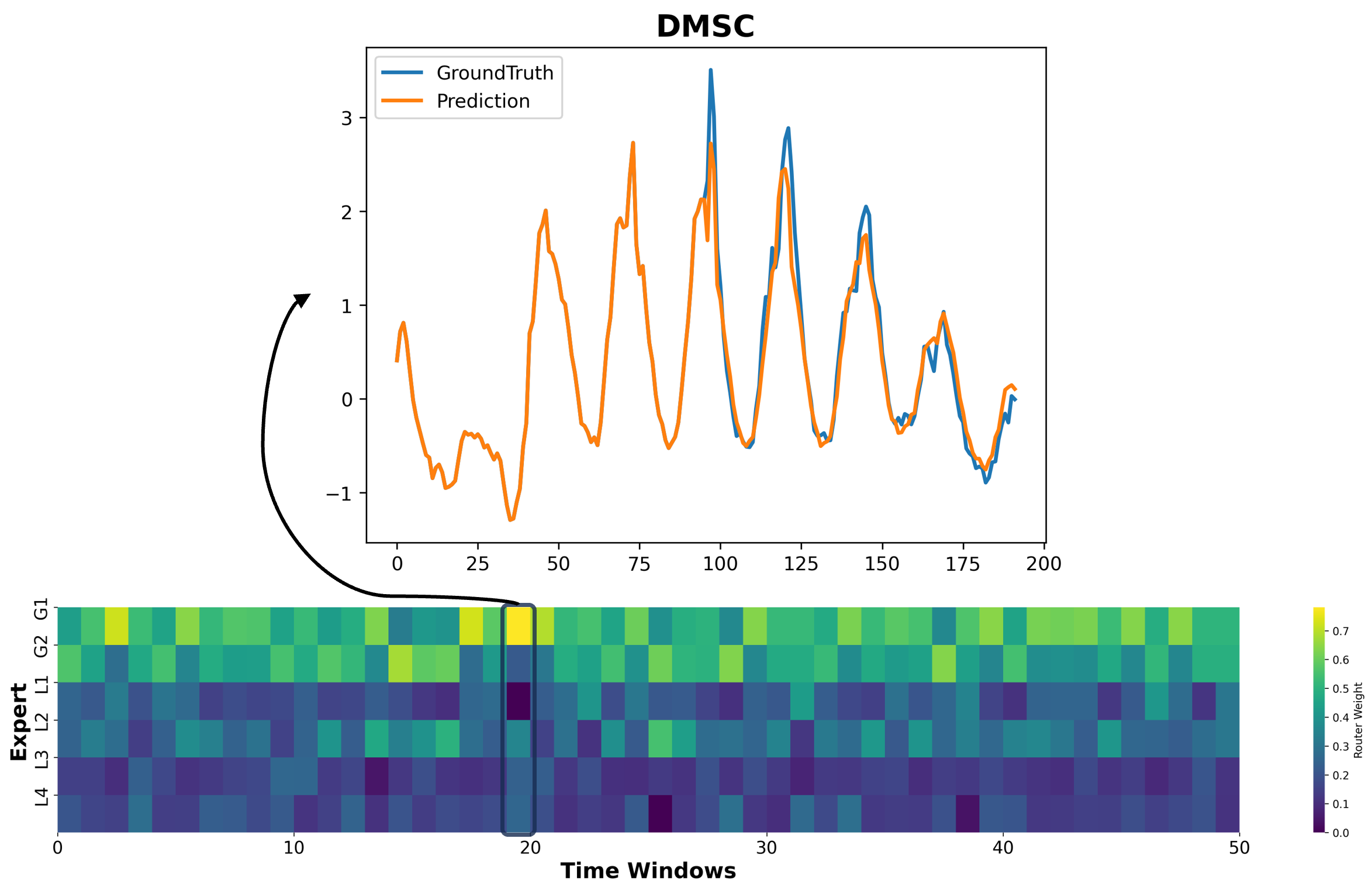}
            \caption{Long-term patterns.}
            \label{expertspecial1}
        \end{subfigure}
        \hfill
        \begin{subfigure}[b]{0.49\textwidth}
            \centering
            \includegraphics[width=\textwidth]{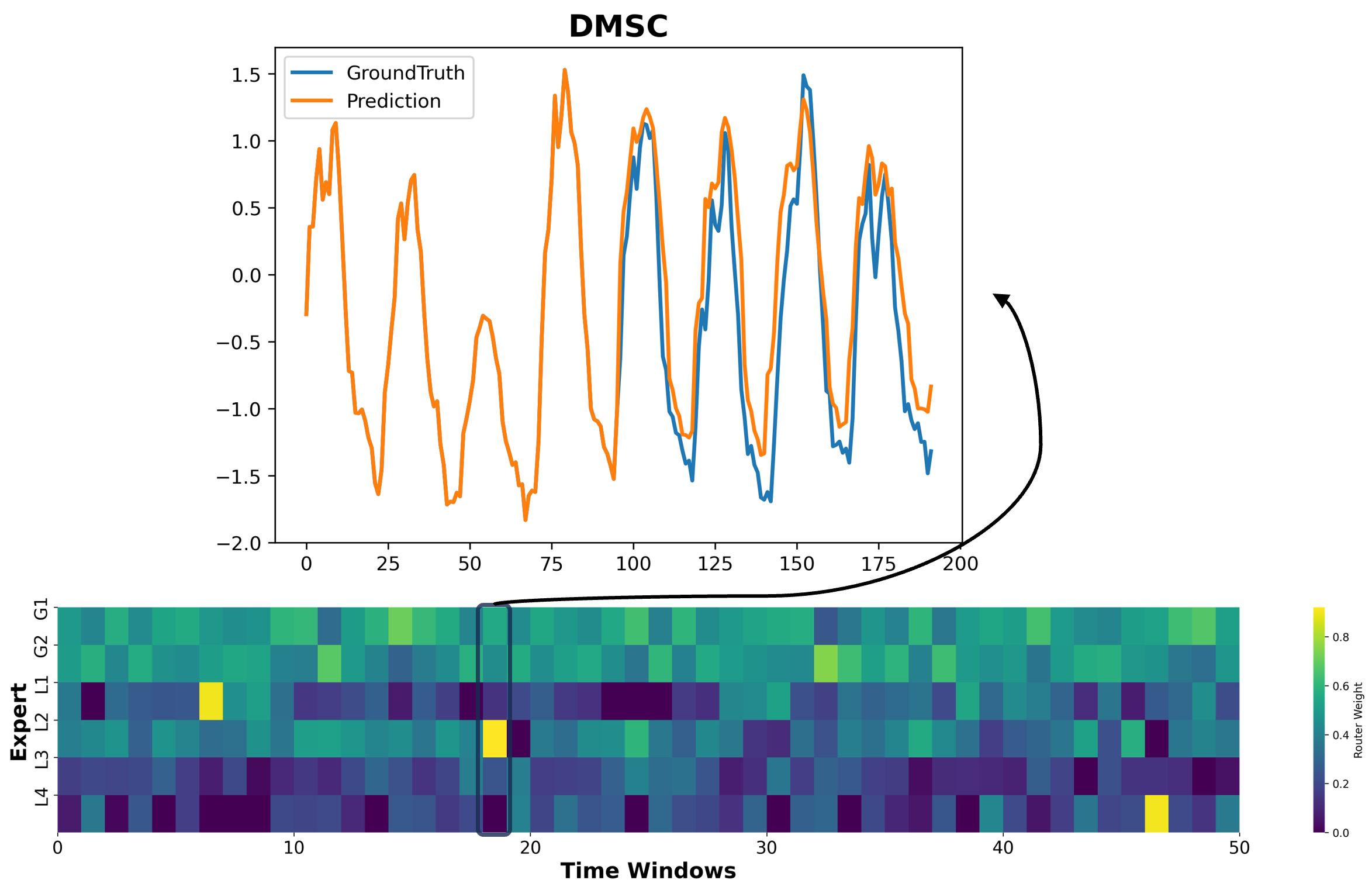}
            \caption{Short-term patterns.}
            \label{expertspecial2}
        \end{subfigure}
        \caption{Case study of expert specialization. Left is dominated by long - term patterns, and right is dominated on short term patterns.}
        \label{expertspecial}
    \end{figure}

\begin{figure}[!t]
  \centering
    \begin{subfigure}{0.47\textwidth}
        \includegraphics[width=\textwidth]{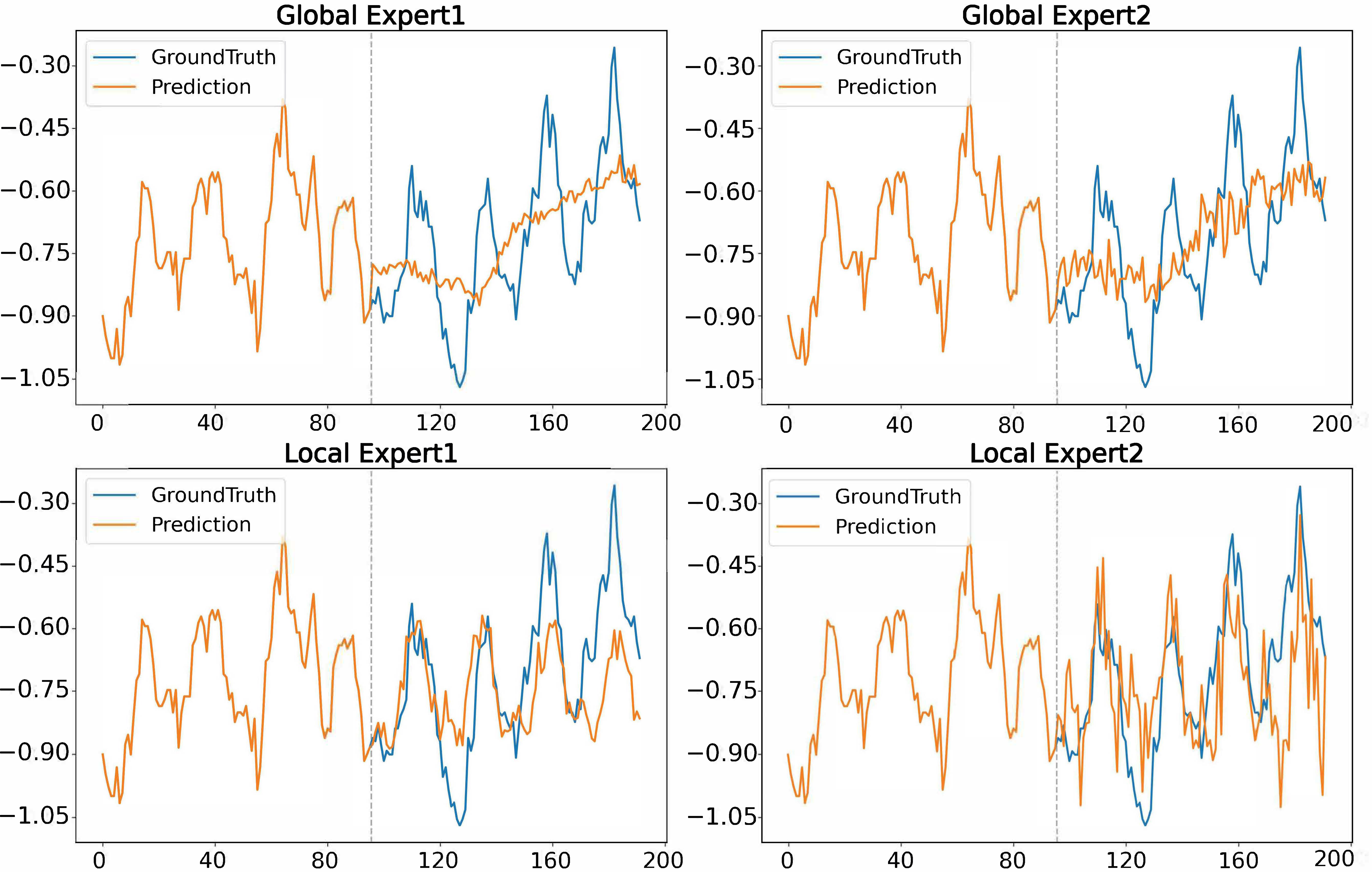}
        \caption{Visualization of different experts.}
        \label{MoE}
    \end{subfigure}
    \hspace{-0.1cm}
    \begin{subfigure}{0.4\textwidth}
        \includegraphics[width=\textwidth]{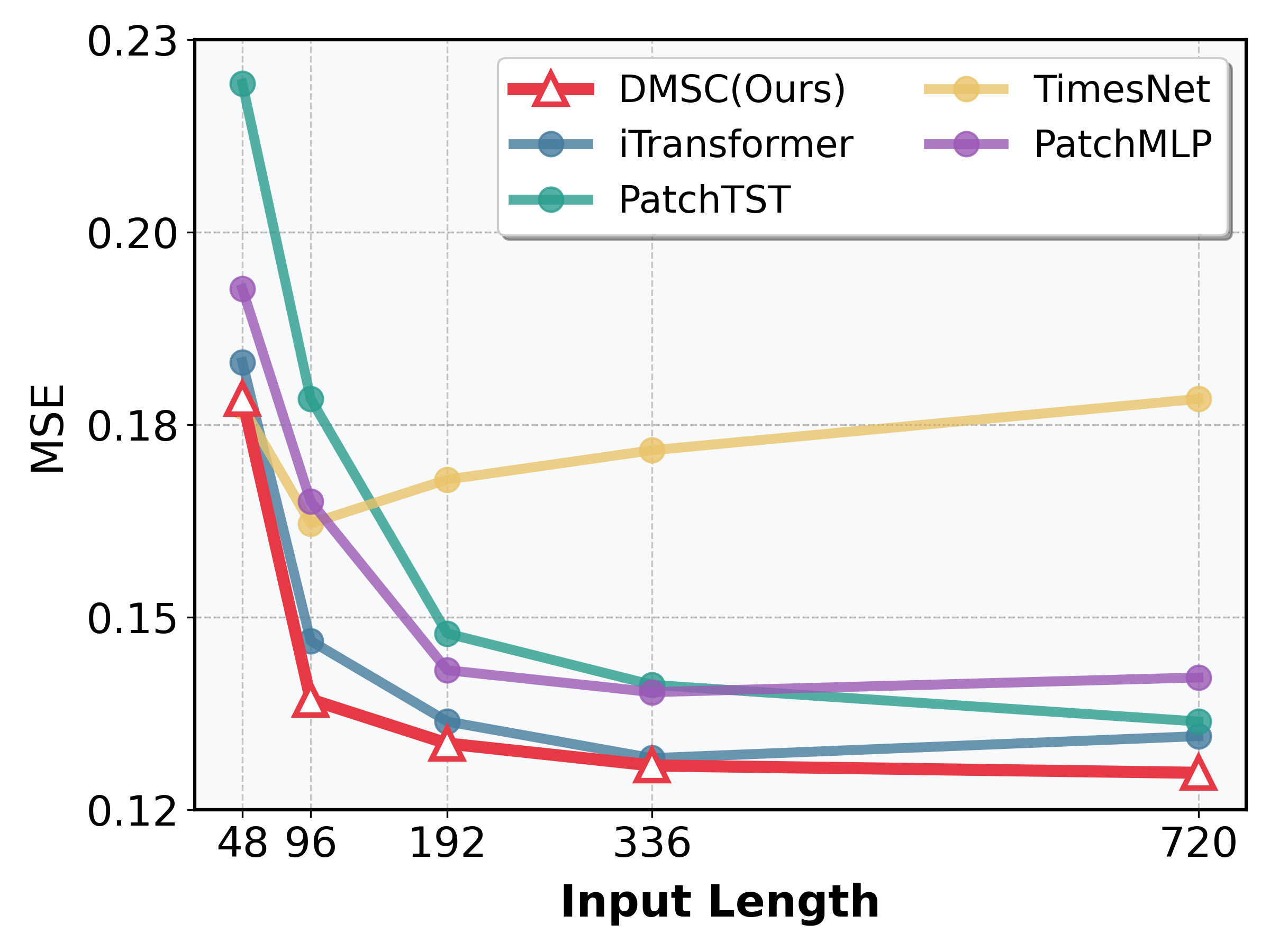}
        \caption{Forecasting results on varying look-back length.}
        \label{input_length}
    \end{subfigure}
    \caption{(a) Visualization of different experts on ETTh1, the look-back and prediction length are set to 96. (b) Forecasting results with varying look-back length on Electricity dataset. Look-back lengths are set to \{48, 96, 192, 336, 720\}, and prediction length is fixed to 96.}
    \label{model_analysis1}
\end{figure}

\textcolor{black}{\textbf{Expert Activation Patterns and Case Study.} To gain a more insight into the functional roles of different experts, we design a case study with two distinct time patterns in Figure \ref{expertspecial}, and visualize the activation patterns across different temporal experts in Figure \ref{MoE}. The heatmaps in figure \ref{expertspecial} reveal strikingly different activation patterns. In the long-term dominated segment, the global experts maintain stable, moderate activations throughout the window, while the local experts remain mostly inactive with low weights. In contrast, in the short-term dominated segment, several local experts become strongly and selectively activated, while the global experts continue to provide a background signal but with reduced relative dominance. Meanwhile, Figure \ref{MoE} shows that distinct specialization profiles emerge: global experts exhibit heightened sensitivity to persistent trends and long-periodic patterns, whereas local experts demonstrate acute responsiveness to short-periodic patterns and high-frequency fluctuations. This dynamic switching directly supports the claim that global experts specialize in persistent long-term patterns and local experts in transient short-term variations.}

\textcolor{black}{Overall, these observed specialization demonstrate that the routing mechanism in ASR-MoE learns a functionally meaningful division of labor, rather than acting as a mere complex fusion layer.}

\subsubsection{Increasing Look-Back Length.}
Theoretically, longer input sequences provide richer historical information, which can enhance the model's ability to capture long-term trends and complex multi-scale dependencies. Our DMSC framework is specifically designed to leverage this characteristic, as its dynamic multi-scale coordination capability inherently facilitates adaptive modeling and effective information utilization across varying look-back lengths. So we evaluate DMSC with look-back lengths selected from \{48, 96, 192, 336, 720\} on Electricity dataset. As shown in Figure \ref{input_length}, DMSC consistently achieves improved performance with longer historical sequences while outperforming other baseline methods. This demonstrates its capacity to extract richer temporal representations from extended contexts and effectively integrate multi-scale temporal patterns.

\subsubsection{Comparison with Static Patch.}
\textcolor{black}{To further assess the benefit of dynamic scale adaptation, we compare EMPD with five manually selected, fixed patch sizes (\(P_{base} \in \{8, 16, 24, 32, 64\}\)). The experiments are conducted on Electricity, Weather, Traffic, and Solar dataset. For each fixed setting, the same multi‑layer cascade is used but with a constant patch base length. The results are listed in Table \ref{diff_patch}, which reveal two consistent patterns. First, no single fixed patch size performs optimally across all datasets: for instance, \(P_{base}=16\) achieves the best result on Weather, while \(P_{base}=32\) is the best on Traffic, indicating that the appropriate scale is highly dataset‑dependent. Second, the dynamic mechanism consistently matches or outperforms the best fixed setting on every dataset, without requiring any per - dataset manual tuning. This confirms that input - adaptive granularity adjustment provides practical gains in real - world scenarios where data characteristics vary.}

\begin{table}[htbp]
	\centering
	\caption{Futher analysis of patch decomposition scheme. We compare EMPD block with different static patch lengths \{8, 16, 24, 32, 64\} in four real-world datasets.}
	\label{diff_patch}
    \resizebox{0.8\textwidth}{!}{
	\begin{tabular}{cc|cc|cc|cc|cc|cc|cc}
		\toprule
		\multicolumn{2}{c|}{\textbf{Patch\_Length}} & \multicolumn{2}{c}{\textbf{Dynamic(ori)}} & \multicolumn{2}{c}{\textbf{8}} & \multicolumn{2}{c}{\textbf{16}} & \multicolumn{2}{c}{\textbf{24}} & \multicolumn{2}{c}{\textbf{32}} & \multicolumn{2}{c}{\textbf{64}} \\
		\cmidrule(lr){3-4} \cmidrule(lr){5-6} \cmidrule(lr){7-8} \cmidrule(lr){9-10} \cmidrule(lr){11-12} \cmidrule(lr){13-14} 
	 	\multicolumn{2}{c|}{\textbf{Metric}} & MSE & MAE & MSE & MAE & MSE & MAE & MSE & MAE & MSE & MAE & MSE & MAE \\
		\midrule
		\multicolumn{2}{c|}{Electricity} & \textbf{0.138 } & \textbf{0.223 } & 0.161  & 0.257  & 0.152  & 0.244  & 0.158  & 0.250  & 0.156  & 0.248  & 0.141  & 0.2300  \\
		\midrule
		\multicolumn{2}{c|}{Weather} & \textbf{0.160 } & \textbf{0.210 } & 0.168  & 0.216  & 0.166  & 0.212  & 0.169  & 0.217  & 0.167  & 0.214  & 0.166  & 0.215 \\
		\midrule
		\multicolumn{2}{c|}{Traffic} & \textbf{0.389 } & \textbf{0.259 } & 0.432  & 0.301  & 0.461  & 0.305  & 0.454  & 0.303  & 0.401  & 0.276 & 0.475  & 0.305 \\
		\midrule
		\multicolumn{2}{c|}{Solar} & \textbf{0.187 } & \textbf{0.231 } & 0.231  & 0.275  & 0.216  & 0.260  & 0.208  & 0.258  & 0.197  & 0.243  & 0.191  & 0.242  \\
		\bottomrule
	\end{tabular}
}
\end{table}

\subsubsection{Analysis of Learned Patch Scales.}\label{alps}
\textcolor{black}{To verify that the scale factor $\alpha$ learned by EMPD is genuinely input-adaptive rather than degenerating into a fixed value, we analyze its behavior under varying data characteristics and forecasting conditions.}

\begin{figure}[!t]
    \centering
    \begin{subfigure}[b]{0.46\textwidth}
        \centering
        \includegraphics[width=\textwidth]{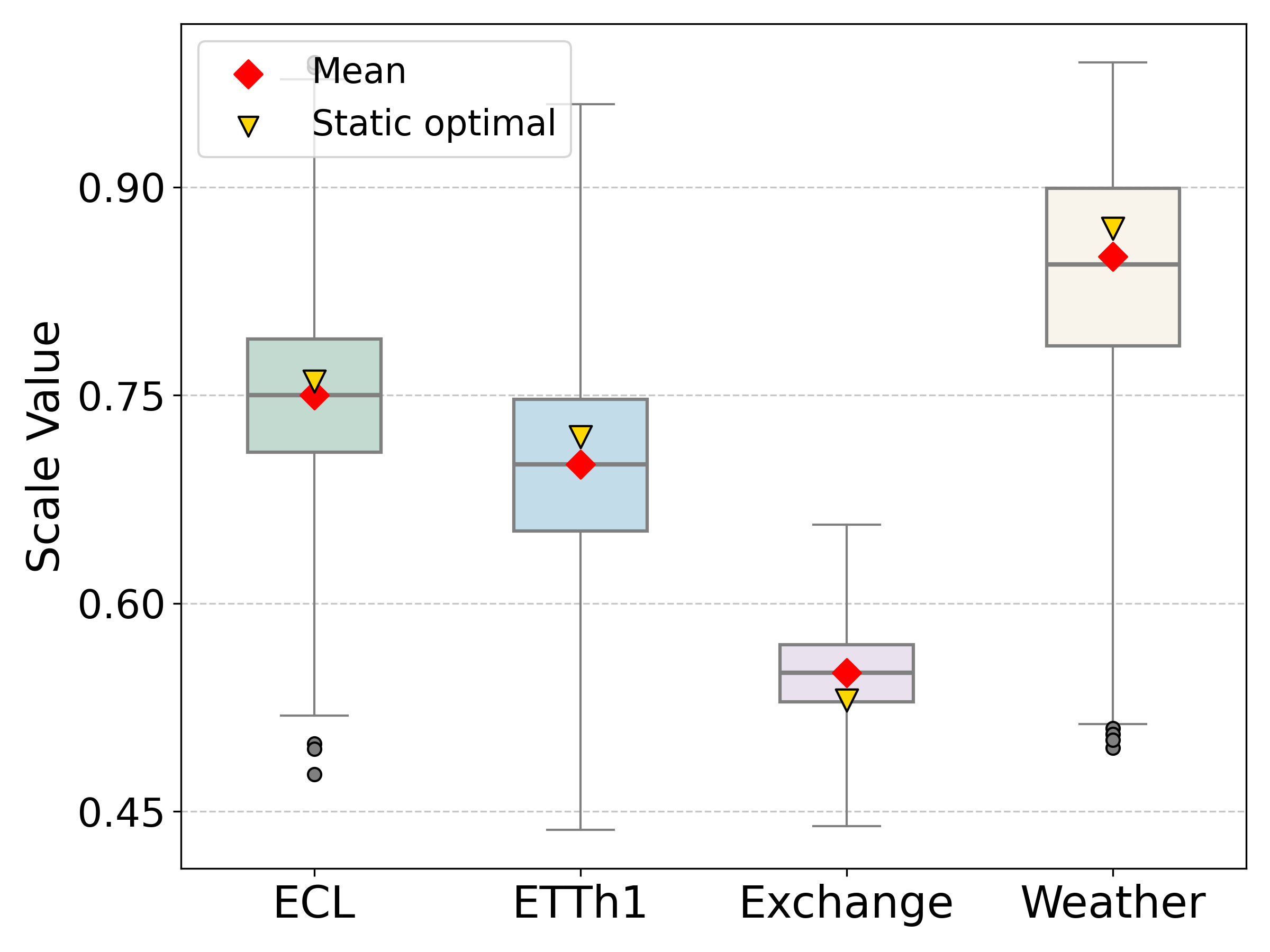}
        \caption{Cross - dataset distribution.}
        \label{numscales1}
    \end{subfigure}
    \hfill
    \begin{subfigure}[b]{0.46\textwidth}
        \centering
        \includegraphics[width=\textwidth]{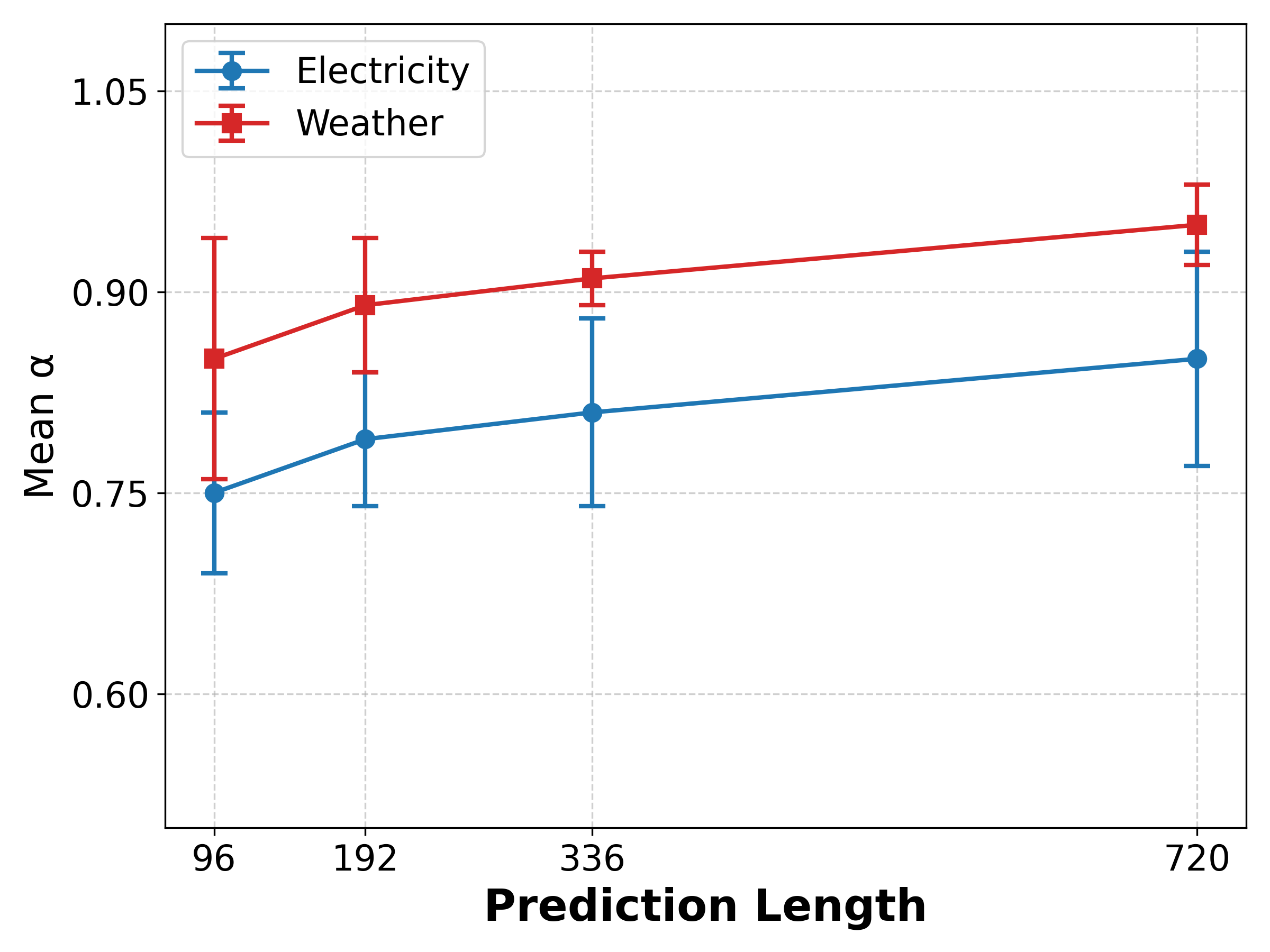}
        \caption{Effect of prediction horizon.}
        \label{numscales2}
    \end{subfigure}
    \caption{Analysis of Learned Patch Scales. Left is distribution of $\alpha$ across four datasets. Right is effect of prediction horizon.}
    \label{numscales}
\end{figure}

\textcolor{black}{\textbf{Cross - dataset distribution.} Figure \ref{numscales1} shows the distribution of $\alpha$ on four representative datasets (ECL, ETTh1, Weather, Exchange). The $\alpha$ values are collected from the test set after training. The boxplots reveal clear inter - dataset differences: high - frequency datasets such as Weather tend to yield larger $\alpha$ (finer patches), while datasets with longer dominant periodicities such as Exchange exhibit smaller $\alpha$ (coarser patches). The dynamic mean $\alpha$ (marked by red diamonds) is close to but not identical to the per - dataset optimal static $\alpha$ (yellow triangles) identified in the fixed - patch ablation, and the non - negligible box widths indicate that $\alpha$ continues to vary across different input windows within the same dataset. This demonstrates that the scale controller adapts the patch granularity to the statistical character of each dataset while still responding to local temporal variations.}

\textcolor{black}{\textbf{Effect of prediction horizon.} Figure \ref{numscales2} reports the mean $\alpha$ and its standard deviation across different prediction lengths ($96$, $192$, $336$, $720$) on Electricity and Weather dataset. A clear increasing trend is observed for both datasets: as the prediction horizon lengthens, the model consistently selects larger $\alpha$, i.e., finer patch granularities. This suggests that longer - range forecasting benefits from more detailed local representations, and the model is capable of adjusting its decomposition strategy accordingly. The small error bars further confirm that this adaptation is stable and systematic.}

\textcolor{black}{Collectively, these analyses confirm that the lightweight scale controller in EMPD learns a meaningful, input - dependent scale factor that varies across datasets, adapts within a dataset according to the prediction task, and never collapses to a single static value.}

\subsubsection{Efficiency Analysis.}\label{ea}
\begin{figure}[!t]
    \centering
    \begin{subfigure}[b]{0.47\textwidth}
        \centering
        \includegraphics[width=\textwidth]{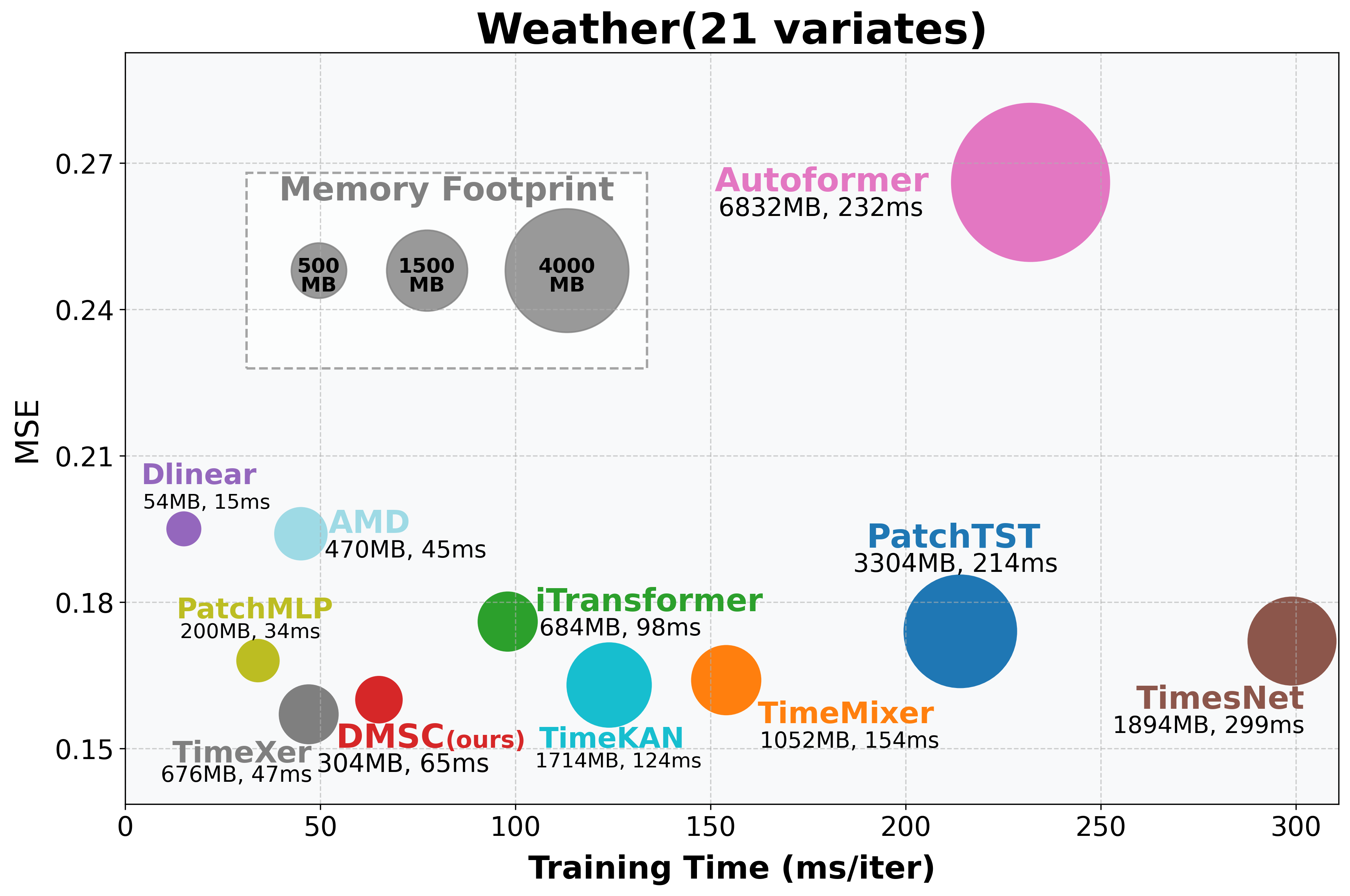}
        \caption{Efficiency analysis on Weather dataset.}
        \label{Efficiency}
    \end{subfigure}
    \hfill
    \begin{subfigure}[b]{0.47\textwidth}
        \centering
        \includegraphics[width=\textwidth]{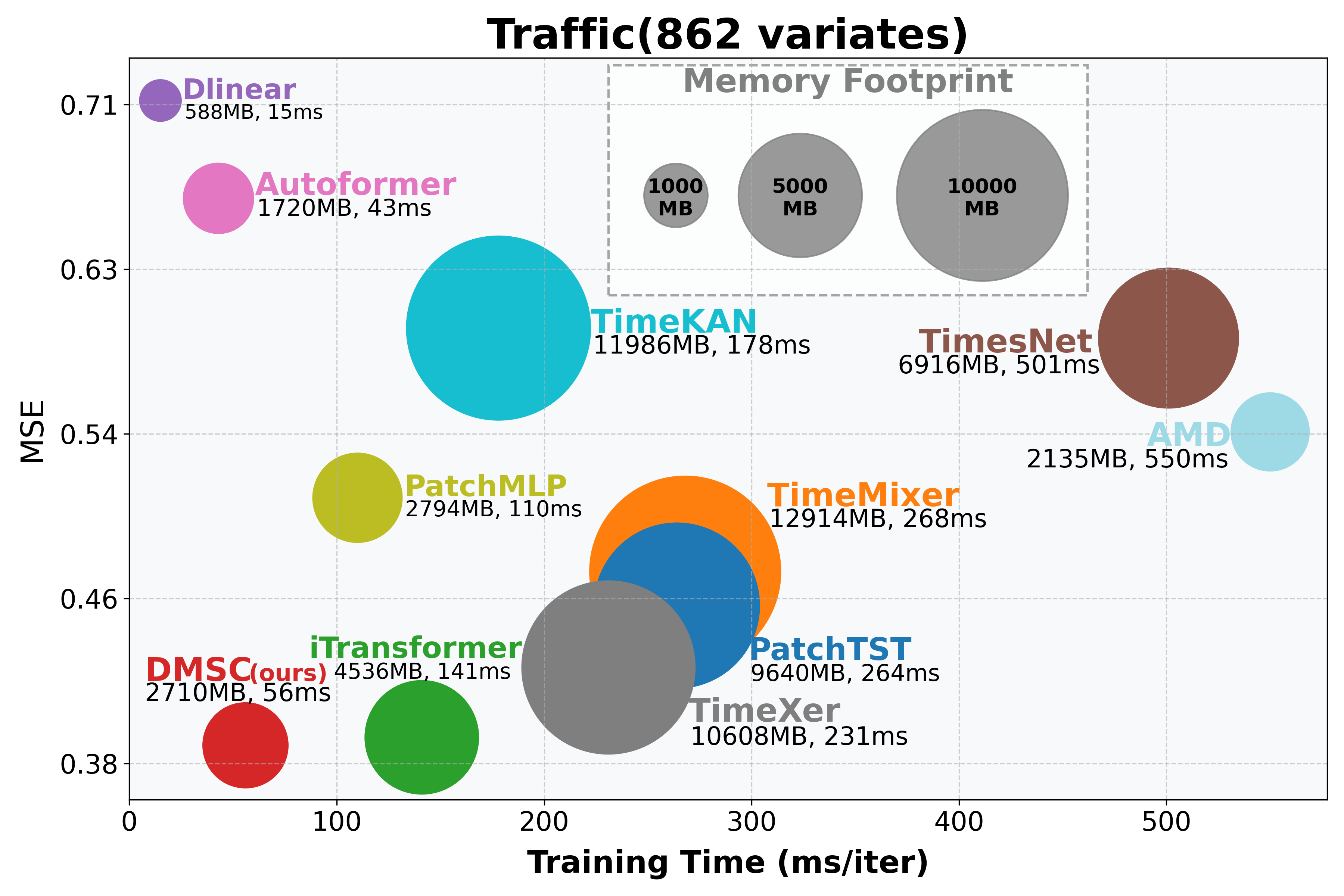}
        \caption{Efficiency analysis on Traffic dataset.}
        \label{Efficiency1}
    \end{subfigure}
    \caption{Model efficiency analysis under 96-look-back length and 96-prediction length. Left(a) is on Weather(21 variates) dataset, while right(b) is on Traffic(862 variates) dataset. Batch size is set to 128(a) and 16(b).}
    \label{Efficanaly}
\end{figure}

We rigorously evaluate the memory usage and training time of DMSC against other SOTA baselines on Weather(21 variates) and Traffic(862 variates) datasets. As demonstrated in Figure \ref{Efficiency} and Figure \ref{Efficiency1}, DMSC consistently achieves superior efficiency in both memory usage and training time while maintaining competitive forecasting accuracy. Crucially, DMSC demonstrates near-linear scalability to long sequences, as outperforming Transformer-based models with quadratic complexity and MLP-based approaches susceptible to parameter explosion. These efficiency gains are particularly pronounced when processing high-dimensional multivariate data (Traffic), where the cross-variable interaction in TIB avoids the computational overhead of exhaustive pairwise attention while preserving modeling capacity. DMSC thus establishes a balance of accuracy, latency, and memory usage for practical deployment scenarios.

\textcolor{black}{To complement the visual overview, we provide a detailed quantitative comparison on the Weather dataset. As shown in Table \ref{efficiency2}, DMSC maintains a moderate parameter count and low FLOPs while achieving the better performance among all compared methods. Its training and inference speeds are competitive with efficient CNN-based and Transformer-based baselines, while GPU memory consumption remains modest. These results confirm that DMSC establishes a favorable accuracy-efficiency trade-off suitable for practical deployment.}

\begin{table}[htbp]
        \centering
        \caption{Computational efficiency comparison on the Weather dataset with under 96-look-back length and 96-prediction length, batch size is set to 128.}
        \resizebox{0.7\textwidth}{!}{
            \begin{tabular}{c|ccccc|c} 
                \toprule
                \multirow{2}{*}{Methods}      & Params & FLOPs & Train Time & Inference Time & GPU Memory & MSE \\
                             & (M)      & (M)   & (ms/iter)   & (ms/batch)     & (MB)      &     \\ \hline
                DMSC (Ours) & 3.4   & 106.34 & 65    & 1.6   & 304   & 0.160 \\ \hline
                iTransformer & 4.8   & 120.78 & 98    & 1.8   & 684   & 0.176 \\
                PatchTST      & 0.5   & 103.61 & 214   & 1.9   & 3304  & 0.174 \\
                Autoformer     &  9.6   & 158.56 & 232   & 13.1  & 6832  & 0.301 \\
                TimesNet      &  1.8   & 201.32 & 299   & 10.9  & 1894  & 0.172 \\
                TimeMixer    & 0.14  & 107.14 & 154   & 5.7   & 1052  & 0.164 \\ \bottomrule    
            \end{tabular}
        }
    
        \label{efficiency2}
\end{table}

\textcolor{black}{Furthermore, we analyze the computational complexity of DMSC and compare it with Transformer-based approaches. A standard Transformer with full self-attention incurs $O(L^2 \cdot d)$ complexity, where $L$ is the sequence length and $d$ is the feature dimension. This quadratic dependence limits its scalability to long sequences. In DMSC, the TIB block uses depthwise separable convolutions ($O(N_l \cdot P_l \cdot C \cdot D)$) and dilated convolutions ($O(N_l \cdot C \cdot D)$) for intra- and inter-patch modeling respectively, both scaling linearly with the number of patches $N_l$. The cross-variable gating involves simple MLP operations, also linear in $C$ and $D$. For ASR-MoE, only the top-$K$ local experts (typically $K=2$) are activated per sample, resulting in $O(K \cdot C \cdot D)$ complexity for the sparse routing component. The overall complexity of DMSC is therefore linear or near-linear with respect to the sequence length, making it inherently more scalable than Transformer-based architectures for long-sequence time series forecasting. This theoretical advantage is reflected in the empirical efficiency results reported in the table above, where DMSC demonstrates competitive FLOPs and inference time while maintaining SOTA accuracy.}

\subsubsection{Statistical Significance Analysis}
\textcolor{black}{To verify the reliability of the reported performance improvements, we conduct statistical significance tests. We run DMSC together with four strong baselines (PatchTST, iTransformer, TimeMixer, TimesNet) under five different random seeds on Electricity, ETTh1, Solar, and Weather dataset. For each configuration (dataset × model × prediction length), we obtain five independent MSE values. All models use the same look-back window ($L=96$) and the four standard prediction horizons ($H \in \{96, 192, 336, 720\}$). Each run uses a distinct random seed for parameter initialization and data shuffling, while keeping all hyperparameters fixed to their default or best - validated values. The resulting MSE values form paired samples for each (dataset, horizon) pair. For each baseline and each experimental setting, we perform a paired two-tailed $t$-test between the five MSEs and MAEs of DMSC and the five of the baseline. The null hypothesis is that the mean difference is zero. Statistical significance is indicated by superscripts:$^\dagger$ denotes $p < 0.05$ and$^\ddagger$ denotes $p < 0.01$, meaning that DMSC achieves a statistically significant improvement over the competing method in that setting.}

\textcolor{black}{The significance annotations in Table \ref{tab:stat_sig} show that DMSC significantly outperforms the baselines in the majority of the tested conditions. On the four selected datasets, over 67.5\% of the comparisons yield $p<0.05$, and 56.8\% yield $p<0.01$. The improvements are particularly pronounced on the Electricity and Weather datasets, where the multivariate dependencies are richer. Even in the few cases where the difference does not reach the $p<0.05$ threshold, DMSC consistently achieves superior forecasting performance. These results provide solid statistical evidence that the performance gains of DMSC are robust and not due to random variation.}

\begin{table}[htbp]
    \centering
    \caption{Long-term forecasting results with statistical significance on representative datasets. MSE and MAE are averaged over 5 random seeds (mean $\pm$ std). $^\dagger$ and$^\ddagger$ indicate that DMSC significantly outperforms the corresponding baseline at $p<0.05$ and $p<0.01$, respectively (paired two-tailed $t$-test). The best result in each setting is in \textbf{bold}.}
    \label{tab:stat_sig}
    \resizebox{\textwidth}{!}{%
    \begin{tabular}{c|l|cc|cc|cc|cc|cc}
        \toprule
        \multirow{2}{*}{\textbf{Dataset}} & \multirow{2}{*}{\centering\textbf{Model}} & \multicolumn{2}{c}{\textbf{96}} & \multicolumn{2}{c}{\textbf{192}} & \multicolumn{2}{c}{\textbf{336}} & \multicolumn{2}{c}{\textbf{720}} & \multicolumn{2}{c}{\textbf{Avg}} \\
        \cmidrule(lr){3-4} \cmidrule(lr){5-6} \cmidrule(lr){7-8} \cmidrule(lr){9-10} \cmidrule(lr){11-12}
         & & MSE & MAE & MSE & MAE & MSE & MAE & MSE & MAE & MSE & MAE \\
        \midrule
        \multirow{5}{*}{Electricity} & DMSC (Ours) & \textbf{0.138 $\pm$ 0.001} & \textbf{0.223 $\pm$ 0.001} & \textbf{0.158 $\pm$ 0.003} & \textbf{0.258 $\pm$ 0.002} & \textbf{0.168 $\pm$ 0.004} & \textbf{0.253 $\pm$ 0.003} & \textbf{0.207 $\pm$ 0.010} & \textbf{0.296 $\pm$ 0.008} & \textbf{0.168} & \textbf{0.257} \\
        & PatchTST    & 0.189 $\pm$ 0.002$^\ddagger$& 0.281 $\pm$ 0.001$^\ddagger$& 0.195 $\pm$ 0.001$^\ddagger$& 0.286 $\pm$ 0.001$^\ddagger$& 0.211 $\pm$ 0.002$^\ddagger$& 0.302 $\pm$ 0.002$^\ddagger$& 0.253 $\pm$ 0.001$^\ddagger$& 0.334 $\pm$ 0.001$^\ddagger$& 0.212 & 0.301 \\
        & iTransformer & 0.165 $\pm$ 0.002$^\ddagger$& 0.268 $\pm$ 0.002$^\ddagger$& 0.183 $\pm$ 0.003$^\dagger$& 0.284 $\pm$ 0.003$^\dagger$& 0.202 $\pm$ 0.005$^\ddagger$& 0.302 $\pm$ 0.004$^\ddagger$& 0.244 $\pm$ 0.022 & 0.334 $\pm$ 0.015$^\dagger$& 0.199 & 0.297 \\
        & TimeMixer   & 0.148 $\pm$ 0.001$^\ddagger$& 0.240 $\pm$ 0.001$^\ddagger$& 0.164 $\pm$ 0.002$^\ddagger$& 0.254 $\pm$ 0.002$^\ddagger$& 0.180 $\pm$ 0.001$^\ddagger$& 0.272 $\pm$ 0.001$^\ddagger$& 0.218 $\pm$ 0.006$^\ddagger$& 0.305 $\pm$ 0.004$^\ddagger$& 0.177 & 0.268 \\
        & TimesNet    & 0.157 $\pm$ 0.001$^\ddagger$& 0.250 $\pm$ 0.002$^\ddagger$& 0.172 $\pm$ 0.001$^\ddagger$& 0.265 $\pm$ 0.002$^\ddagger$& 0.189 $\pm$ 0.002$^\ddagger$& 0.282 $\pm$ 0.003$^\ddagger$& 0.230 $\pm$ 0.003$^\dagger$& 0.318 $\pm$ 0.004$^\ddagger$& 0.187 & 0.279 \\
        \midrule
         \multirow{5}{*}{ETTh1} & DMSC (Ours) & \textbf{0.372 $\pm$ 0.006} & \textbf{0.396 $\pm$ 0.004} & \textbf{0.404 $\pm$ 0.005} & \textbf{0.421 $\pm$ 0.004} & \textbf{0.418 $\pm$ 0.004} & \textbf{0.432 $\pm$ 0.004} & \textbf{0.476 $\pm$ 0.021} & \textbf{0.464 $\pm$ 0.012} & \multicolumn{1}{r}{\textbf{0.417}} & \textbf{0.428} \\
        & PatchTST    & 0.409 $\pm$ 0.001$^\ddagger$& 0.416 $\pm$ 0.001$^\ddagger$& 0.458 $\pm$ 0.001$^\ddagger$& 0.444 $\pm$ 0.001$^\ddagger$& 0.499 $\pm$ 0.001$^\ddagger$& 0.465 $\pm$ 0.001$^\ddagger$& 0.500 $\pm$ 0.002 & 0.488 $\pm$ 0.001$^\ddagger$& 0.467 & 0.453 \\
        & iTransformer & 0.387 $\pm$ 0.001 & 0.404 $\pm$ 0.001 & 0.441 $\pm$ 0.002$^\ddagger$& 0.436 $\pm$ 0.001$^\ddagger$& 0.487 $\pm$ 0.004$^\ddagger$& 0.458 $\pm$ 0.002$^\ddagger$& 0.502 $\pm$ 0.013$^\dagger$& 0.489 $\pm$ 0.008$^\ddagger$& 0.481 & 0.468 \\
        & TimeMixer   & 0.385 $\pm$ 0.005 & 0.400 $\pm$ 0.002 & 0.445 $\pm$ 0.011$^\ddagger$& 0.433 $\pm$ 0.006$^\ddagger$& 0.492 $\pm$ 0.027$^\ddagger$& 0.460 $\pm$ 0.011$^\dagger$& 0.578 $\pm$ 0.056$^\dagger$& 0.512 $\pm$ 0.025$^\dagger$& 0.454 & 0.447 \\
        & TimesNet    & 0.413 $\pm$ 0.016$^\dagger$& 0.429 $\pm$ 0.011$^\ddagger$& 0.481 $\pm$ 0.007$^\ddagger$& 0.466 $\pm$ 0.004$^\ddagger$& 0.513 $\pm$ 0.010$^\ddagger$& 0.480 $\pm$ 0.007$^\ddagger$& 0.518 $\pm$ 0.009$^\ddagger$& 0.495 $\pm$ 0.003$^\ddagger$& 0.475 & 0.451 \\
        \midrule
         \multirow{5}{*}{Solar} & DMSC (Ours) & \textbf{0.191 $\pm$ 0.014} & \textbf{0.237 $\pm$ 0.010} & \textbf{0.218 $\pm$ 0.004} & \textbf{0.261 $\pm$ 0.004} & \textbf{0.211 $\pm$ 0.020} & \textbf{0.262 $\pm$ 0.013} & \textbf{0.217 $\pm$ 0.006} & \textbf{0.265 $\pm$ 0.004} & \textbf{0.209} & \textbf{0.256} \\
        & PatchTST    & 0.222 $\pm$ 0.002$^\ddagger$& 0.276 $\pm$ 0.002$^\ddagger$& 0.253 $\pm$ 0.001$^\ddagger$& 0.296 $\pm$ 0.004$^\ddagger$& 0.273 $\pm$ 0.001$^\ddagger$& 0.306 $\pm$ 0.001$^\ddagger$& 0.275 $\pm$ 0.005$^\ddagger$& 0.305 $\pm$ 0.002$^\ddagger$& 0.256 & 0.296 \\
        & iTransformer & 0.204 $\pm$ 0.003 & 0.238 $\pm$ 0.002 & 0.238 $\pm$ 0.001 & 0.262 $\pm$ 0.001$^\dagger$& 0.250 $\pm$ 0.001$^\dagger$& 0.274 $\pm$ 0.002 & 0.250 $\pm$ 0.001$^\ddagger$& 0.275 $\pm$ 0.001$^\ddagger$& 0.235 & 0.262 \\
        & TimeMixer   & 0.207 $\pm$ 0.009 & 0.259 $\pm$ 0.009$^\dagger$& 0.250 $\pm$ 0.045 & 0.296 $\pm$ 0.066 & 0.252 $\pm$ 0.008$^\ddagger$& 0.278 $\pm$ 0.008 & 0.238 $\pm$ 0.008$^\ddagger$& 0.285 $\pm$ 0.015 & 0.237 & 0.279 \\
        & TimesNet    & 0.220 $\pm$ 0.005$^\dagger$& 0.256 $\pm$ 0.013 & 0.260 $\pm$ 0.010$^\ddagger$& 0.279 $\pm$ 0.008$^\dagger$& 0.285 $\pm$ 0.011$^\ddagger$& 0.294 $\pm$ 0.004$^\dagger$& 0.285 $\pm$ 0.006$^\ddagger$& 0.301 $\pm$ 0.009$^\ddagger$& 0.262 & 0.282 \\
        \midrule
        \multirow{5}{*}{Weather} & DMSC (Ours) & \textbf{0.161 $\pm$ 0.003} & \textbf{0.207 $\pm$ 0.003} & \textbf{0.208 $\pm$ 0.001} & \textbf{0.250 $\pm$ 0.001} & \textbf{0.253 $\pm$ 0.002} & \textbf{0.285 $\pm$ 0.001} & \textbf{0.313 $\pm$ 0.001} & \textbf{0.333 $\pm$ 0.001} & \multicolumn{1}{r}{\textbf{0.233}} & \textbf{0.268} \\
        & PatchTST    & 0.178 $\pm$ 0.001$^\ddagger$& 0.219 $\pm$ 0.001$^\ddagger$& 0.225 $\pm$ 0.001$^\ddagger$& 0.259 $\pm$ 0.002$^\ddagger$& 0.279 $\pm$ 0.001$^\ddagger$ & 0.298 $\pm$ 0.002$^\ddagger$ & 0.355 $\pm$ 0.001$^\ddagger$ & 0.348 $\pm$ 0.001$^\ddagger$ & 0.259 & 0.281 \\
        & iTransformer & 0.177 $\pm$ 0.002$^\ddagger$ & 0.216 $\pm$ 0.002$^\ddagger$ & 0.227 $\pm$ 0.002$^\ddagger$ & 0.259 $\pm$ 0.002$^\ddagger$ & 0.283 $\pm$ 0.001$^\ddagger$ & 0.300 $\pm$ 0.001$^\ddagger$ & 0.359 $\pm$ 0.001$^\ddagger$ & 0.350 $\pm$ 0.001$^\ddagger$ & 0.262 & 0.281 \\
        & TimeMixer   & 0.163 $\pm$ 0.005 & 0.217 $\pm$ 0.017 & 0.213 $\pm$ 0.007 & 0.253 $\pm$ 0.005 & 0.277 $\pm$ 0.017 $^\dagger$& 0.302 $\pm$ 0.012 $^\dagger$& 0.345 $\pm$ 0.001$^\ddagger$ & 0.345 $\pm$ 0.001$^\ddagger$ & 0.573 & 0.346 \\
        & TimesNet    & 0.171 $\pm$ 0.001$^\ddagger$ & 0.220 $\pm$ 0.002$^\ddagger$ & 0.221 $\pm$ 0.001$^\ddagger$ & 0.262 $\pm$ 0.001$^\ddagger$ & 0.278 $\pm$ 0.001$^\ddagger$ & 0.301 $\pm$ 0.001$^\ddagger$ & 0.354 $\pm$ 0.001$^\ddagger$ & 0.351 $\pm$ 0.001$^\ddagger$ & 0.256 & 0.283 \\
        \bottomrule
    \end{tabular}%
    }
\end{table}

\subsubsection{Forecasting Visualization}
\begin{figure}[!t]
    \centering
    \begin{subfigure}{\textwidth}
        \begin{subfigure}{0.245\textwidth}
            \includegraphics[width=\textwidth]{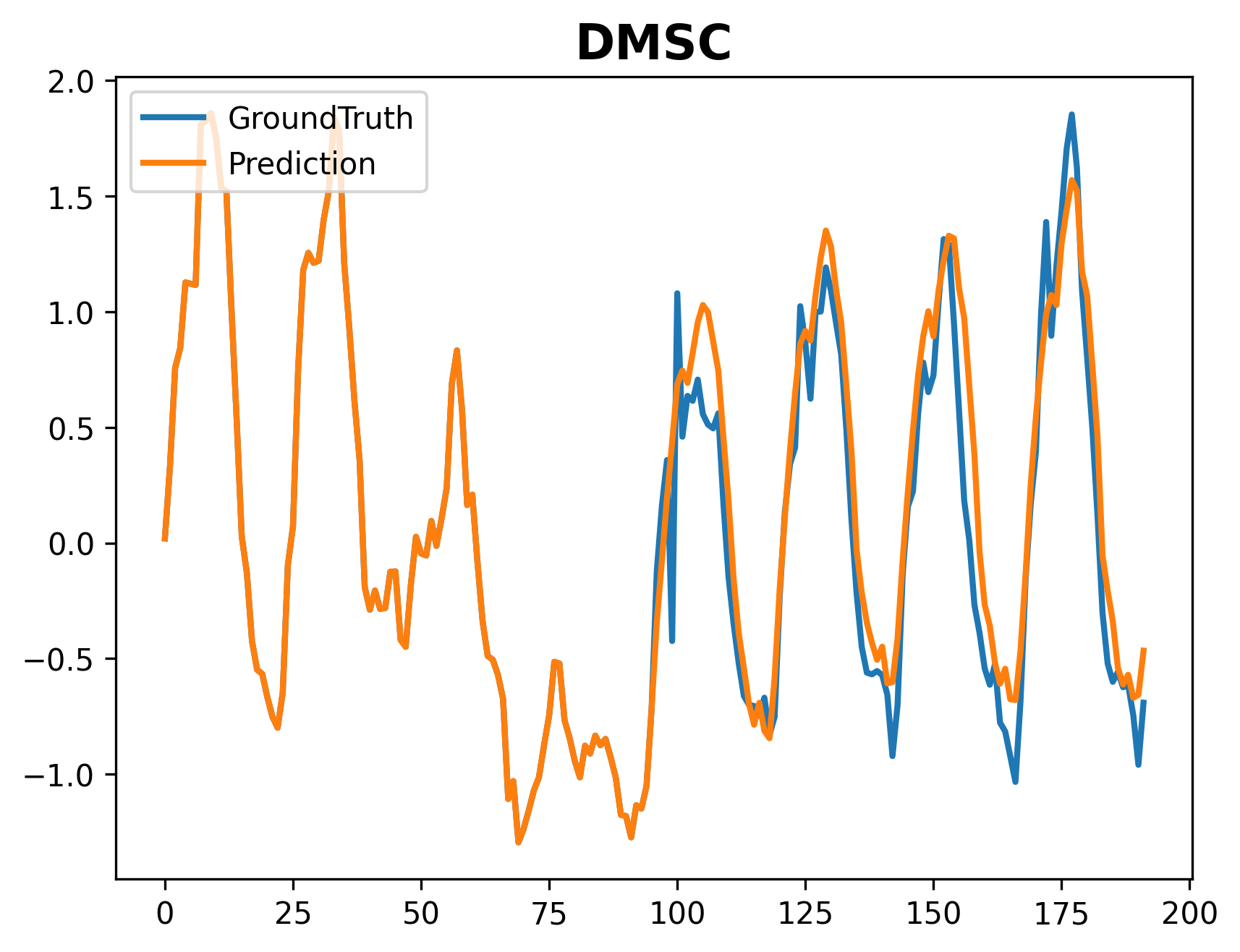}
        \end{subfigure}
        \hspace{-2pt}
        \begin{subfigure}{0.245\textwidth}
            \includegraphics[width=\textwidth]{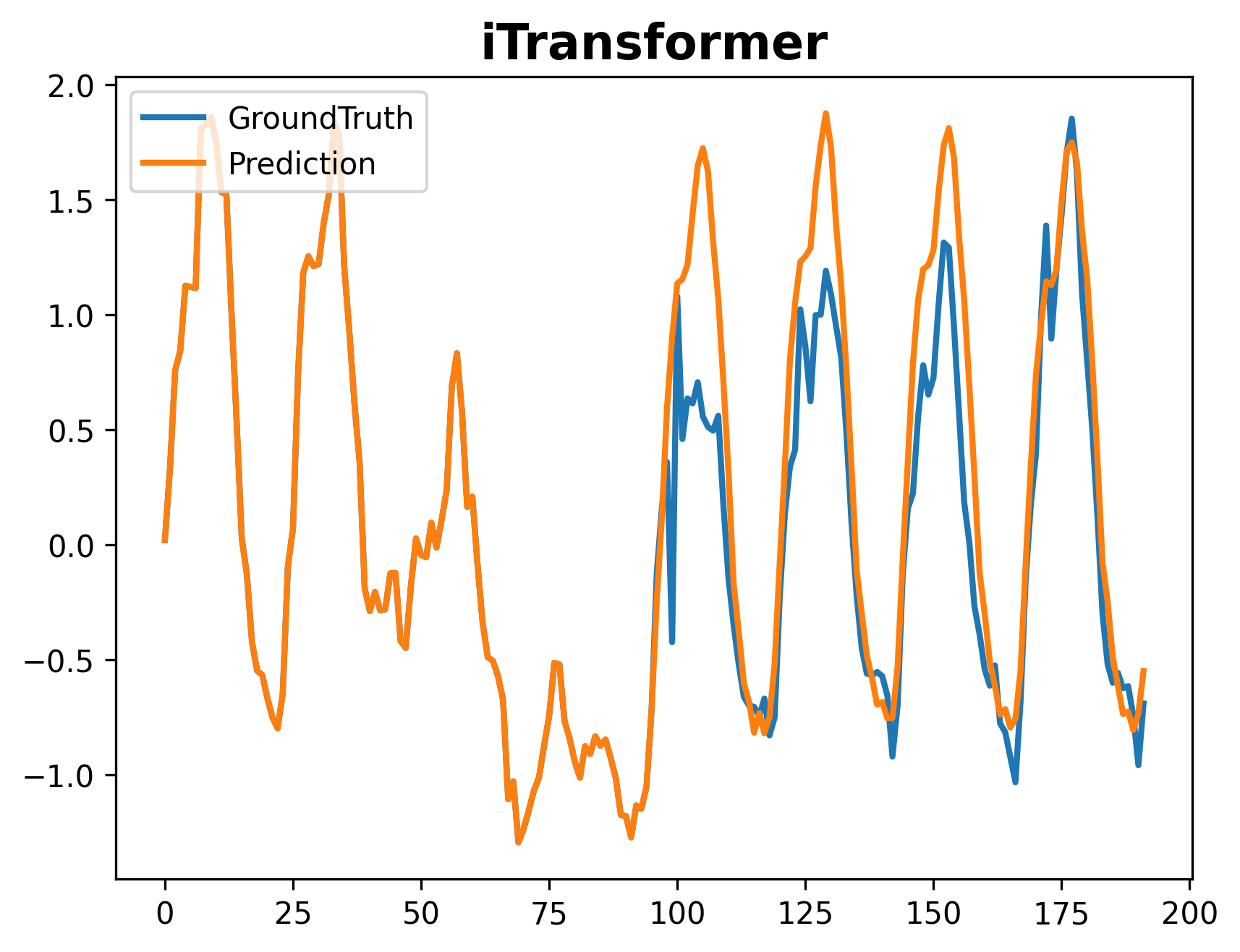}
        \end{subfigure}
        \hspace{-2pt}
        \begin{subfigure}{0.245\textwidth}
            \includegraphics[width=\textwidth]{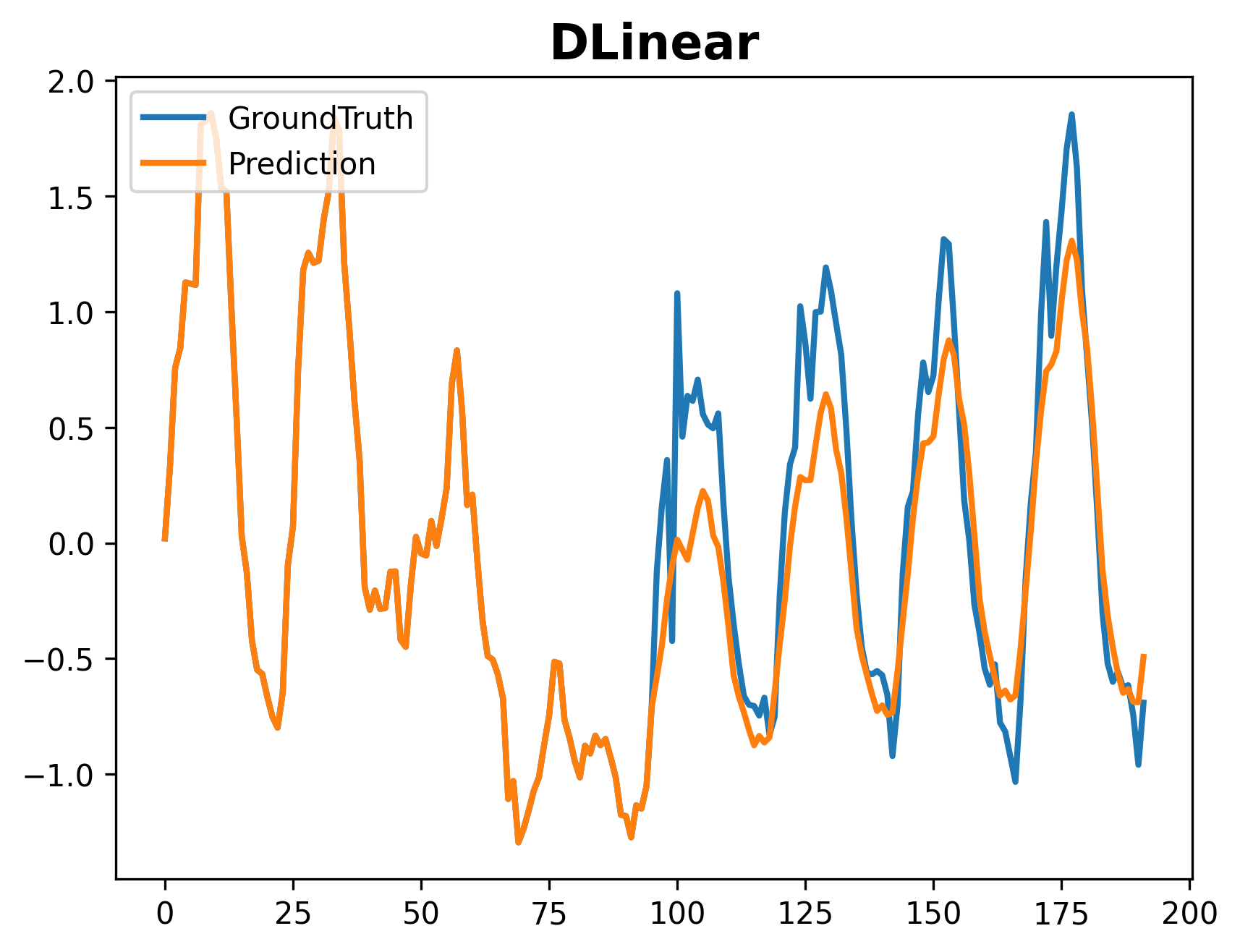}
        \end{subfigure}
        \hspace{-2pt}
        \begin{subfigure}{0.245\textwidth}
            \includegraphics[width=\textwidth]{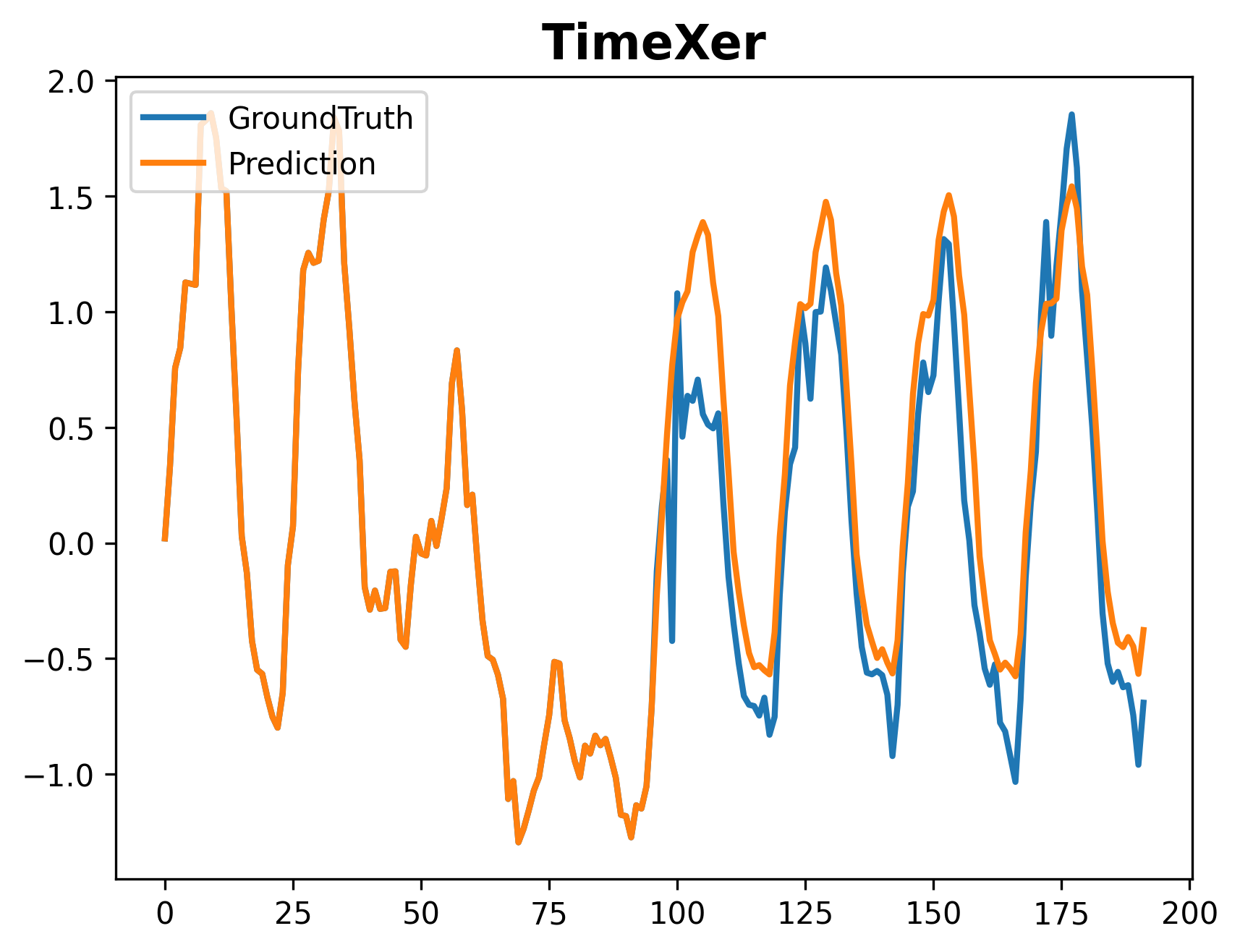}
        \end{subfigure}
    \caption{Visualization on Electricity dataset.}
    \end{subfigure}
    \\
    \begin{subfigure}{\textwidth}
        \begin{subfigure}{0.245\textwidth}
            \includegraphics[width=\textwidth]{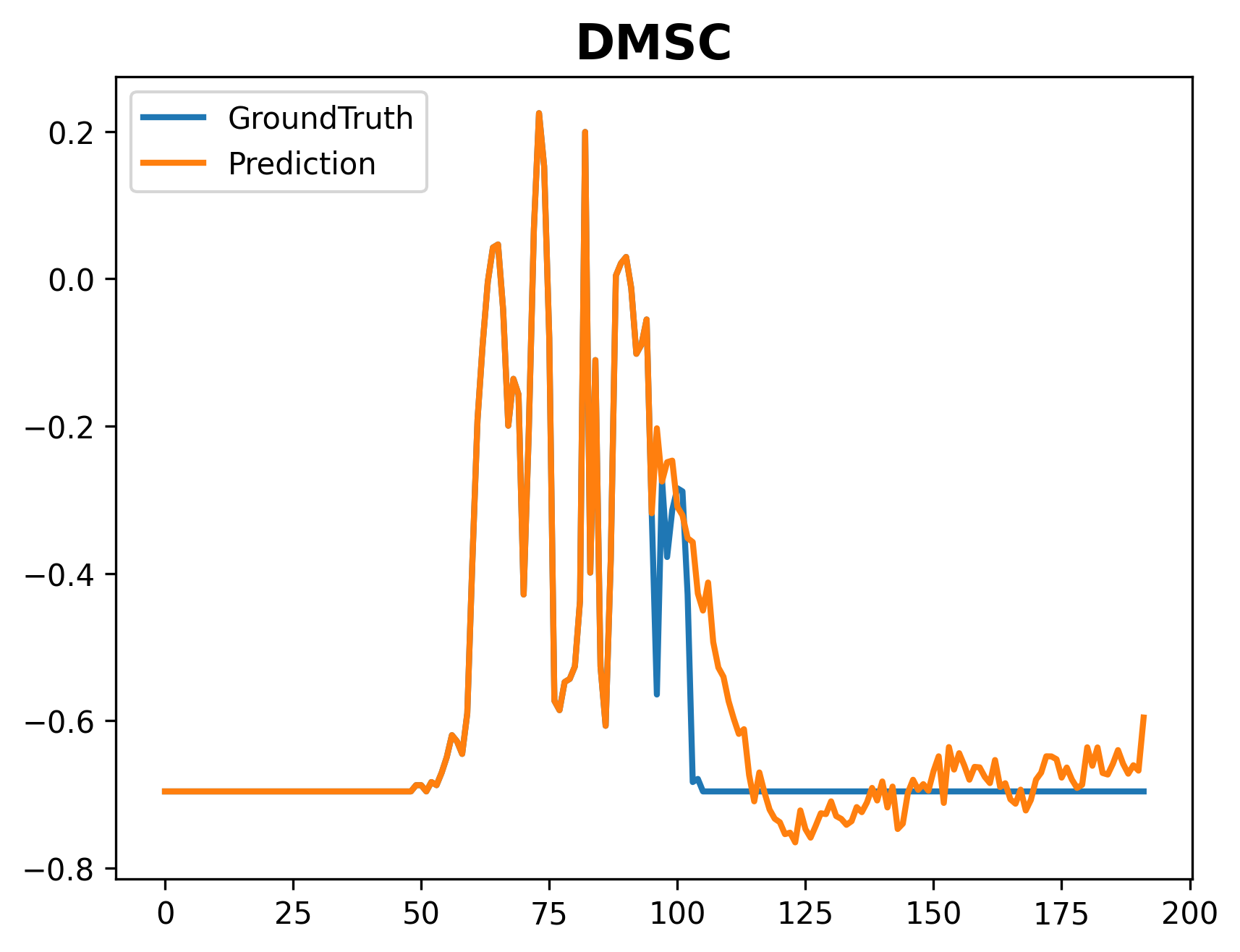}
        \end{subfigure}
        \hspace{-2pt}
        \begin{subfigure}{0.245\textwidth}
            \includegraphics[width=\textwidth]{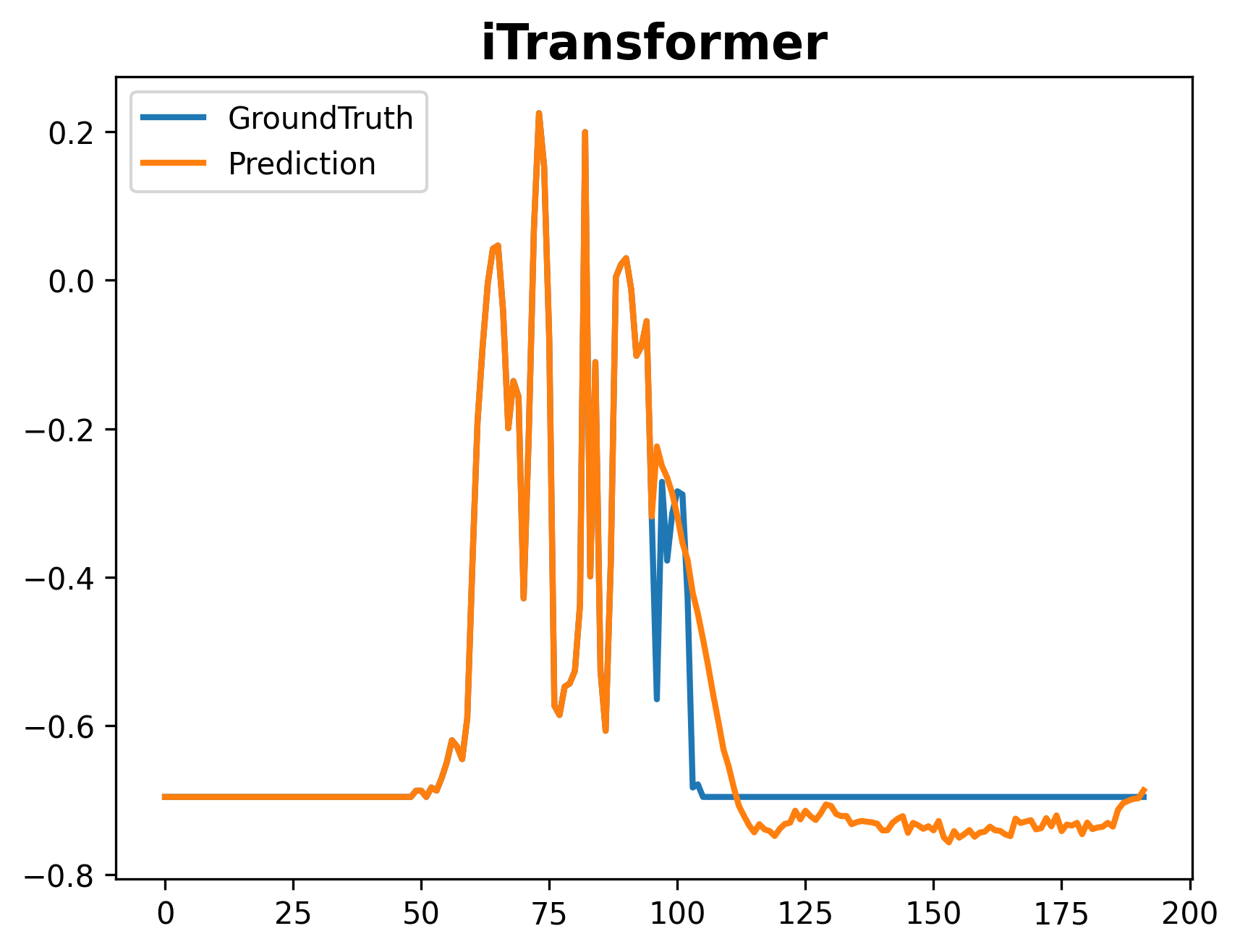}
        \end{subfigure}
        \hspace{-2pt}
        \begin{subfigure}{0.245\textwidth}
            \includegraphics[width=\textwidth]{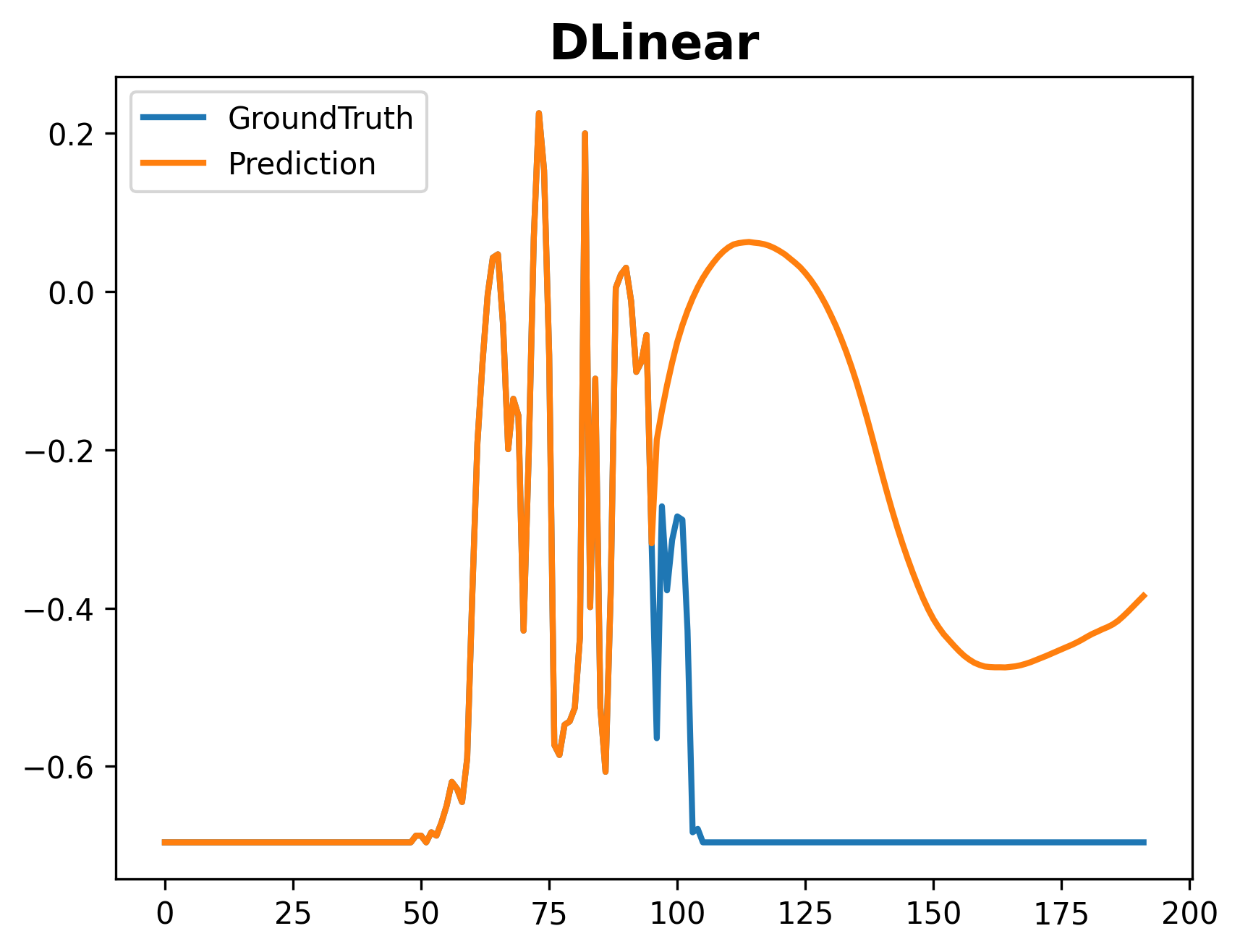}
        \end{subfigure}
        \hspace{-2pt}
        \begin{subfigure}{0.245\textwidth}
            \includegraphics[width=\textwidth]{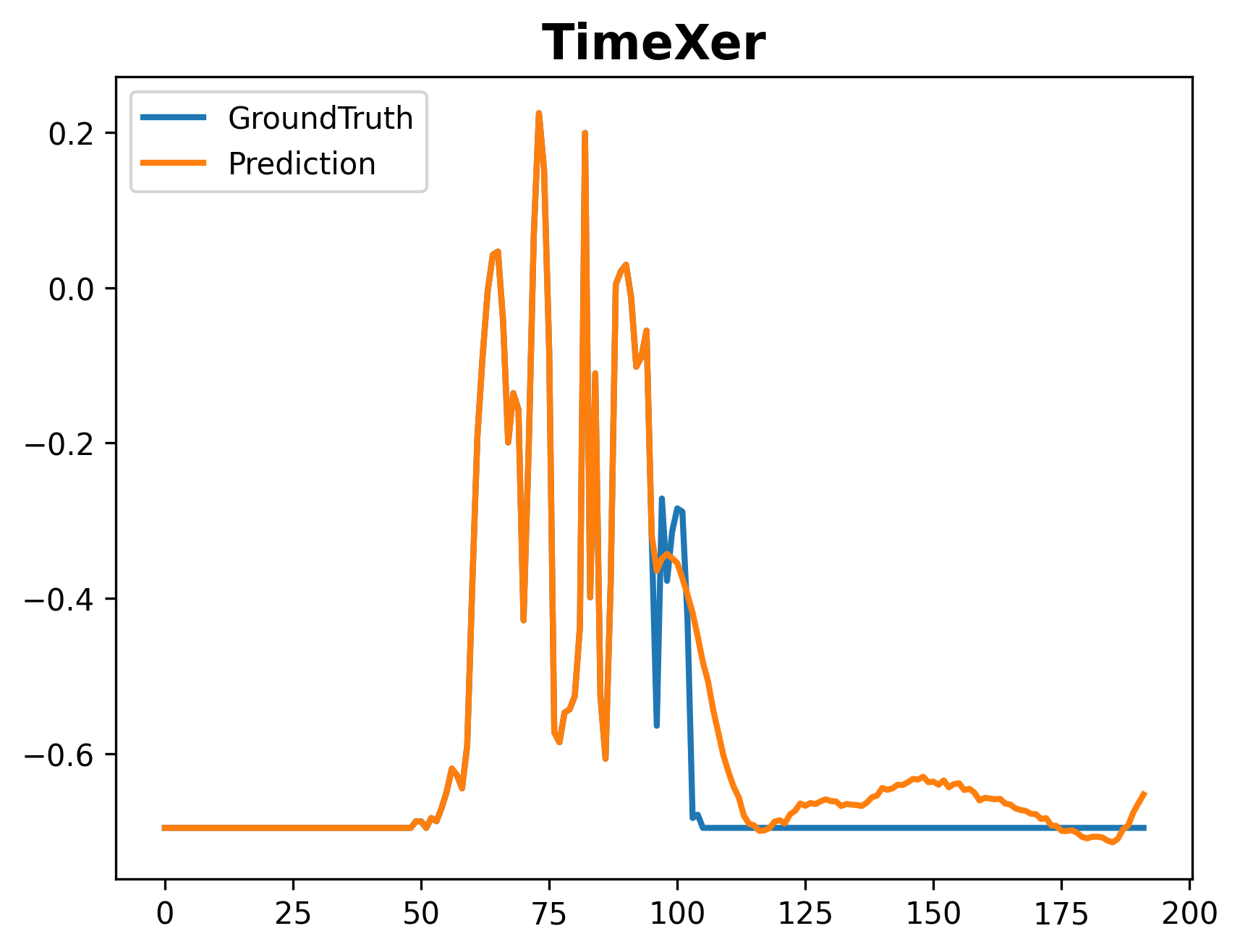}
        \end{subfigure}
    \caption{Visualization on Solar dataset.}
    \end{subfigure}
    \\
    \begin{subfigure}{\textwidth}
        \begin{subfigure}{0.245\textwidth}
            \includegraphics[width=\textwidth]{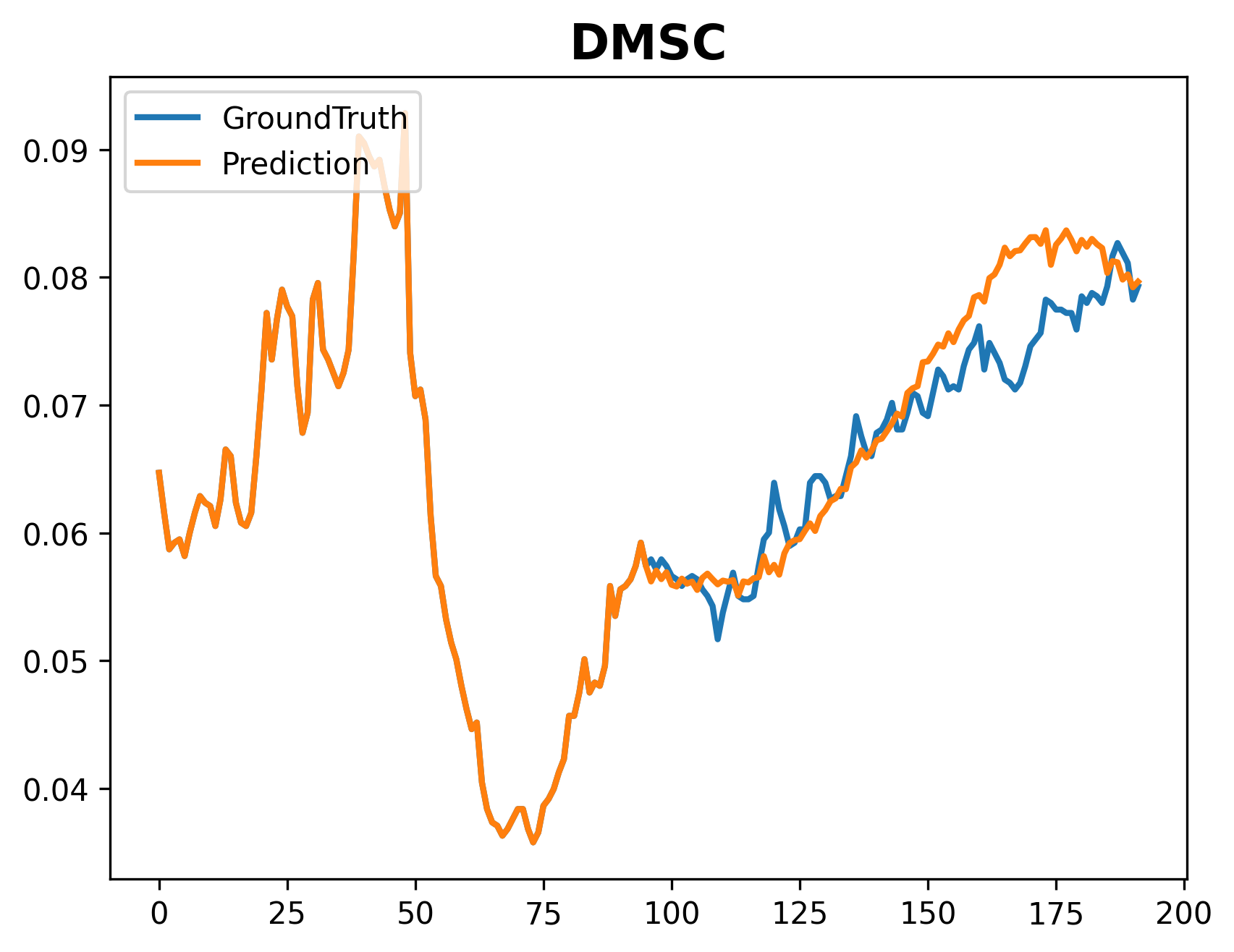}
        \end{subfigure}
        \hspace{-2pt}
        \begin{subfigure}{0.245\textwidth}
            \includegraphics[width=\textwidth]{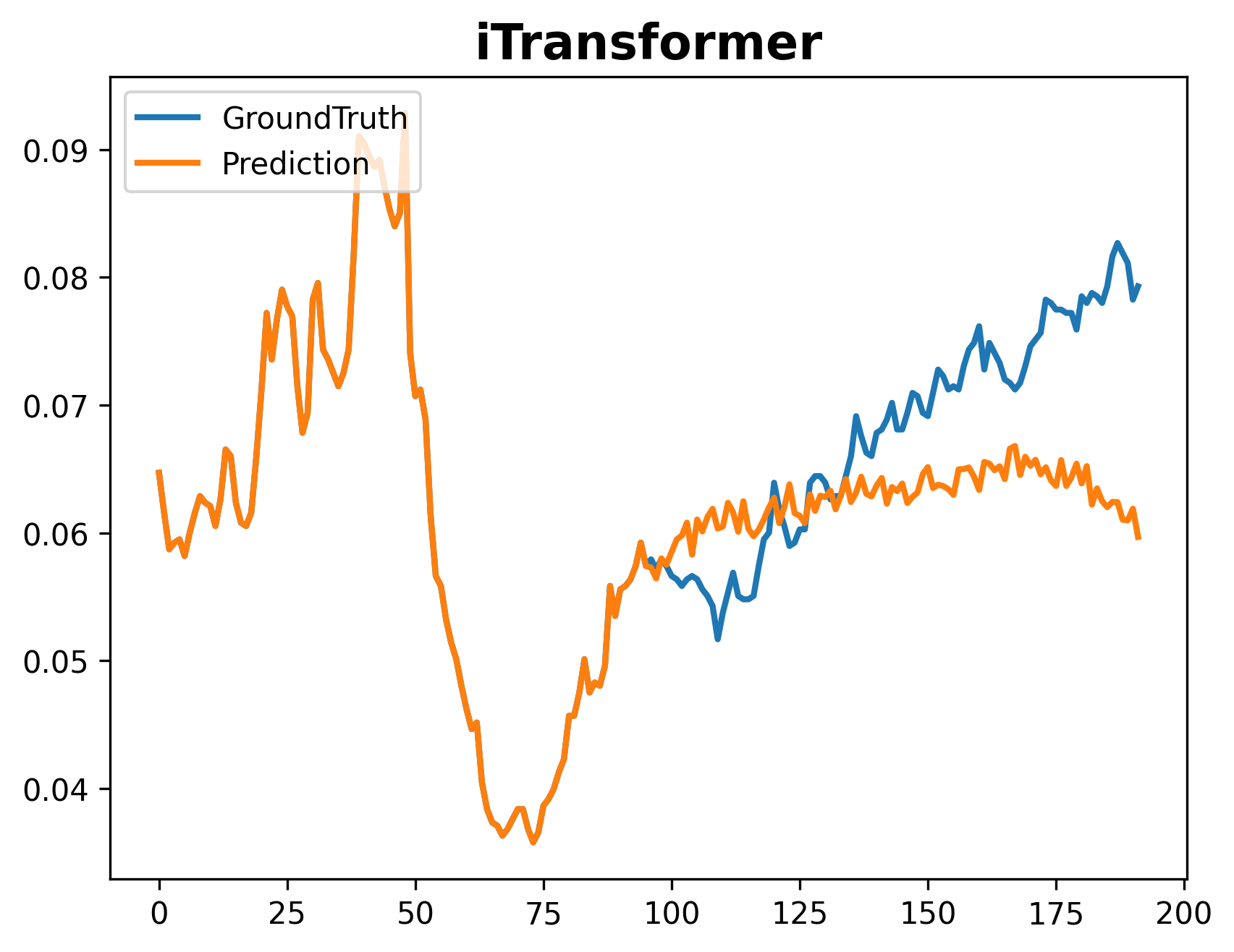}
        \end{subfigure}
        \hspace{-2pt}
        \begin{subfigure}{0.245\textwidth}
            \includegraphics[width=\textwidth]{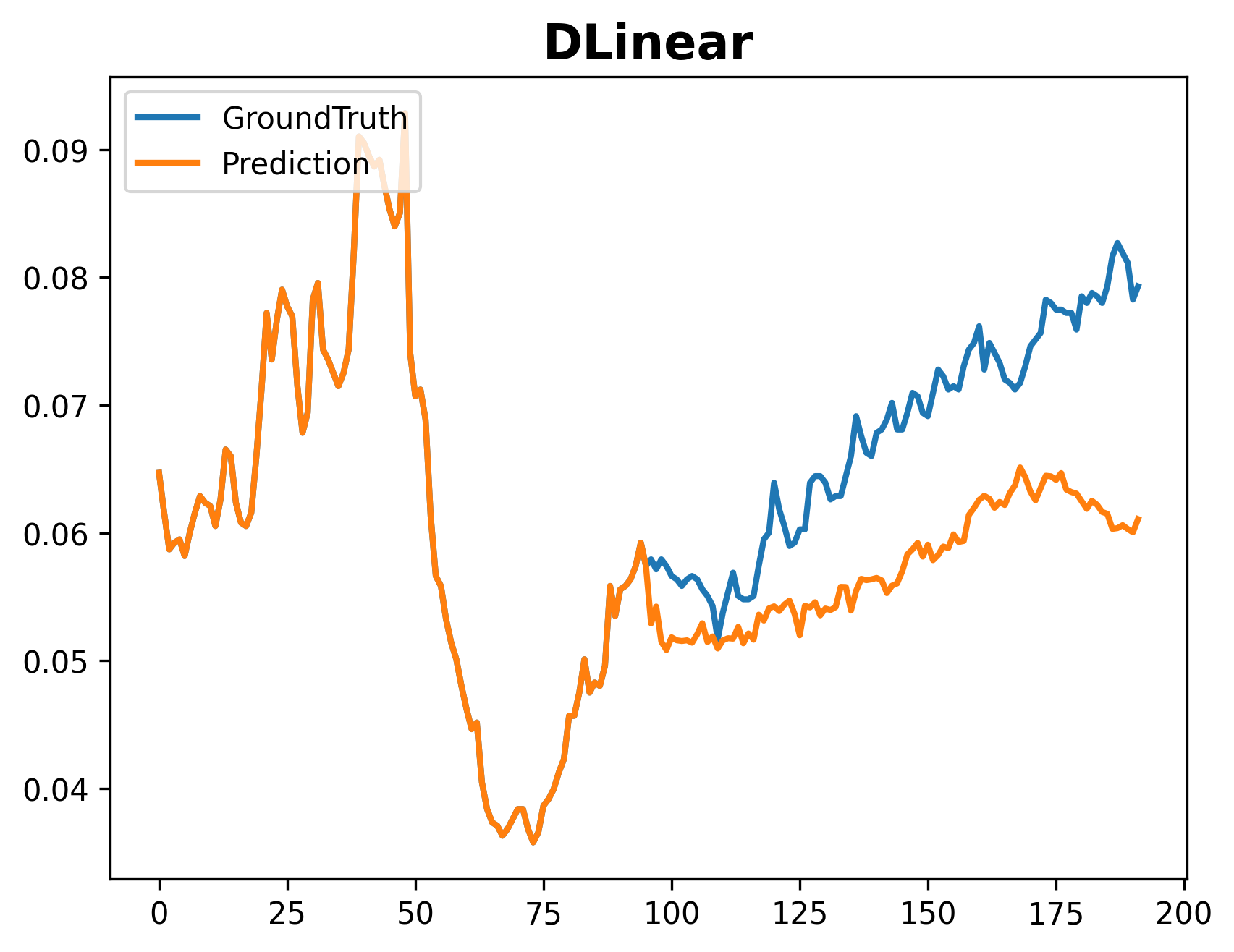}
        \end{subfigure}
        \hspace{-2pt}
        \begin{subfigure}{0.245\textwidth}
            \includegraphics[width=\textwidth]{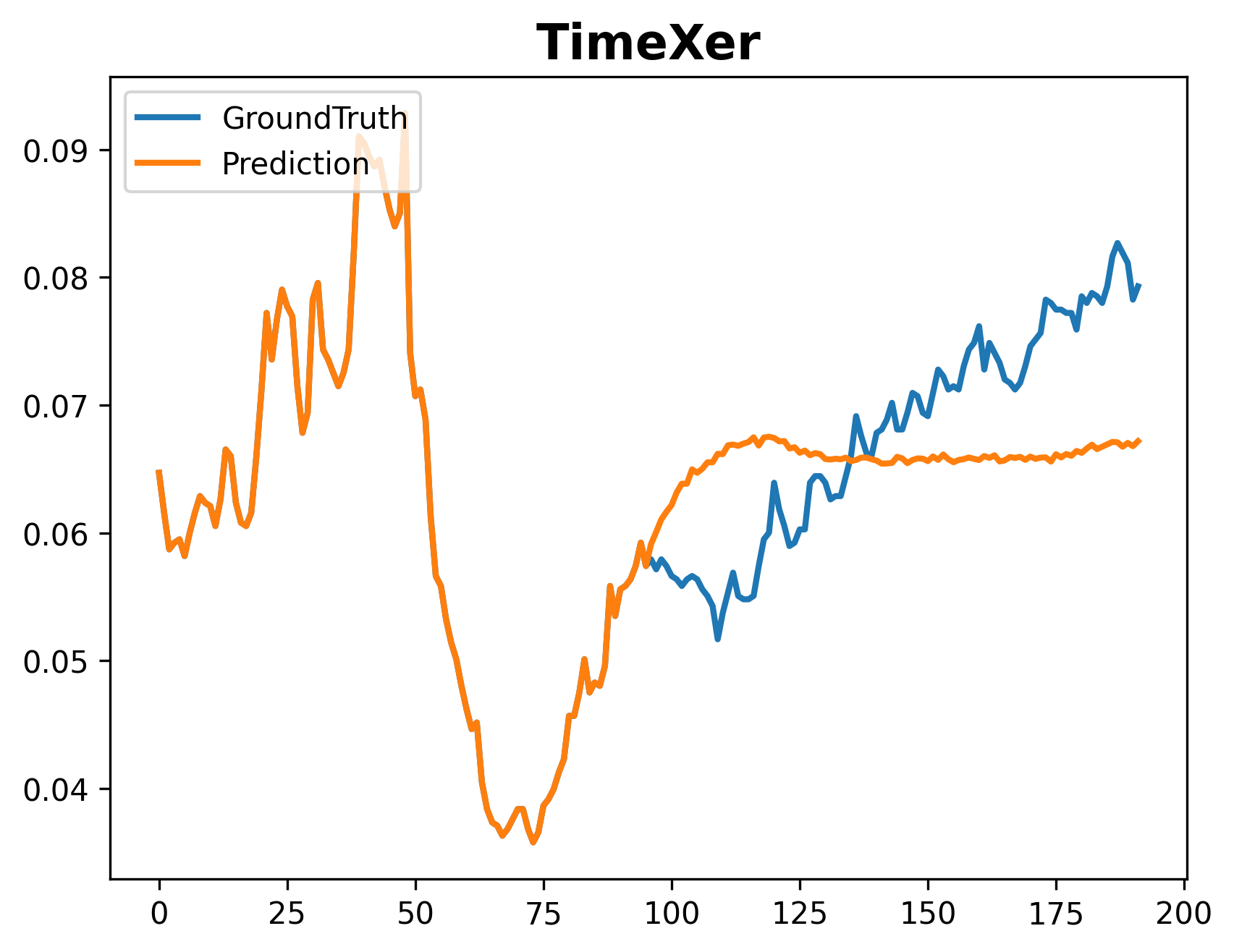}
        \end{subfigure}
    \caption{Visualization on Weather dataset.}
    \end{subfigure}
    \caption{Forecasting visualization of different backbones. The look-back length and prediction length is fixed to 96.}
    \label{forecast_visual}
\end{figure}

To provide an intuitive and comprehensive view of our DMSC framework, we present forecasting visualizations comparing the predictions with other backbones across four representative datasets. As shown in \ref{forecast_visual}, the predictions with our framework are closer to the ground truth. These visualization results serve as qualitative evidence that our DMSC framework effectively captures the temporal dynamics of time series data and yields more accurate forecasting results.

\subsection{Hyperparameter Sensitivity Analysis}
\subsubsection{Embedding Dimension.}
\begin{figure}[!t]
  \centering
  \begin{subfigure}[b]{0.47\textwidth}
    \centering
    \includegraphics[width=\textwidth]{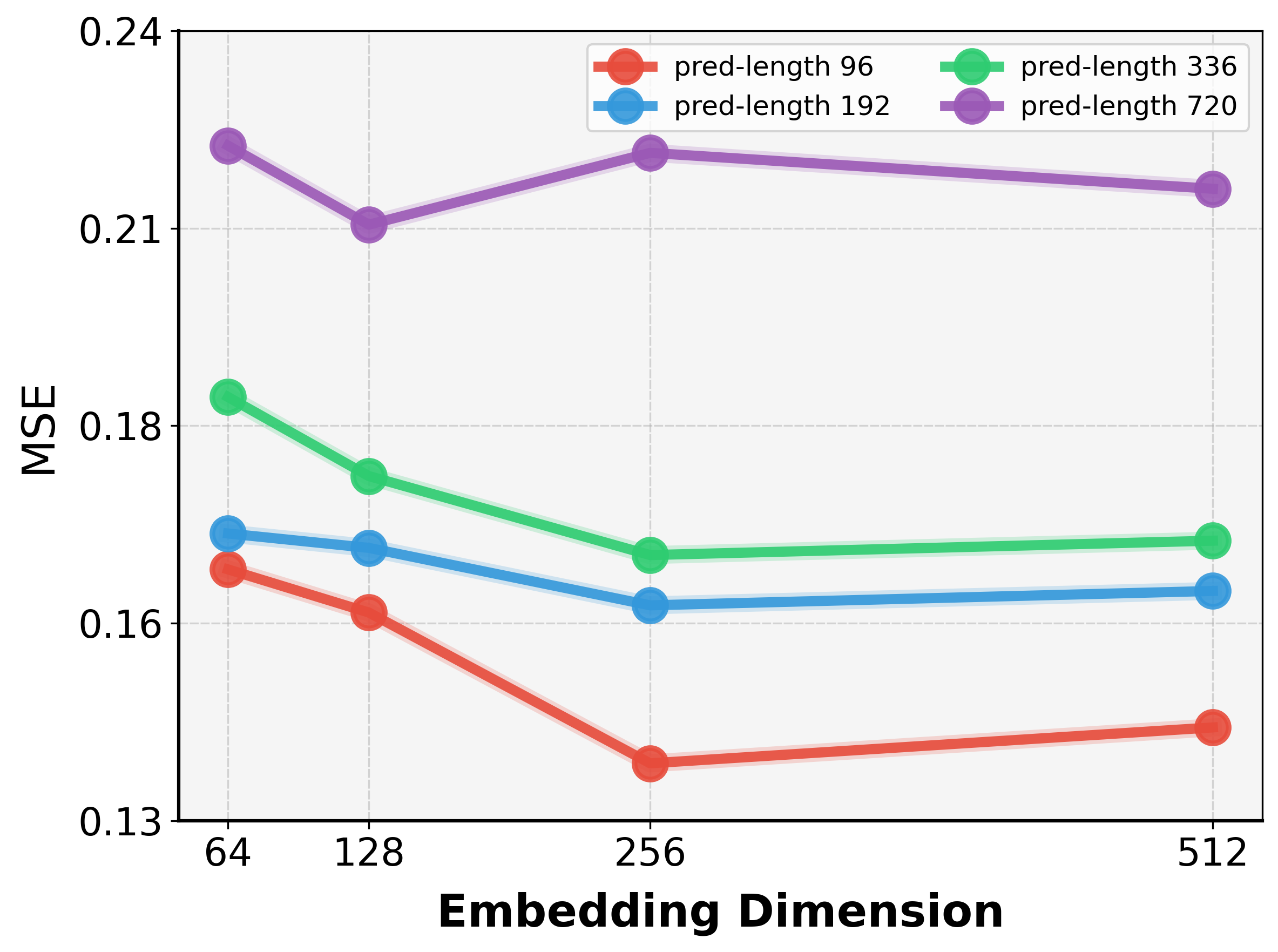}
    \caption{Performance on Electricity dataset.}
    \label{emb}
  \end{subfigure}
  \hfill
  \begin{subfigure}[b]{0.47\textwidth}
    \centering
    \includegraphics[width=\textwidth]{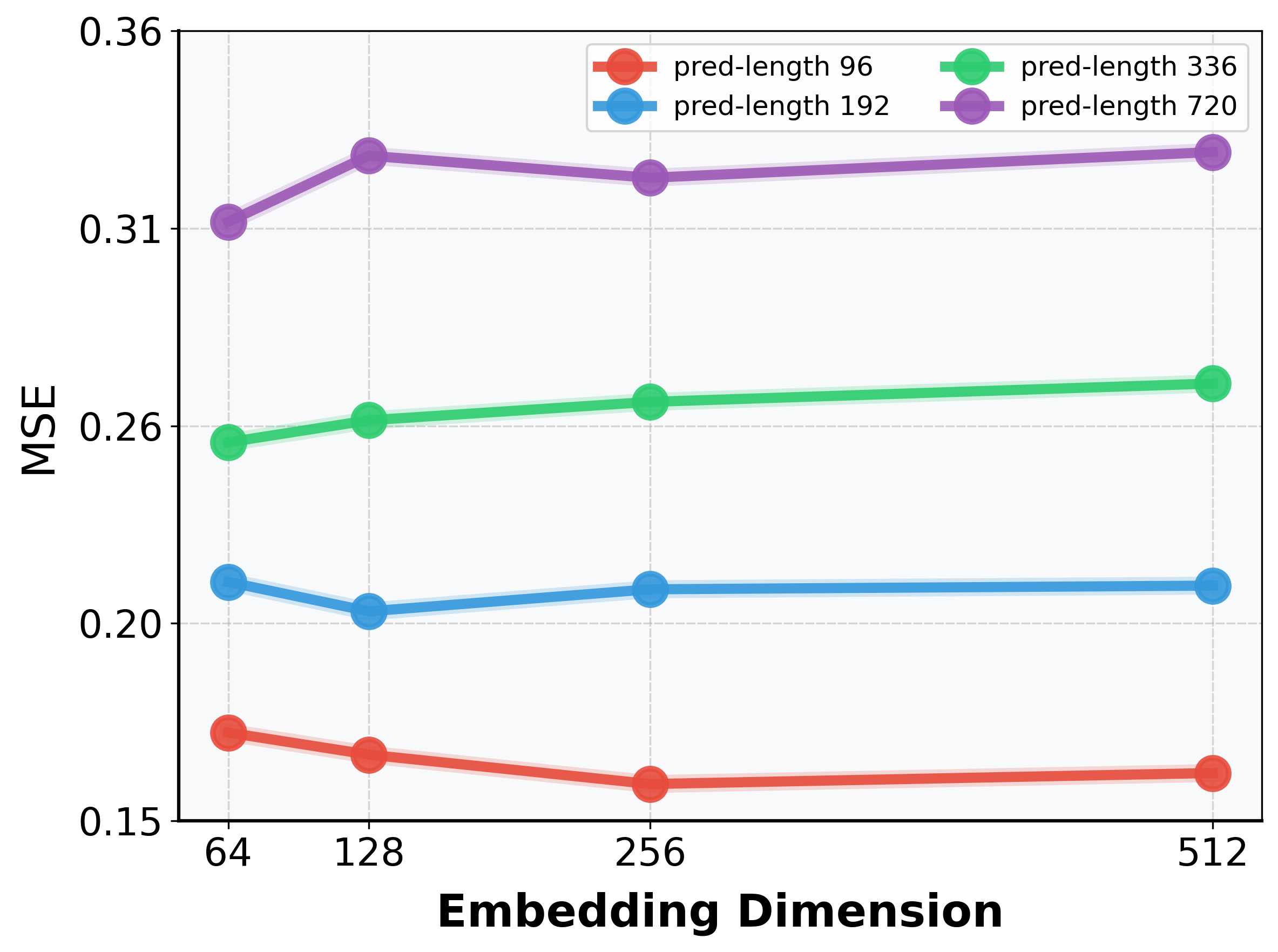}
    \caption{Performance on Weather dataset.}
    \label{emb1}
  \end{subfigure}
  \caption{Hyperparameter sensitivity analysis under different embedding dimension, $d_{\mathrm{model}}$ is set to \{64, 128, 256, 512\}. Left is conducted on Electricity dataset and right is on Weather dataset. Look-back length is fixed to 96, and prediction lengths are set to \{96, 192, 336, 720\}.}
  \label{fig:combined}
\end{figure}

The embedding dimension determines the richness of feature representations. We evaluate the model's performance on ECL and Weather datasets with embedding dimensions $d_{\mathrm{model}} \in \{64, 128,$ $256, 512\}$. The results in Figure \ref{fig:combined} show that the optimal dimension is at 128 / 256, as smaller embedding impairs multi-scale feature separation, while larger configurations yield marginal gains with higher memory cost, indicating dimensionality saturating.

\subsubsection{EMPD Patch Length Decay.}
\begin{figure}[!t]
  \centering
  \begin{subfigure}[b]{0.47\textwidth}
    \includegraphics[width=\textwidth]{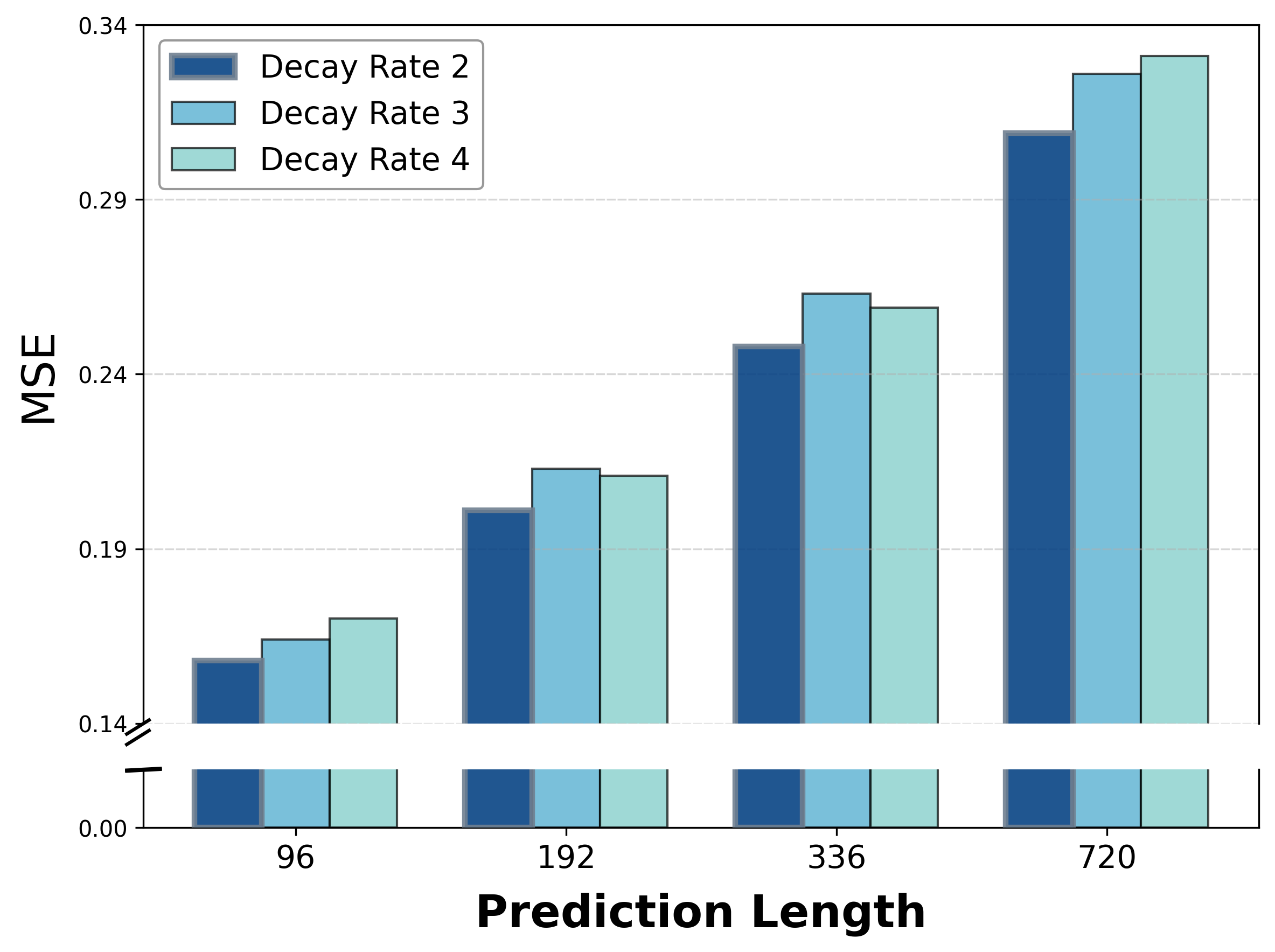}
    \caption{Different patch length decay rate.}
    \label{Patch}
  \end{subfigure}
  \hfill
  \begin{subfigure}[b]{0.47\textwidth}
    \centering
    \includegraphics[width=\textwidth]{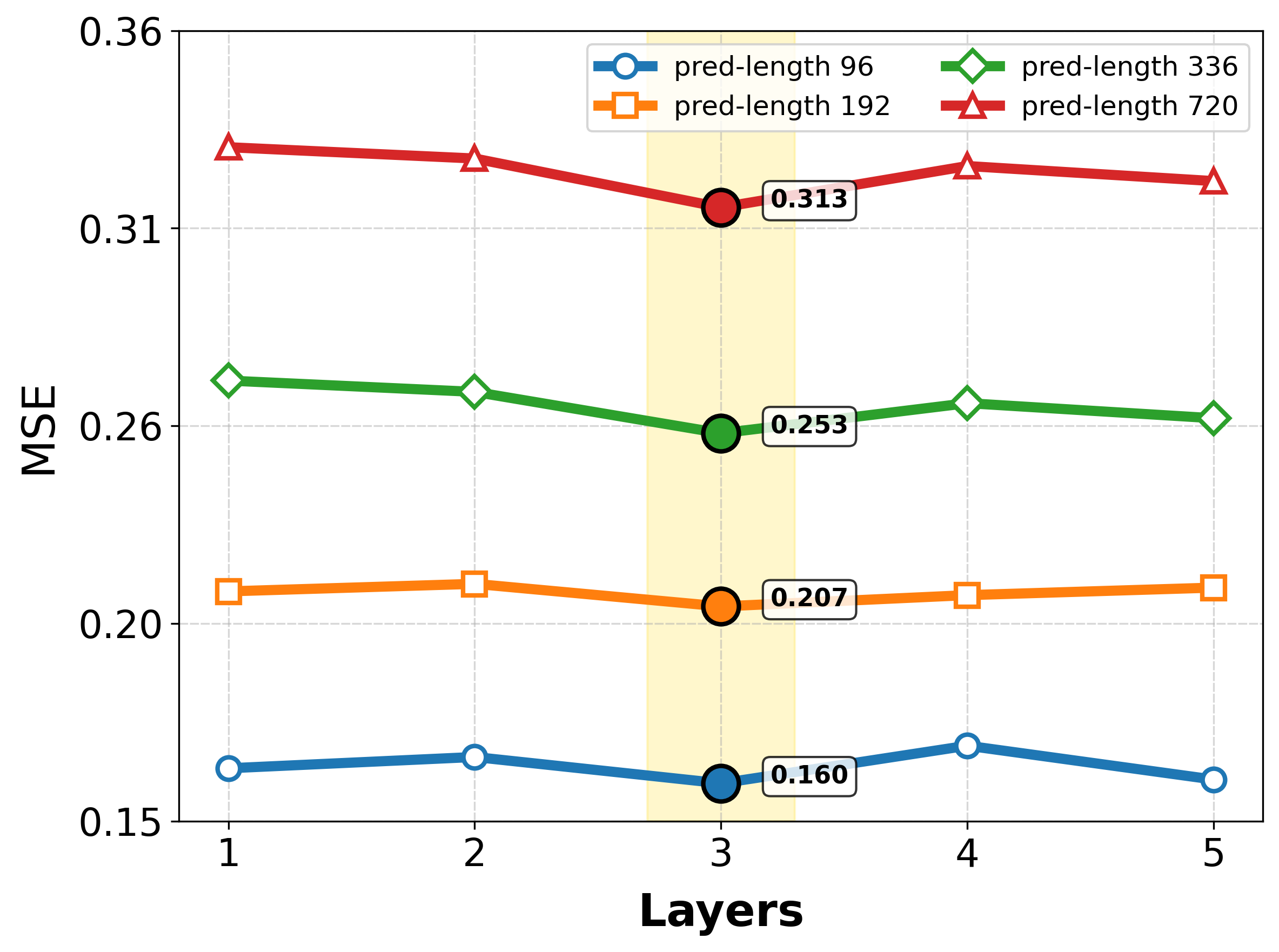}
    \caption{Different progressive cascade layers.}
    \label{Layers}
  \end{subfigure}
  \caption{Hyperparameter sensitivity analysis on Weather dataset. Left is different patch length decay rate, right is different progressive cascade layers.}
  \label{fig:combined1}
\end{figure}

EMPD employs exponentially decaying patch lengths across layers to capture multi-scale dependencies. We evaluate the impact of the EMPD decay rate by testing values of $\tau\in\{2, 3, 4\}$ on Weather dataset. As shown in Figure \ref{Patch}, a decay factor of $\tau=2$ achieves the best forecasting performance, which balances the preservation of sufficient local details in fine-grained layers with the extraction of broader trends in coarse-grained layers.

\subsubsection{Number of Progressive Cascade Layers.}
The number of progressive cascade layers ($l\in\{1, 2, 3, 4, 5\}$) influences hierarchical feature extraction. As shown in Figure \ref{Layers}, three layers can achieve optimal efficiency, shallower stacks (1 - 2 layers) fail to capture fine-grained interactions, while deeper configurations (4 - 5 layers) introduce substantially increased latency with diminishing performance returns.

\subsubsection{Number of global and local experts.}
\begin{figure}[!t]
      \centering
      \begin{subfigure}[b]{0.47\textwidth}
        \includegraphics[width=\textwidth]{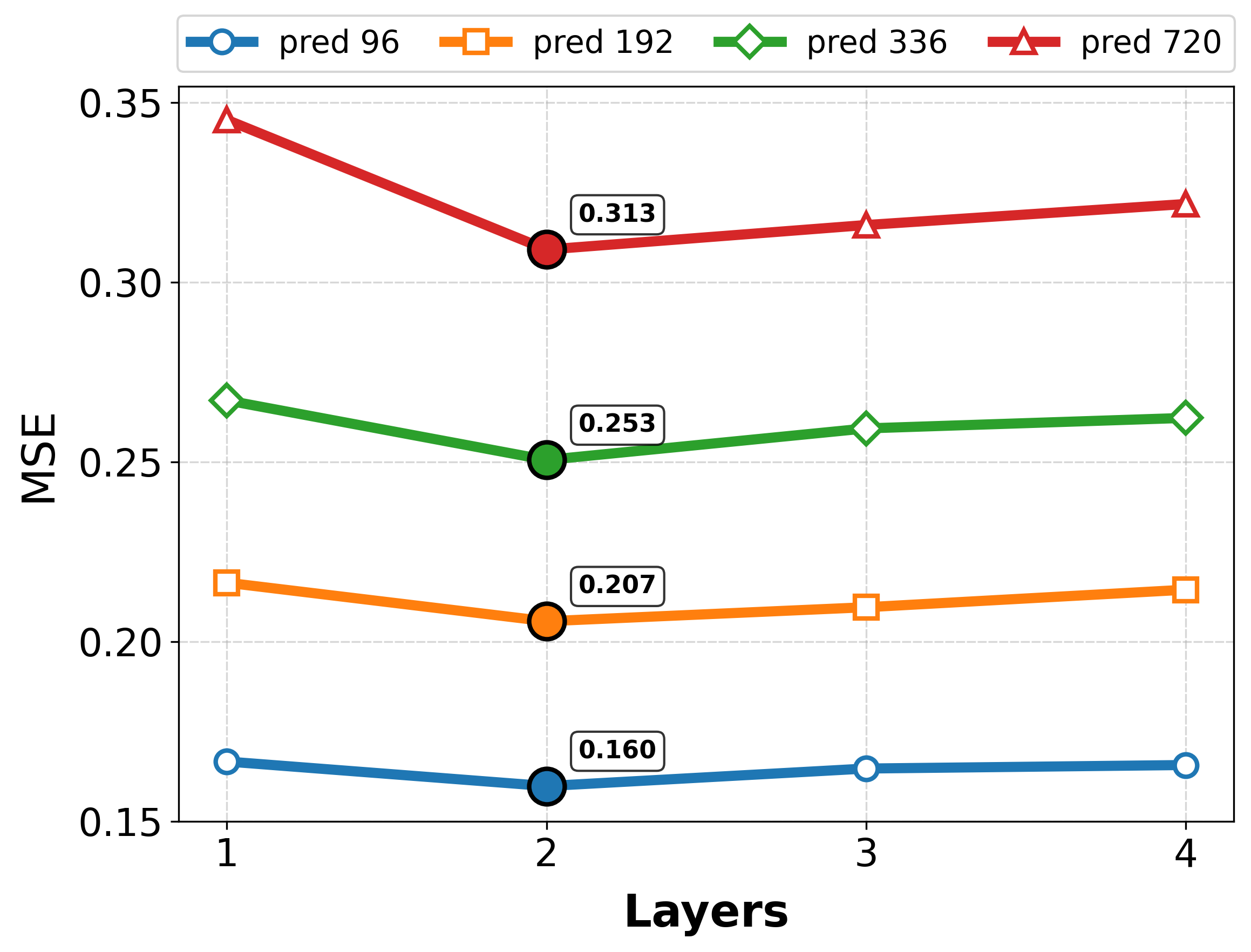}
        \caption{Different number of global experts.}
        \label{global_experts}
      \end{subfigure}
      \hfill
      \begin{subfigure}[b]{0.47\textwidth}
        \centering
        \includegraphics[width=\textwidth]{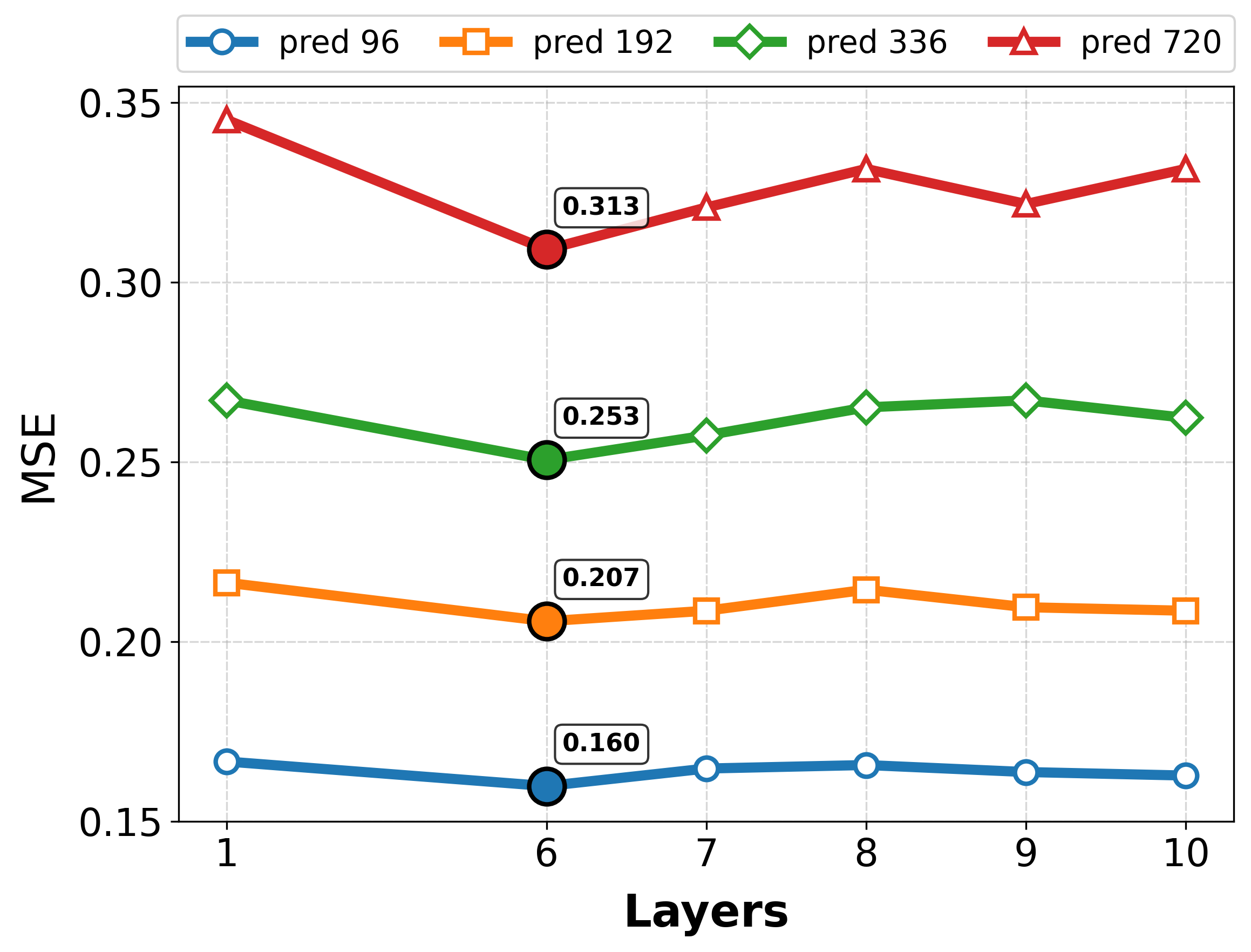}
        \caption{Different number of local experts.}
        \label{local_experts}
      \end{subfigure}
      \caption{Hyperparameter sensitivity analysis of number of experts. Left is global experts, and right is local experts.}
      \label{fig:numofexperts}
\end{figure}

\textcolor{black}{We further examine the sensitivity of DMSC to the number of global experts $m$ and local experts $n$. Figure \ref{fig:numofexperts} presents the forecasting performance on the Weather dataset under varying configurations: $m \in \{1, 2, 3, 4\}$ and $n \in \{1, 6, 7, 8, 9, 10\}$. The results show moderate fluctuations across different combinations, with the optimal performance consistently achieved at $m=2$ global experts and $n=6$ local experts. Using a single expert of either type leads to a noticeable performance drop, as the model lacks sufficient capacity to capture the diverse long - term and short - term patterns. Conversely, increasing the number of experts beyond the optimal point brings marginal gains or slight degradation, likely due to increased routing difficulty and over - parameterization. These results confirm that a moderate number of experts strikes a favorable balance between model capacity and generalization, and that DMSC is not overly sensitive to these hyperparameters within a reasonable range.}

\subsubsection{Effect of balance loss weight $\lambda$.}
\begin{figure}[!t]
	\centering
	\includegraphics[width=0.55\textwidth]{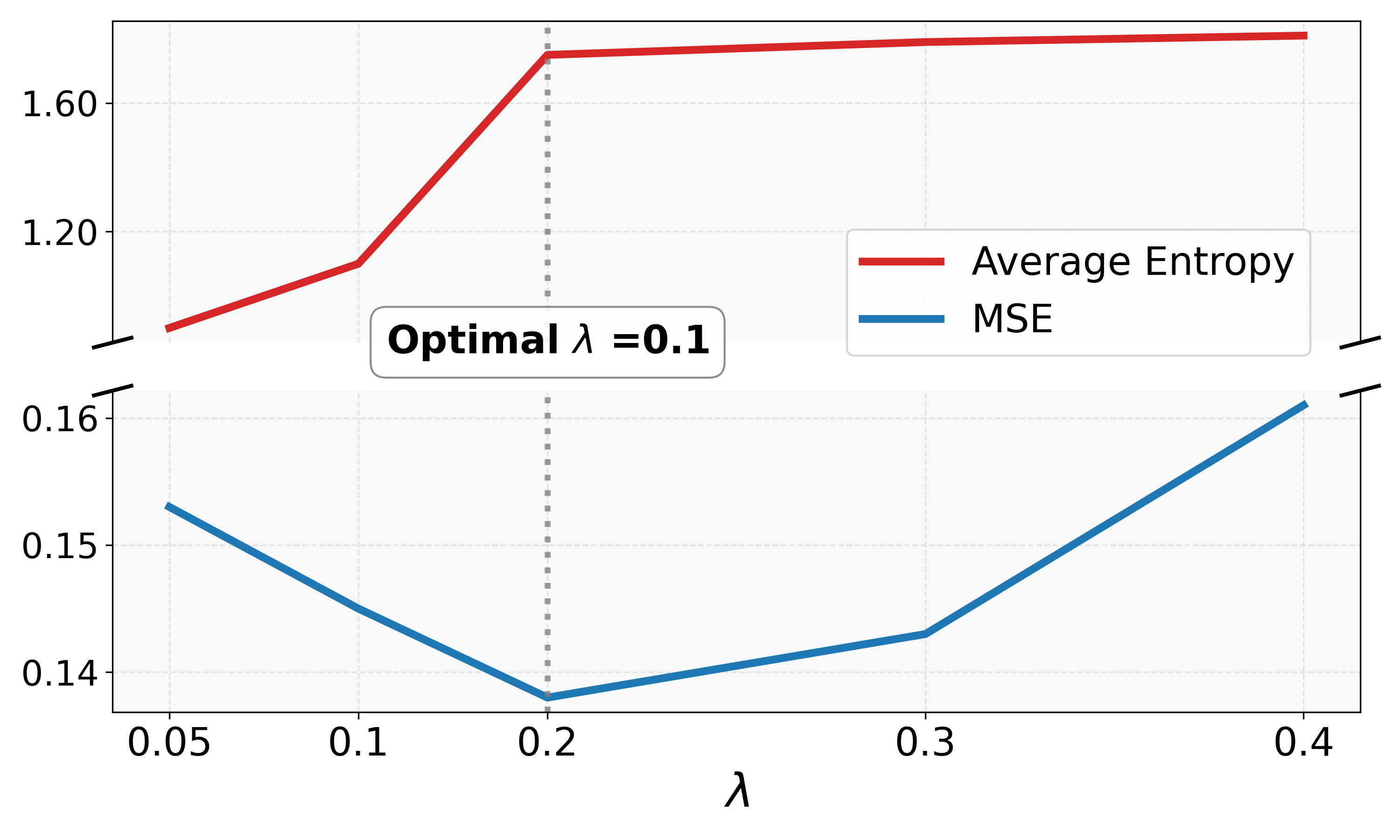}
	\caption{Hyperparameter sensitivity analysis of balance loss weight $\lambda$.}
	\label{lambdaanaly}
\end{figure}

\textcolor{black}{We further examine the influence of the balance loss weight $\lambda$ on both the routing behavior and the forecasting performance. Figure \ref{lambdaanaly} shows the MSE and the average routing entropy $H(\Omega)$ on the Electricity dataset for $\lambda \in \{0.05, 0.1, 0.2, 0.3, 0.4\}$, with all other hyperparameters held constant. As $\lambda$ increases, the routing entropy rises substantially, confirming that the auxiliary loss effectively encourages more uniform expert utilization. The MSE improves steadily and reaches its minimum at $\lambda = 0.2$, where the entropy approaches its highest level. Beyond this point, further increasing $\lambda$ yields only marginal entropy gains while the MSE begins to deteriorate. This indicates that an excessively large $\lambda$ forces the router toward a near - uniform distribution, which diminishes expert specialization and harms predictive accuracy. We therefore adopt $\lambda = 0.2$ as the default value, which achieves the optimal balance between load balancing and forecasting performance.}

\section{Discussion: Applicability and Limitations}
\textcolor{black}{The experimental results presented above demonstrate the strong and consistent performance of DMSC across diverse benchmarks. Here we provide a more fine-grained discussion on the scenarios where DMSC is expected to be most beneficial, as well as its current limitations.}

\textcolor{black}{\textbf{High-dimensional multivariate forecasting.} DMSC shows particularly pronounced improvements on datasets with many variables, such as Electricity (321 variables) and Traffic (862 variables). The cross-variable gating mechanism in TIB adaptively captures inter-series dependencies without a predefined graph, which becomes increasingly valuable as the number of variables grows. For univariate or low-dimensional series, simpler architectures such as DLinear may already achieve competitive performance with lower complexity.}

\textcolor{black}{\textbf{Non-stationary time series.} The dynamic scale adaptation in EMPD and the temporal-aware routing in ASR-MoE are specifically designed to handle distribution shifts and time-varying patterns. The scale factor analysis (Section \ref{alps}) confirms that DMSC adjusts its patch granularity in response to structural changes in the input. DMSC is therefore expected to offer greater advantages on non-stationary data. On highly regular, stationary series, the benefit of dynamic adaptation diminishes, and a static decomposition may suffice.}

\textcolor{black}{\textbf{Long-horizon forecasting.} The hierarchical coarse-to-fine cascade and the global experts in ASR-MoE are tailored for capturing persistent long-term trends. The increasing look-back experiment (Figure \ref{input_length}) shows that DMSC benefits from longer histories, and the relative gain over baselines tends to grow with prediction length. For very short horizons, lightweight models may provide comparable accuracy with lower cost.}

\textcolor{black}{\textbf{Computational considerations.} Although DMSC achieves favorable efficiency (Section \ref{ea}), its multi-layer architecture and MoE prediction head incur a moderate parameter and computation overhead compared to extremely simple linear models. In deployment scenarios with severe resource constraints (e.g., edge devices with very limited memory), lightweight MLP-based methods remain the most practical choice.}

\section{Conclusion}
This paper presents the Dynamic Multi-Scale Coordination (DMSC) framework, a \textcolor{black}{unified} approach that advances time series forecasting through comprehensive dynamic multi-scale coordination. DMSC has three key contributions: 1) Embedded Multi-Scale Patch Decomposition(EMPD) dynamically decomposes time series into hierarchical patches, eliminating predefined scale constraints through input-adaptive granularity adjustment; 2) Triad Interaction Block (TIB) jointly models intra-patch, inter-patch, and cross-variable dependencies, forming a coarse-to-fine feature pyramid through progressive cascade layers; 3) Adaptive Scale Routing MoE (ASR-MoE) dynamically fuses multi-scale predictions via temporal-aware weighting of specialized global and local experts. Extensive experiments on thirteen benchmarks demonstrate that DMSC achieves SOTA performance and superior efficiency \textcolor{black}{through systematic integration}. Future work will extend the framework to multi-task learning and explore optimizations on complex real-world environments, thereby enhancing its applicability across diverse scenarios.

\bibliographystyle{ACM-Reference-Format}
\bibliography{ref}

@article{Alvarez2011,
  title={Energy time series forecasting based on pattern sequence similarity},
  author={Alvarez, Francisco Martinez and Troncoso, Alicia and Riquelme, Jose C and Ruiz, Jesus S Aguilar},
  journal={IEEE Transactions on Knowledge and Data Engineering},
  volume={23},
  number={8},
  pages={1230--1243},
  year={2010},
  publisher={IEEE}
}

@article{Guo2015,
  title={Ecomark 2.0: empowering eco-routing with vehicular environmental models and actual vehicle fuel consumption data},
  author={Guo, Chenjuan and Yang, Bin and Andersen, Ove and Jensen, Christian S and Torp, Kristian},
  journal={GeoInformatica},
  volume={19},
  number={3},
  pages={567--599},
  year={2015},
  publisher={Springer}
}

@inproceedings{Luan2019,
  title={Representation learning for early sepsis prediction},
  author={Tran, Luan and Nguyen, Manh and Shahabi, Cyrus},
  booktitle={2019 Computing in Cardiology (CinC)},
  pages={1--4},
  year={2019},
  organization={IEEE}
}

@article{Kaimin2022,
  title={Cancer classification with data augmentation based on generative adversarial networks},
  author={Wei, Kaimin and Li, Tianqi and Huang, Feiran and Chen, Jinpeng and He, Zefan},
  journal={Frontiers of Computer Science},
  volume={16},
  number={2},
  pages={162601},
  year={2022},
  publisher={Springer}
}

@article{Chenjuan2020,
  title={Context-aware, preference-based vehicle routing},
  author={Guo, Chenjuan and Yang, Bin and Hu, Jilin and Jensen, Christian S and Chen, Lu},
  journal={The VLDB Journal},
  volume={29},
  number={5},
  pages={1149--1170},
  year={2020},
  publisher={Springer}
}

@article{Jin2021,
  title={TrafficBERT: Pre-trained model with large-scale data for long-range traffic flow forecasting},
  author={Jin, KyoHoon and Wi, JeongA and Lee, EunJu and Kang, ShinJin and Kim, SooKyun and Kim, YoungBin},
  journal={Expert Systems with Applications},
  volume={186},
  pages={115738},
  year={2021},
  publisher={Elsevier}
}

@article{Haixu2023,
  title={Interpretable weather forecasting for worldwide stations with a unified deep model},
  author={Wu, Haixu and Zhou, Hang and Long, Mingsheng and Wang, Jianmin},
  journal={Nature Machine Intelligence},
  volume={5},
  number={6},
  pages={602--611},
  year={2023},
  publisher={Nature Publishing Group UK London}
}

@article{Kaifeng2023,
  title={Accurate medium-range global weather forecasting with 3D neural networks},
  author={Bi, Kaifeng and Xie, Lingxi and Zhang, Hengheng and Chen, Xin and Gu, Xiaotao and Tian, Qi},
  journal={Nature},
  volume={619},
  number={7970},
  pages={533--538},
  year={2023},
  publisher={Nature Publishing Group UK London}
}

@article{Chen2023,
   title={ChatGPT Informed Graph Neural Network for Stock Movement Prediction},
   ISSN={1556-5068},
   journal={SSRN Electronic Journal},
   publisher={Elsevier BV},
   author={Chen, Zihan and Zheng, Lei and Lu, Cheng and Yuan, Jialu and Zhu, Di},
   year={2023} }

@misc{Yux2023,
  title={Temporal data meets LLM--explainable financial time series forecasting},
  author={Yu, Xinli and Chen, Zheng and Ling, Yuan and Dong, Shujing and Liu, Zongyi and Lu, Yanbin},
  journal={arXiv preprint arXiv:2306.11025},
  year={2023}
}

@inproceedings{Zhong2024,
  title={Urbangpt: Spatio-temporal large language models},
  author={Li, Zhonghang and Xia, Lianghao and Tang, Jiabin and Xu, Yong and Shi, Lei and Xia, Long and Yin, Dawei and Huang, Chao},
  booktitle={Proceedings of the 30th ACM SIGKDD Conference on Knowledge Discovery and Data Mining},
  pages={5351--5362},
  year={2024}
}

@inproceedings{Yuan2024,
  title={Unist: A prompt-empowered universal model for urban spatio-temporal prediction},
  author={Yuan, Yuan and Ding, Jingtao and Feng, Jie and Jin, Depeng and Li, Yong},
  booktitle={Proceedings of the 30th ACM SIGKDD Conference on Knowledge Discovery and Data Mining},
  pages={4095--4106},
  year={2024}
}

@article{Qiuz2024,
title={TFB: Towards Comprehensive and Fair Benchmarking of Time Series Forecasting Methods},
author={Qiu, Xiangfei and Hu, Jilin and Zhou, Lekui and Wu, Xingjian and Du, Junyang and Zhang, Buang and Guo, Chenjuan and Zhou, Aoying and Jensen, Christian S. and Sheng, Zhenli and Yang, Bin},
journal={Proceedings of the VLDB Endowment},
issue={9},
pages={2363-2377},
year={2024},
}

@article{Zonglei2023,
  title={Long sequence time-series forecasting with deep learning: A survey},
  author={Chen, Zonglei and Ma, Minbo and Li, Tianrui and Wang, Hongjun and Li, Chongshou},
  journal={Information Fusion},
  volume={97},
  pages={101819},
  year={2023},
  publisher={Elsevier}
}

@incollection{GuyPNason2006,
title = {Stationary and non-stationary time series},
author = {Nason, G. P.},
booktitle = {Statistics in Volcanology},
publisher = {Geological Society of London},
year = {2006},
month = {01},
}

@book{Seasonal2016,
title={Seasonal Adjustment Methods and Real Time Trend-Cycle Estimation},
author={Estela Bee Dagum and Silvia Bianconcini},
publisher={Springer Nature},
volume={8},
year={2016},
}

@article{Zezhi2025,
title={Exploring Progress in Multivariate Time Series Forecasting: Comprehensive Benchmarking and Heterogeneity Analysis},
author={Zezhi Shao and Fei Wang and Yongjun Xu and Wei Wei and Chengqing Yu and Zhao Zhang and Di Yao and Tao Sun and Guangyin Jin and Xin Cao and Gao Cong and Christian S. Jensen and Xueqi Cheng},
journal={IEEE Transactions on Knowledge and Data Engineering},
issue={1},
pages={291-305},
year={2025},
}

@misc{TimeMixer2025,
  title={TimeMixer++: A General Time Series Pattern Machine for Universal Predictive Analysis}, 
  author={Shiyu Wang and Jiawei Li and Xiaoming Shi and Zhou Ye and Baichuan Mo and Wenze Lin and Shengtong Ju and Zhixuan Chu and Ming Jin},
  year={2025},
  eprint={2410.16032},
  archivePrefix={arXiv},
}

@misc{TimeMixer2024,
  title={TimeMixer: Decomposable Multiscale Mixing for Time Series Forecasting}, 
  author={Shiyu Wang and Haixu Wu and Xiaoming Shi and Tengge Hu and Huakun Luo and Lintao Ma and James Y. Zhang and Jun Zhou},
  year={2024},
  eprint={2405.14616},
  archivePrefix={arXiv},
}

@misc{TimeKAN2025,
  title={TimeKAN: KAN-based Frequency Decomposition Learning Architecture for Long-term Time Series Forecasting}, 
  author={Songtao Huang and Zhen Zhao and Can Li and Lei Bai},
  year={2025},
  eprint={2502.06910},
  archivePrefix={arXiv},
}

@misc{TVNet2025,
  title={TVNet: A Novel Time Series Analysis Method Based on Dynamic Convolution and 3D-Variation}, 
  author={Chenghan Li and Mingchen Li and Ruisheng Diao},
  year={2025},
  eprint={2503.07674},
  archivePrefix={arXiv},
}

@inproceedings{TimesNet2023,
title={TimesNet: Temporal 2D-Variation Modeling for General Time Series Analysis},
author={Wu, Haixu and Hu, Tengge and Liu, Yong and Zhou, Hang and Wang, Jianmin and Long, Mingsheng},
booktitle={11th International Conference on Learning Representations, ICLR 2023},
year={2023},
}

@misc{TSL2024,
  title={Deep Time Series Models: A Comprehensive Survey and Benchmark}, 
  author={Yuxuan Wang and Haixu Wu and Jiaxiang Dong and Yong Liu and Mingsheng Long and Jianmin Wang},
  year={2024},
  eprint={2407.13278},
  archivePrefix={arXiv},
}

@misc{NBEATS2019,
  title={N-BEATS: Neural basis expansion analysis for interpretable time series forecasting}, 
  author={Boris N. Oreshkin and Dmitri Carpov and Nicolas Chapados and Yoshua Bengio},
  year={2020},
  eprint={1905.10437},
  archivePrefix={arXiv},
}

@article{FreTS2023,
  title={Frequency-domain MLPs are more effective learners in time series forecasting},
  author={Yi, Kun and Zhang, Qi and Fan, Wei and Wang, Shoujin and Wang, Pengyang and He, Hui and An, Ning and Lian, Defu and Cao, Longbing and Niu, Zhendong},
  journal={Advances in Neural Information Processing Systems},
  volume={36},
  pages={76656--76679},
  year={2023}
}

@article{chi2021t,
  title={TOHAN: A one-step approach towards few-shot hypothesis adaptation},
  author={Chi, Haoang and Liu, Feng and Yang, Wenjing and Lan, Long and Liu, Tongliang and Han, Bo and Cheung, William and Kwok, James},
  journal={Advances in neural information processing systems},
  volume={34},
  pages={20970--20982},
  year={2021}
}

@misc{TCN2018,
  title={An Empirical Evaluation of Generic Convolutional and Recurrent Networks for Sequence Modeling}, 
  author={Shaojie Bai and J. Zico Kolter and Vladlen Koltun},
  year={2018},
  eprint={1803.01271},
  archivePrefix={arXiv},
}

@inproceedings{SCINet2022,
 title = {SCINet: Time Series Modeling and Forecasting with Sample Convolution and Interaction},
 author = {LIU, Minhao and Zeng, Ailing and Chen, Muxi and Xu, Zhijian and LAI, Qiuxia and Ma, Lingna and Xu, Qiang},
 booktitle = {Advances in Neural Information Processing Systems},
 pages = {5816--5828},
 volume = {35},
 year = {2022}
}

@inproceedings{MiCN2023,
title={MICN: Multi-scale Local and Global Context Modeling for Long-term Series Forecasting},
author={Wang, Huiqiang and Peng, Jian and Huang, Feihu and Wang, Jince and Chen, Junhui and Xiao, Yifei},
booktitle={11th International Conference on Learning Representations, ICLR 2023},
year={2023},
}

@inproceedings{moderntcn2024,
  title={Moderntcn: A modern pure convolution structure for general time series analysis},
  author={Luo, Donghao and Wang, Xue},
  booktitle={The twelfth international conference on learning representations},
  pages={1--43},
  year={2024}
}

@inproceedings{Informer2021,
  title={Informer: Beyond efficient transformer for long sequence time-series forecasting},
  author={Zhou, Haoyi and Zhang, Shanghang and Peng, Jieqi and Zhang, Shuai and Li, Jianxin and Xiong, Hui and Zhang, Wancai},
  booktitle={Proceedings of the AAAI conference on artificial intelligence},
  volume={35},
  pages={11106--11115},
  year={2021}
}

@article{Autoformer2021,
  title={Autoformer: Decomposition transformers with auto-correlation for long-term series forecasting},
  author={Wu, Haixu and Xu, Jiehui and Wang, Jianmin and Long, Mingsheng},
  journal={Advances in neural information processing systems},
  volume={34},
  pages={22419--22430},
  year={2021}
}

@article{wang20241,
  title={Tackling noisy labels with network parameter additive decomposition},
  author={Wang, Jingyi and Xia, Xiaobo and Lan, Long and Wu, Xinghao and Yu, Jun and Yang, Wenjing and Han, Bo and Liu, Tongliang},
  journal={IEEE Transactions on Pattern Analysis and Machine Intelligence},
  volume={46},
  number={9},
  pages={6341--6354},
  year={2024},
  publisher={IEEE}
}

@article{Vaswani2017,
  title={Attention is all you need},
  author={Vaswani, Ashish and Shazeer, Noam and Parmar, Niki and Uszkoreit, Jakob and Jones, Llion and Gomez, Aidan N and Kaiser, {\L}ukasz and Polosukhin, Illia},
  journal={Advances in neural information processing systems},
  volume={30},
  year={2017}
}

@inproceedings{Ailing2023,
  title={Are transformers effective for time series forecasting?},
  author={Zeng, Ailing and Chen, Muxi and Zhang, Lei and Xu, Qiang},
  booktitle={Proceedings of the AAAI conference on artificial intelligence},
  volume={37},
  pages={11121--11128},
  year={2023}
}

@misc{tang2024,
  title={Unlocking the Power of Patch: Patch-Based MLP for Long-Term Time Series Forecasting}, 
  author={Peiwang Tang and Weitai Zhang},
  year={2024},
  eprint={2405.13575},
  archivePrefix={arXiv},
}

@inproceedings{PatchTST2023,
title={A Time Series is Worth 64 Words: Long-term Forecasting with Transformers},
author={Nie, Yuqi and Nguyen, Nam H. and Sinthong, Phanwadee and Kalagnanam, Jayant},
booktitle={11th International Conference on Learning Representations, ICLR 2023},
year={2023},
}

@misc{iTransformer2023,
  title={iTransformer: Inverted Transformers Are Effective for Time Series Forecasting}, 
  author={Yong Liu and Tengge Hu and Haoran Zhang and Haixu Wu and Shiyu Wang and Lintao Ma and Mingsheng Long},
  year={2024},
  eprint={2310.06625},
  archivePrefix={arXiv},
}

@misc{TimeXer2024,
  title={TimeXer: Empowering Transformers for Time Series Forecasting with Exogenous Variables}, 
  author={Yuxuan Wang and Haixu Wu and Jiaxiang Dong and Guo Qin and Haoran Zhang and Yong Liu and Yunzhong Qiu and Jianmin Wang and Mingsheng Long},
  year={2024},
  eprint={2402.19072},
  archivePrefix={arXiv},
}

@misc{AMD2025,
  title={Adaptive Multi-Scale Decomposition Framework for Time Series Forecasting}, 
  author={Yifan Hu and Peiyuan Liu and Peng Zhu and Dawei Cheng and Tao Dai},
  year={2025},
  eprint={2406.03751},
  archivePrefix={arXiv},
}

@misc{liu2021,
      title={Swin Transformer: Hierarchical Vision Transformer using Shifted Windows}, 
      author={Ze Liu and Yutong Lin and Yue Cao and Han Hu and Yixuan Wei and Zheng Zhang and Stephen Lin and Baining Guo},
      year={2021},
      archivePrefix={arXiv},
      primaryClass={cs.CV},
}

@article{Yifan2024,
title={Multi-Scale Transformer Pyramid Networks for Multivariate Time Series Forecasting},
author={Yifan Zhang and Rui Wu and Sergiu M. Dascalu and Frederick C. Harris},
journal={IEEE Access},
issue={ },
pages={14731-14741},
year={2024},
}

@inproceedings{stgcn, 
   series={IJCAI-2018},
   title={Spatio-Temporal Graph Convolutional Networks: A Deep Learning Framework for Traffic Forecasting},
   booktitle={Proceedings of the Twenty-Seventh International Joint Conference on Artificial Intelligence},
   publisher={International Joint Conferences on Artificial Intelligence Organization},
   author={Yu, Bing and Yin, Haoteng and Zhu, Zhanxing},
   year={2018},
   pages={3634–3640},
   collection={IJCAI-2018} }

@misc{mtgnn,
      title={Connecting the Dots: Multivariate Time Series Forecasting with Graph Neural Networks}, 
      author={Zonghan Wu and Shirui Pan and Guodong Long and Jing Jiang and Xiaojun Chang and Chengqi Zhang},
      year={2020},
      eprint={2005.11650},
      archivePrefix={arXiv},
}

@misc{agcrn,
      title={Adaptive Graph Convolutional Recurrent Network for Traffic Forecasting}, 
      author={Lei Bai and Lina Yao and Can Li and Xianzhi Wang and Can Wang},
      year={2020},
      eprint={2007.02842},
      archivePrefix={arXiv},
}

@misc{sae2018,
      title={Squeeze-and-Excitation Networks}, 
      author={Jie Hu and Li Shen and Samuel Albanie and Gang Sun and Enhua Wu},
      year={2019},
      eprint={1709.01507},
      archivePrefix={arXiv},
}

@inproceedings{Tsung2017,
title={Feature Pyramid Networks for Object Detection},
author={Tsung-Yi Lin and Piotr Dollár and Ross Girshick and Kaiming He and Bharath Hariharan and Serge Belongie},
booktitle={2017 IEEE Conference on Computer Vision and Pattern Recognition (CVPR)},
year={2017},
}

@inproceedings{Shazeer2017,
title={Outrageously large neural networks: The sparsely-gated mixture-of-experts layer},
author={Shazeer, Noam and Mirhoseini, Azalia and Maziarz, Krzysztof and Davis, Andy and Le, Quoc and Hinton, Geoffrey and Dean, Jeff},
booktitle={5th International Conference on Learning Representations, ICLR 2017},
year={2017},
}

@misc{timemoe2025,
      title={Time-MoE: Billion-Scale Time Series Foundation Models with Mixture of Experts}, 
      author={Xiaoming Shi and Shiyu Wang and Yuqi Nie and Dianqi Li and Zhou Ye and Qingsong Wen and Ming Jin},
      year={2025},
      eprint={2409.16040},
      archivePrefix={arXiv},
}


\end{document}